\documentclass[3p,final,11pt,authoryear]{elsarticle}

\usepackage{rotating}
\usepackage{amsmath}
\usepackage{tikz}
\usepackage{booktabs}
\usepackage{graphicx}
\usepackage{stackengine}
\usepackage{stackengine}
\usepackage{caption}
\usepackage{amssymb}
\usepackage{amsthm}
\usepackage{tablefootnote}
\usepackage{algorithm}
\usepackage[noend]{algpseudocode}

\usepackage{multirow}
\usepackage{txfonts}
\usepackage{mathdots}

\newsavebox{\measurebox}

\usepackage{lineno}
\usepackage{array,calc}
\usepackage{lscape}
\usepackage{float}
\usepackage[unicode]{hyperref}
\hypersetup{pdfauthor={Name}}
\usepackage[nameinlink,capitalize]{cleveref}
\usepackage{subcaption}

\floatstyle{plaintop}
\restylefloat{table}

\usepackage{xcolor}

\usepackage[symbol,multiple]{footmisc}
\usepackage{array,calc}
\usepackage{makecell}

\usepackage{pdfrender}
\usepackage{amssymb}
    \newcommand*{\boldcheckmark}{%
      \textpdfrender{
        TextRenderingMode=FillStroke,
        LineWidth=1pt,
      }{\checkmark}%
    }

\journal{***}

\begin{document}
\begin{frontmatter}
\title{A Coalition Game for On-demand Multi-modal 3D Automated Delivery System}
\author[1]{Farzan Moosavi\footnote{Corresponding author}}
\author[1]{Bilal Farooq}

\address[1]{Laboratory of Innovations in Transportation (LiTrans),

Toronto Metropolitan University, Toronto, Canada}

\begin{abstract}

In urban logistics, Unmanned Aerial Vehicles (UAVs) and Autonomous Delivery Robots (ADRs) present promising alternatives for user and cooperative delivery solutions, especially when it comes to on-demand delivery. We introduce a multi-modal autonomous delivery optimization framework as a coalition game for a fleet of UAVs and ADRs operating in two overlaying networks \textcolor{black}{to address last-mile delivery in urban environments, including high-density areas and time-critical applications. }
\textcolor{black}{In particular, a centralized dispatch system is designed to assign optimal delivery modes and solve the vehicle routing problem. The problem is defined as multiple depot pickup and delivery with time windows constrained over operational restrictions, such as vehicle battery limitation, precedence time window, and building obstruction.} 
\textcolor{black}{Utilizing the coalition game theory, we investigate cooperation structures among the modes to capture how strategic collaboration can improve overall routing efficiency}. To do so, a generalized reinforcement learning model is designed to evaluate the cost-sharing and allocation to different modes to learn the cooperative behaviour with respect to various realistic scenarios.
Our methodology leverages an end-to-end deep multi-agent policy gradient method augmented by a novel spatio-temporal adjacency neighbourhood graph attention network using a heterogeneous edge-enhanced attention model and transformer architecture.
Several numerical experiments on last-mile delivery applications have been conducted, showing the results from the case study in the city of Mississauga, which shows that despite the incorporation of an extensive network in the graph for two modes and a complex training structure, the model addresses realistic operational constraints and achieves high-quality solutions compared with the existing transformer-based and classical methods. \textcolor{black}{It can perform well on non-homogeneous data distribution, generalizes well on different scales and configurations, and demonstrates a robust cooperative performance under stochastic scenarios across various tasks, which is effectively reflected by coalition analysis and cost allocation to signify the advantage of cooperation}.

\end{abstract}

\end{frontmatter}
\section{Introduction}
\label{S:TInt}

\textcolor{black}{In recent years, on-demand delivery has gained remarkable attention. The market trend for online shopping demonstrates that the worldwide e-commerce sales are expected to top $7$ trillion in 2025, growing $15\%$ annually in North America since 2022 \citep{lvmtechNavigatingSuccess}. This could potentially cause unpredictable traffic congestion and delivery delays in urban networks as cities grow rapidly. To effectively cope with the resulting time delays, there is a growing push to reshape the on-demand urban delivery frameworks. For instance, in the meal delivery industry, on-demand delivery services like Uber Eats have been investigated by several studies to facilitate continuous operation, optimize routing efficiency and maintain customer satisfaction \citep{wang2023reinforcement, chen2023matching, mehra2023deliverai}. Nevertheless, their model incorporates ground riders, which cannot completely mitigate the long-term urban-related delivery issues. In particular, urban infrastructure layout and limited road network capacity significantly impact the timely and efficient last-mile delivery. Therefore, scaling the ground vehicles and current means of transportation will add more load to traffic flow and cause increased congestion, pollution, and risk of accidents \citep{mohammad2023innovative}. 
}

\textcolor{black}{
Other use cases include deliveries during medical emergencies and natural disasters \citep{shi2022bi}, as well as demand-responsive and urgent deliveries in high-density areas \citep{he2022route}. These cases underscore the need to leverage air mobility to alleviate congestion and shorten delivery times. In this regard, autonomous air delivery and unmanned aerial vehicles (UAVs) have become an attractive solution to reduce delivery time and cost, access remote and difficult-to-reach areas, and bypass city traffic and inaccessible roads \citep{archetti2021recent}. Furthermore, their flexibility allows congestion-free delivery without recurrent stops \citep{beliaev2023congestion}. A variety of UAV on-demand last-mile delivery in urban environments has been explored. For example, parcel and meal delivery considering city structure, although all paths are considered point-to-point direct lines and have been designed for problem-specific delivery \citep{liu2019optimization, elsayed2020impact}. On the other hand, there exist studies of on-demand delivery in high-density areas which incorporated a conflict-free aerial network considering urban infrastructures rather than a straight line by building a layout-inspired and road-based network \citep{mohamed2018preliminary}, corridor-based route network planning, \citep{he2022route}, and optimal height delivery in multiple flight levels \citep{kim2024drone}, though mostly simulated a small case network and real-world delivery scenarios like peak hours demands have not been considered. In this regard, a conflict-free, scalable aerial network independent of physical city structure is required for autonomous on-demand urban delivery. }

\textcolor{black}{
Nevertheless, practical issues have been identified that degrade the UAVs' performance and restrict them from being the most effective solution in urban delivery scenarios. Regardless of noise and privacy issues, air regulations like no-fly zones and operational layers, including limited capacity and battery, directly affect the level of service. Despite their fast and agile delivery, these shortcomings may not reflect UAVs' point of strength in cluttered urban areas. For example, a limitation of physical landing pads exclusively for a UAV landing in case of very low-level urban airspace delivery \citep{doole2020estimation}, in addition to the excessive energy consumption required for several take-offs and landings in multiple short-range flights in nearby locations \citep{zhang2021energy}.  
As an alternative, collaborative delivery is identified as one of the promising solutions to exploit the individual strengths of other vehicles to overcome UAVs' practical challenges. Namely, truck-drone hybrid delivery has been proposed for no-fly zone areas for the limited payload \citep{jeong2019truck}, and truck-and-robot multi-modal delivery \citep{ostermeier2023multi} is suggested to save more cost and reduce traffic congestion. However, when it comes to inner-city networks, where dense residential areas exist, trucks bring more congestion. Autonomous Delivery Robots (ADRs), on the other hand, are another delivery candidate. They are designed to operate at low speeds, e.g., pedestrian speed, safely share existing sidewalks and bike lanes with people and provide service for a limited network \citep{alfandari2022tailored}. As such, a more efficient multi-modal choice for such an environment, a hybrid UAV-ADR, is proposed in a simple showcase one-to-one matching, leading to a higher level of service  \cite{samouh2020multimodal}. Hence, we leverage a collaborative UAV-ADR system following this study as a multi-modal delivery to address more realistic urban last-mile delivery scenarios. 
}

\begin{figure}[!ht]
\centering
     \includegraphics[width=\textwidth]{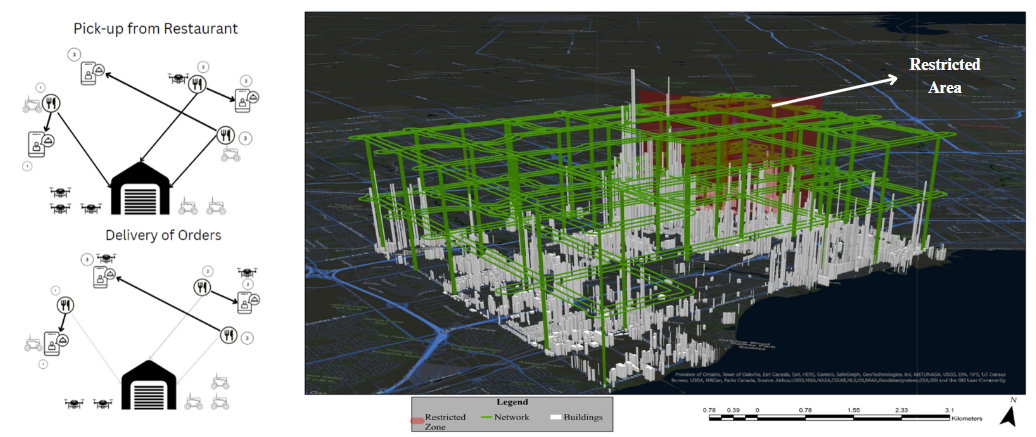}
     \caption{Aerial network for UAVs, including a  restricted zone marked as a red arced region}
     \label{fig8b}
\end{figure}

\textcolor{black}{
In the context of on-demand delivery, the research problem aims to optimize the operational cost for the city-scale delivery of food or medical supplies, where packages from pickup locations, like restaurants and medical centers, are to be delivered by a third-party or online delivery service to customers' delivery locations. Initially, the sequential decision-making optimization scheme is designed as a centralized controller to constitute the node visiting sequence of each vehicle that minimizes the vehicles' travel time and customers' waiting time during the peak hour demand in an urban network. Subsequently, a collaborative delivery is offered that benefits from the synergy and interaction of each mode in the system, first to distribute the workload between vehicles and secondly to investigate its impact on the cost savings of individual modes.
}
We assume the ADRs operate on the existing sidewalk/bike lane network. Similarly, a predefined virtual network is extended in the air (3rd dimension) for the operations of UAVs managed by the municipality. Figure \ref{fig8b} represents a pickup and delivery environment for the UAVs' operation network in the city of Mississauga. 
As a result, given the battery-based vehicles and realistic aspects of urban delivery, this problem can be classified as a collaborative electric capacitated multi-vehicle pick-up and delivery problem with time windows (CE-CPDPTW).

\textcolor{black}{
It is noted that throughout this research, the term collaborative means that each mode works together and cooperates in the form of groups towards a common objective by collective decision-making rather than individual strategic choices. In the field of game theory, this is also called a coalition game and aims to model the players' cooperation behaviour given heterogeneous agents and the complexity of the environment \citep{ahmad2023applications}.
Specifically, in urban last-mile delivery, collaborative delivery has gained increasing interest to reduce transportation costs and improve operational efficiency \citep{gansterer2020shared,zhang2022heterogeneous}, especially to economically incentivize individuals' cooperation by fair cost allocation to balance cost savings \citep{pingale2024collaborative}. However, there is a gap in the collaborative UAV-ADR delivery literature, and customer service sharing in the CE-CPDPTW application of last-mile delivery has not been sufficiently studied, specifically in what real-world urban scenarios vehicle coalitions are beneficial and how their joint effort can be divided among each player. Therefore, solving CE-CPDPTW, we aim to obtain the sub-coalitions of two modes in which the cooperation gain is higher than the individual gain.  
While each participant can act independently with local observation, in this work, all players require full information sharing since they are managed by a central dispatcher.  
}

\textcolor{black}{
In terms of the vehicle routing problem (VRP), heuristic methods were among the earliest approaches to solving the VRP variant. They are, however, frequently problem-specific and manually developed and cannot generalize or scale to other problem settings \citep{das2020synchronized, alfandari2022tailored, gu2023dynamic}. 
On the other hand, deep reinforcement learning (DRL) approaches have been presented as a powerful tool to model complex objectives given complicated constraints and uncertainty in last-mile delivery problems \citep{james2019online}. 
Specifically, they can handle larger-scale problems with different constraints and outperform traditional and metaheuristic algorithms and generalize from trained models to solve VRP instances of varying sizes and configurations \cite{li2022overview, bogyrbayeva2022learning}. Furthermore, the majority of the recent efforts on DRL-based VRP have employed encoder-decoder deep learning architectures to simulate sequential decision-making. For instance, transformers  \citep{vaswani2017attention}, for node sequence decoding \citep{kool2018attention} and graph attention networks (GAT) \citep{velickovic2017graph}, for graph encoding \citep{lei2022solve}, have exhibited promising alternatives in terms of computation time, generalization, and dealing with stochasticity due to capability of strong information embedding in graph structure using attention mechanisms. Inspired by them, we aim to solve CE-CPDPTW for the various applications of on-demand services, such as food and medicine delivery, by proposing a centralized deep multi-agent reinforcement learning (MARL) approach optimization. To model the sequential decision-making, we employ a heterogeneous, edge-enhanced graph attention network encoder based on temporal and spatial features and a transformer architecture decoder for a multi-modal fleet conditioned on the urban environment.}

\textcolor{black}{Finally, the notion of the core coalition game is brought to this study first to evaluate the cooperating efficiency of two modes if not to operate individually, and second, to promote single-mode last-mile delivery companies to horizontally collaborate upon sharing information or to form a coalition and benefit from both UAV and ADR to distribute the operational and customer service cost to each coalition. To the best of the author's knowledge, this is the first study to propose a collaborative autonomous UAV-ADR fleet addressing the urban-related on-demand last-mile delivery applications, firstly, by introducing a dual GAT-transformer mechanism with priority-based and conflict-free mode assignment, followed by a coalitional evaluation and cost allocation mechanism to incentivize the agents to illustrate the advantage of working in coalitions across various case studies.  
}
\textcolor{black}{
To summarize, the main contributions of our work are
listed as follows:
}
\textcolor{black}{
\begin{itemize}
    \item We tackle the CE-CPDPTW problem for a multi-modal and multi-depot fleet with limited payload capacity and battery size operation for pickup and delivery problems with a time window to minimize the delay and travel time, which accounts for real-world urban delivery cases, such as time-critical and non-uniform distribution delivery, including weather uncertainty and urban layouts.
    \item We propose an end-to-end multi-agent reinforcement learning methodology with a dual edge-enhanced spatial-temporal-aware encoder incorporated with an extensive yet effective graph attention module to differentiate the multi-modal network. Then, we generate the visiting node sequence of the optimal travel tour by adaptive speed routing using a transformer decoder through a conflict resolution layer. 
    \item Furthermore, a coalition game is defined to capture the interactive behaviour of the system to obtain potential coalitions to cooperate, if they exist, to suggest accommodating UAV-ADR combination and provide the managerial insights for online delivery services for performance attribution and regulatory trade-off of their fleet by evaluating the impact of the cost allocation mechanism of the savings on individual modes' collaboration incentive.
\end{itemize}
}

The rest of this paper is organized as follows: a review of relevant studies \textcolor{black}{that have motivated our design choices from a methodological and collaborative perspective of urban last-mile delivery} provided in \cref{S:tBack}. \cref{S:t3} briefly discusses problem statements, mathematical formulation, \textcolor{black}{and coalitional game theory preliminaries}. Methodology and detailed GAT-transformer architecture are described in \cref{methodology}. The application of our proposed framework through an extensive experimental result analysis,  \textcolor{black}{as well as cooperative game evaluation,} is discussed in \cref{S:t5}. Finally, \cref{S:t6} is dedicated to conclusions, final remarks, and future works.

\section{Background}
\label{S:tBack}

This section reviews the existing literature \textcolor{black}{on the CE-CPDPTW routing problem}, specifically when utilizing encoder and decoder DRL architectures. In addition, \textcolor{black}{collaborative delivery strategies as well as the} coalition game applications \textcolor{black}{regarding the urban last-mile delivery} are reviewed. \textcolor{black}{The research on modelling the on-demand last-mile delivery applications and incorporating complex real-world constraints is well established. However, to the best of our knowledge, there are only a limited number of studies that focus on solving the CE-CPDPTW problem using two interactive modes of UAVs and ADRs, particularly addressing realistic urban delivery. }

\textcolor{black}{
\subsection{CE-CPDPTW Approaches}
}
\cite{chu2021data} considered how food delivery platforms can solve the joint order assignment and routing problem of last-mile delivery service in on-demand delivery using efficient mini-batching gradient and simulated annealing algorithm. Moreover, \cite{liu2019optimization} comprehensively designed a dynamic rolling horizon for UAV food delivery. Although practical factors such as the orders’ location uncertainty, variable demand, carrying capacity, battery consumption, and battery swapping operation have been considered, the computational cost exponentially grows for larger networks in both examples. Additionally, the CE-CPDPTW problem for last-mile delivery is constrained by pairing and precedence relationships, urban physical layout, and time window delivery. Due to its NP-hard nature, it remains difficult for conventional methods, including exact and heuristic algorithms, to solve it optimally in a short computation time \citep{zong2022mapdp}, in which these constraints limit the solution space. In contrast, reinforcement learning (RL) approaches \textcolor{black}{in combination with sequential deep learning models} are proposed to automatically learn the rules in traditional heuristic methods for solving routing problems, which produces results with much faster computation. 

Initially, RL has found applications in fleet dispatching and on-demand delivery and various variants of vehicle routing problems. For example, \cite{jahanshahi2022deep} proposed a food delivery service as a Markov decision process (MDP) using deep Q-networks (DQN) to optimize courier assignment. Likewise, \cite{mehra2023deliverai} introduced DeliverAI, a multi-agent reinforcement learning system that allows food delivery networks to provide dynamic routing by a novel distributed Q-learning path-sharing algorithm. However, the training process becomes more complex in both cases as models require specialized tuning and can be time-consuming due to limited buffer memory.

\textcolor{black}{Deep reinforcement learning, in addition, provides powerful memory-based architectures to account for sequence-to-sequence model learning, such as node visit sequence. These models are based on an encoder-decoder policy network,} and mostly use policy-based methods such as the REINFORCE algorithm or actor-critic architecture to evaluate an action using a combined value-based approach \citep{williams1992simple}. The encoder processes the initial features and maps them to a high-dimensional representation. Subsequently, the decoder outputs each action step by step by extracting information from the internal memory by comparing it with a critical baseline. \textcolor{black}{The first deep learning model for the sequential decision-making solution of VRP is introduced by \cite{vinyals2015pointer}, a pointer network for the travelling salesperson problem (TSP). Following this work, \cite{nazari2018reinforcement} argued that the pointer network fails in static features decoding; instead, they utilize the recurrent neural network (RNN) decoder coupled with an attention mechanism. Nevertheless, a wide variety of research has used transformers for both encoder and decoder due to the attention architecture, which can be designed to learn the node relationships and enhance the solution quality in terms of time, generalization, and scalability following \cite{kool2018attention}. 
}
\textcolor{black}{For instance, \cite{li2021heterogeneous} suggested a heterogeneous attention-based network to learn the pickup and delivery relation using six types of heterogeneous attentions of different nodes. A more general notion of this notion can be found in \cite{lowens2022solving}, where heterogeneous attention is sparsified based on chains of precedence constraints. However, their encoder embedding included only node features. To this end, \cite{liu2023edge} proposed a UAV-based pickup and delivery problem, where attention layers are characterized by dot-product stochastic edge features merging into the node embedding. They represented a real-world case study delivery subjected to wind, and a stochastic edge-enhanced technique accounts for non-Euclidean distance. Their model outperforms \cite{li2021heterogeneous} and \cite{kool2018attention}, though their scenario involves one-to-one matching and without a capacity constraint.}

\textcolor{black}{
In particular, taking the edge information into account in the graph structure and residual connections between encoder layers has been shown to be more effective in VRP problems by \cite{lei2022solve} and \cite{fellek2023graph}. Their research shows that edge embedding-based message aggregation has superior graph topology representation power due to its inherent architecture of sharing information through edges. This mechanism can help refine edge embedding based on its connectivity in the graph. However, in these studies, the entire graph has been incorporated in the GAT by distance-based edge weight, and all customers ended up adding unnecessary embedding for faraway nodes. Specifically, this can be worse when a node contains spatial and temporal features. For example, if two nodes are in a close distance in the graph, but they have a large gap in their time window visit, distance-based message passing does not reflect their true relationship, as they can be prioritized in encoding over those nodes which have an actual closer time window. In this regard, \cite{zhang2023graph} has demonstrated that it is necessary to take advantage of graph neural networks to extract local spatial–temporal information to improve the solution. As a result, edge embeddings in the graph attention network will be used, and an adjacency-based masking for close spatial and temporal features is applied to capture the coupling distribution of the location and demand time window, especially when there are two modes with two different networks. Another aspect of the importance of edge-enhancement is to take the urban network into account, which is more similar to a road-style network, where not necessarily a straight line can be found to link the nodes. For example, there might be several intermediate nodes connecting a pickup to its delivery node, where they are mostly considered as a direct Euclidean distance in the literature. In fact, \cite{james2019online} proposed pickup and delivery vehicle routing with a pointer network, which considers network-based routing utilizing the entire graph, not just the customer node, for the cost adjacency matrix. Likewise, our study incorporates road-based network routing, where a larger graph is used from which a candidate subset of nodes is selected, and their connection is computed by Dijkstra's algorithm.}

\textcolor{black}{ 
In addition, most of the above-mentioned studies were simulated for single-vehicle systems and did not consider time window constraints for the pickup and delivery problem. Addressing more complex constraints, \cite{soroka2023deep} added more layers and created a multi-vehicle CPDPTW, and took an additional encoder for vehicles to enrich the initial feature embedding. Despite the novel design performing well on a large-scale network, their model objectives only minimize the routing cost.
In contrast, \cite{santiyuda2024multi}, \cite{zhang2022transformer}, and \cite{zhang2023two} presented the CPDPTW problem by multi-agent routing, simultaneously minimizing travel time and the late time arrival violation. Each has designed a unique transformer architecture to better handle the node relational information. For example, the encoding approach comprises a joint encoding scheme to encode the spatiotemporal information in the first two, and the latter uses a graph clustering method based on the hop-neighbourhood network. 
Nevertheless, although each work gives more importance to one of the real-world constraints of vehicle delivery, electric vehicle specifications like battery consumption, and the urban-related issues, such as incorporating physical structure to avoid obstacles, are ignored. }
\textcolor{black}{As a result, to address these challenges regarding CE-CPDPTW using the UAV-ADR system, the scope of our methodology is defined as follows.}

\begin{figure}[!ht]
\centering
     \includegraphics[width=1\textwidth]{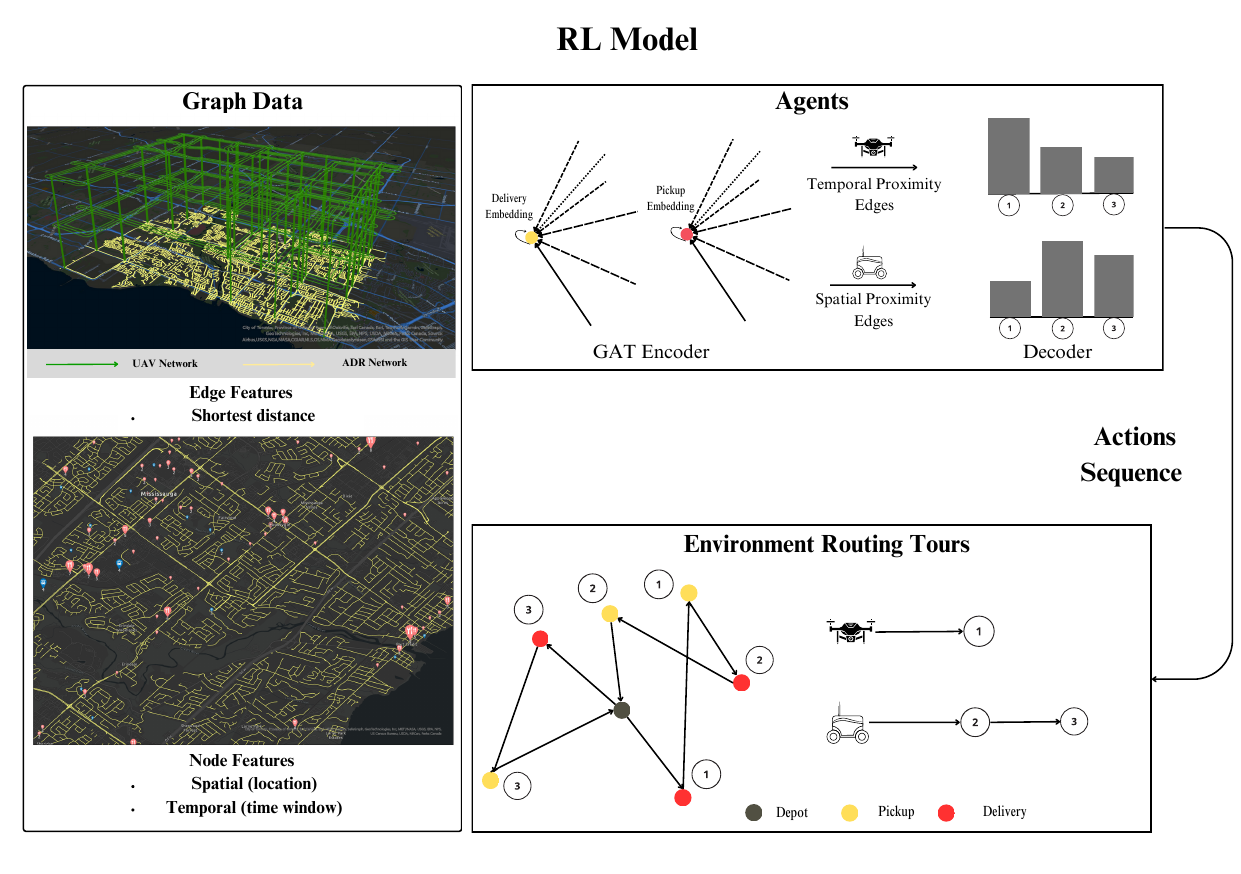}
     \caption{The reinforcement learning framework. Agents take a sequence of actions based on the reward feedback from the environment. Besides, the graph embedding encoder extracts the spatiotemporal joint correlation from the initial graph network features.}
     \label{fig11}
 \end{figure}

This study integrates a novel customized graph attention network encoder and a transformer decoder architecture using an end-to-end approach to solve CE-CPDPTW and minimize the delivery travel time and time delay. To do so, RL agents build up a tour in a sequential step through node to vehicle assignment with the highest probability based on the reward feedback from the decoder. Moreover, to better distinguish node-edge relationships for either of the modes in initial graph encoding, a dual heterogeneous encoder is introduced to account for precedence constraints and, more importantly, customers' spatial and temporal correlation, which focuses on coupling relation learning of location and time window.  
The flowchart of this methodology is shown in Figure \ref{fig11}.

\textcolor{black}{
\subsection{Collaborative Approaches}
}

\textcolor{black}{
The MARL collaborative framework for air mobility has been studied by \cite{fernando2023graph} in a decentralized manner under partial observations coupled with a fleet re-balancing mask to maximize the delivery fulfillment. Although decentralized training leads to learning independent policies for agents, and does need a global communication layer, fleet coordination can be challenging for a very large-scale fleet. Additionally, due to the sparse nature of the reward in such problems, training a decentralized cooperative policy is difficult, as it does not perform competitively \citep{park2021schedulenet}. On the other hand, a variety of studies presented centralized cooperative MARL with a global coordination layer. For instance, \cite{son2023solving} proposed a novel equity-transformer for min-max pickup and delivery routing problems, which contains two key inductive biases: multi-agent positional encoding for order bias and context encoder for equity context. The former considers additional depot nodes mimicking the single agent starting building tours. Conversely, the latter considers crucial factors such as temporal tour length, the target tour length, and the desired number of cities to be visited, thereby enhancing the fairness of the generated tours. 
Furthermore, \cite{fuertes2023solving} studied a system for cooperative is split into two stages: initial planning and routing solving. The initial planning proposes a clustering strategy to assign initial regions to each UAV, and the routing solving considers the initial and shared regions assigned during the initial planning to maximize the team reward for the orienteering problem. Moreover, \cite{zong2022mapdp} presented a cooperative MARL for minimizing the travel cost, and demonstrated a significant difference in performance for unbalanced node distribution in the case of halting time in the simultaneous decision-making to resolve a node conflict. Lastly, \cite{zhang2022transformer}
considered multiple ways of making action decisions, such as assigning all delivery tasks to vehicles based on a preset fleet order or the least travelling time. However, none of the research above has simulated a non-homogeneous fleet; therefore, no impact of interactive gain and cost allocation for each vehicle. In other words, there were no case studies indicating that a heterogeneous fleet can be exploited to minimize the system-level and common objective, such as customer waiting time. In fact, they either performed toward agent-specific cost, like minimum travel time or did not evaluate an effective combination of the fleet and profit distribution given real-world scenarios. 
}

Finally, a summary of the related studies is depicted in Table \ref{tab2}. According to the table, our research has brought several aspects that have yet to be addressed, such as considering the uncertain and stochastic nature of the delivery system and external factors impacting the level of service for a multi-modal fleet in urban real-world constraints. As a result, this study, to the best of the authors' knowledge, is the first that considers the on-demand food delivery problem modelled by collaborative electric-capacitated pickup and delivery with time windows (CE-CPDPTW) in the presence of wind and physical city structure.

\begin{table}[!ht]
\footnotesize
    \centering
    \caption{Summary of the most relevant studies in the literature}
\begin{tabular}{l|cc|cccc|ccc}
\hline \multirow[b]{2}{*}{Previous Work} & \multicolumn{2}{c}{Solution Approach} & \multicolumn{4}{c}{Problem Characteristic} & \multicolumn{3}{c}{Model Specification} \\
\cline{2-10} & H-GAT & TRL & EV & CVRP & PDPTW & UI & MA & EE & Stochasticity \\
\hline \cite{james2019online} &  &  & $\checkmark$ & $\checkmark$ & $\checkmark$ &  &  & $\checkmark$ & $\checkmark$ \\
\cite{zhang2022transformer} & $\checkmark$ & $\checkmark$ &  & $\checkmark$ & $\checkmark$ &  & $\checkmark$ &  & $\checkmark$ \\
\cite{zong2022mapdp} &  &  & $\checkmark$ &  &  & $\checkmark$ & $\checkmark$ &  & $\checkmark$ \\
\cite{lei2022solve} &  & $\checkmark$ & $\checkmark$ & &  &  &  & $\checkmark$ &  \\
\cite{zhang2023two} & $\checkmark$ & $\checkmark$ &  & $\checkmark$ & $\checkmark$ &  & $\checkmark$ &  &  \\
\cite{zhang2023graph} & $\checkmark$ & $\checkmark$ &  & $\checkmark$ & $\checkmark$ &  & $\checkmark$ &  &  \\
\cite{fuertes2023solving} & $\checkmark$ & $\checkmark$ & $\checkmark$ & $\checkmark$ &  &  & $\checkmark$ &  &  \\
\cite{santiyuda2024multi} & $\checkmark$ & $\checkmark$ &  & $\checkmark$ &   & $\checkmark$ & $\checkmark$ &  &  \\
\hline \textbf{Our Study} & $\boldcheckmark$ & $\boldcheckmark$ & $\boldcheckmark$ & $\boldcheckmark$ & $\boldcheckmark$ & $\boldcheckmark$ & $\boldcheckmark$ & $\boldcheckmark$ & $\boldcheckmark$ \\
\hline
\end{tabular}

    {{ H-GAT: Heterogeneous Graph Attention Network, TRL: Transformer Reinforcement Learning, EV: Electric Vehicle, CVRP: Collaborative Vehicle Routing Problem, PDPTW: Pickup and Delivery with Time Windows, UI: Urban Infrastructure EE: Edge-embedding, MA: Multi-agent}}
    \label{tab2}
\end{table}

\normalsize

\textcolor{black}{
In the next part, fundamental principles of coalition game theory, as well as cost allocation mechanisms, and how they can be applied to our problem, are discussed.}

\textcolor{black}{
\subsubsection{Coalition Game}
}\label{S:t22}

\textcolor{black}{
One of the popular ways to investigate such facets is the coalitional game with a group of cooperating players.
Coalition game theory has been applied to various centralized collaborative transportation and delivery problems, particularly focusing on cost-sharing mechanisms}. For instance, the vehicle routing and location-routing problems, \citep{osicka2020cooperative}. The study demonstrates that while core allocations are not guaranteed in collaborative location-routing scenarios, they frequently exist in most cases. Cooperative strategies can be beneficial as long as the core is not empty.  Moreover, the multi-depot vehicle routing problem \citep{zibaei2016cooperative} and shared charging station for electric vehicles \citep{wang2023collaborative} have also incorporated cooperative game theory, including the Shapley value and core analysis, showing that the cost can be fairly distributed.

The coalition game theory examines how groups are formed so that no member is incentivized to leave and play individually. \textcolor{black}{In such games, agents agree on a cooperative task, maximizing the common goal \citep{lui2010introduction}. A coalition game $ (K, C(S)) $ is defined by a set of players $K$, and $C$ is called the characteristic function, a real number that represents the cost resulting from a coalition $S\subset N$ in the game $(K, C(S))$, which is a group of cooperating players. If $S=K$, the $C(K)$ is the cost of forming the coalition of all users, known as the grand
coalition, with the assumption of $C(\emptyset)=0$. In addition, there is a cost allocation vector called user cost, $\phi_i$, which is the portion of $C(S)$ received by a player $i$ in coalition $S$ for each agent in coalitions \citep{chalkiadakis2022computational}.
}

\textcolor{black}{
The first goal is to identify which coalitions can be formed to yield less cost owing to cooperative behaviour. The second is to determine a fair cost allocation mechanism by which potential coalitions should divide the user payoff among their members.
Coalition game has several subclasses, such as monotone game, super-additive game, and convex game. Among which, a super-additive game is always profitable for two groups of players to join forces \citep{chalkiadakis2022computational}. A game is called super-additive if it is defined by Equation \ref{sup}, in which agents can always work without interfering with one another in disjoint coalitions, and no loss is involved in merging coalitions.
}
\begin{equation}\textcolor{black}{
C(S_1 \cup S_2) \leq C(S_1)+C(S_2) \quad \forall S_1, S_2 \subseteq K: S_1 \cap S_2=\varnothing }
\label{sup}    
\end{equation}

\textcolor{black}{Where $S_1$ and $S_2$ are sub-coalitions from the greatest coalition exist; all agents working the coalition, therefore, $K$ is the total number of agents. Specifically, the cooperative case of our problem is established such that coalitions can be formed in two separate groups of UAV and ADR and not jointly from both groups simultaneously. In other words, each distinct coalition can consist of only UAVs or ADRs in its group. It is assumed that either these vehicles are owned by different entities or a parent company is planning to rent a profitable combination of each fleet. Therefore, a super-additive game can find coalitions with no incentive to leave the grand coalition. For example, if the UAV and ADR fleet sets are shown by $S_U$ and $S_A$, respectively, the ground coalition is then $K = S_A \cup S_U$. Therefore, CE-CPDPTW can be formed as a super-additive game if Equation \ref{sup} holds for the given coalition sets and their cost. The analysis of such games can be twofold. First is stability, according to \cite{roger1991game}, the core of a game is defined as all allocation costs, such that no coalition wants to deviate from the grand coalition. The core can be empty as well, indicating no preference over cooperation. For the non-empty core, there are two necessary conditions regarding cost allocations, efficiency and rationality, which are shown in Equations \ref{eff} and \ref{rat}, respectively. 
}

\begin{equation}\textcolor{black}{
    \sum_{j \in N} \phi_j=C(K) }
    \label{eff}
\end{equation}
\begin{equation}\textcolor{black}{
    \sum_{j \in S} \phi_j \leq C(S) \quad \forall S \subseteq K}
\label{rat}
\end{equation}

\textcolor{black}{
Upon satisfying these conditions, the core property of the game shows stability; thus, it incentivizes the formation of a grand coalition.  Moreover, Shapley value fairly distributes the cost $C(K)$ that could be obtained by the grand coalition \citep{roth1988shapley}, to the unique cost allocation vector as shown in Equation \ref{shape}. In this equation, each agent receives a proportional amount to their average marginal contribution, averaging over all the different sequences according to which the grand coalition could be built up from the empty coalition. 
}
\textcolor{black}{
\begin{equation}
\phi_i(K, C)=\frac{1}{K!} \sum_{S \subseteq K \backslash\{i\}}|S|!(|K|-|S|-1)![C(S \cup\{i\})-C(S)] 
\label{shape}    
\end{equation}
}

\textcolor{black}{
Therefore, for a given characteristic function and combination of disjoint groups, the Shapley value can be used to split revenue or savings, based on the cost function.
}


\section{Problem Description}
\label{S:t3}

\textcolor{black}{
The research problem is to minimize the vehicles' travel time and customer waiting time for on-demand last-mile delivery in an urban environment using two distinct sets of autonomous vehicles. To be more specific, a time-sensitive scenario, such as food delivery in peak hour demand at a city network level, is considered to be solved for the given multi-modal fleet. In what follows, the preliminaries and mathematical formulation are provided, followed by a discussion on a complete operational network in the case study for UAV and ADR.}

Initially, the delivery problem network comprises two unique graphs for a fleet of UAVs and ADRs based on the road network of an urban area. The nodes are represented by a directed graph for UAV network $G^d=(D^{\prime} \cup P \cup D, E^d)$, where $P=\left\{x_1, \ldots, x_n\right\}$ denotes the set of $n$ pickup nodes, $D=\left\{x_{n+1}, \ldots, x_{2n}\right\}$ as corresponding delivery nodes, and $D^{\prime}=\left\{x_0, \ldots, x_{|D^{\prime}|}\right\}$ are depots set. $x_i$ denotes the location of node $i$, and $|D^{\prime}|$ shows the number of depots. In addition, $E^d=$ $\left\{e^d_{ij} \mid (i, j) \in N \right\}$ denotes the set of edges connecting the locations and $N= \left\{ P\cup D \cup D^{\prime} \right\}$ represent all the nodes in the graph. The same representation is applied for the ADR graph network $G^r=(D^{\prime} \cup P \cup D, E^r)$ with the same nodes but different edges $E^r=$ $\left\{e^r_{ij} \mid (i, j) \in N \right\}$. The $i$th order demand and time window is denoted by $q_i$ and $\left[e_i, l_i\right]$ ($e_i$ for pickup and $l_i$ for delivery node), respectively with $q_i>0$ and $q_{i+N}=-q_i$, for corresponding delivery node. Each vehicle must serve the pickup and delivery of requests together, accounting for precedence constraints within the time window of each point; otherwise, they get delayed, and the penalty is considered. There are $N^d$ and $N^r$ UAVs and ADRs, respectively. 
The $k$th vehicle, $k \in\left\{1, \ldots, K \right\}$, where $K$ is total number of vehicles, has a capacity $Q_k$ and a battery size $B_k$. If the $k$th vehicle is used, it departs from a depot with a sequence of locations and returns to a depot after its final delivery or when it needs recharging. In other words, a delivery scenario has $n$ customer requests, and each is constituted into a pickup task $i \in P$ and a delivery task $i+n \in D$. A vehicle with a fully charged battery is sent to accomplish all the assigned tasks; the vehicle can be recharged at any depot. 

\textcolor{black}{
Furthermore, the initial graph is sampled from a larger graph that is generated randomly with a number of nodes five times greater than the size of pickup and delivery nodes, and edges are also generated randomly from half of the maximum possible number of edges to full size. By doing so, after sampling the delivery network graph, the nodes' locations remain the same, although the edges are not necessarily a direct link. In this regard, the travel time between any two nodes is the time on the road network based on the impedance weight computed by Dijkstra's algorithm. 
Note that the vehicle's graph is built upon the customer nodes (pickup and delivery), and every other node is considered an intermediary point that connects these nodes, at which vehicles would not stop.}
The UAV network varies by wind, meaning its impedance updates at different weather conditions. We adopted the energy consumption model for UAVs in the effect of the wind as proposed in \citep{liu2023edge}. Solving the vehicle delivery problem aims to find a route that minimizes the delivery time, including the delay respected to time windows \citep{santiyuda2024multi}, while being subjected to constraints regarding the delivery mission and vehicle properties. Further assumptions used in this study are listed as follows:
 
\begin{enumerate}
    \item Each customer is served by the same vehicle.
    \item \textcolor{black}{Both modes can serve more than one request simultaneously.}
    \item Each mode has a certain battery consumption and capacity, as well as the battery lower bound threshold to return to the nearest depot for recharging.
    \item Each mode can operate within its associated unrestricted network areas.
    \item Each vehicle can start from different depots at the first step.
    
\end{enumerate}

\subsection{Mathematical Formulation}

The mathematical notations and formulation of the Electric Pickup and Delivery Problem with Time Windows (CE-PDPTW) are proposed as follows, and Table \ref{tab3} depicts the variables and parameters definition in this notation. 
The objectives and constraints in the mathematical formulation are listed as follows. The objective function is shown in Equation \ref{eqobj}.

\begin{equation}
 \text{min}\; \alpha_1\sum_{k \in N^d} \sum_{(i, j) \in N} t_{i j k} X_{i j k} + \alpha_2\sum_{k \in N^r } \sum_{(i, j) \in N} t_{i j k} X_{i j k} +\sum_{i \in P \cup D}   \alpha_3\left|T_{i k}-e_i\right|+\alpha_4\sum_{i \in P \cup D}  max\left\{T_{i k}-l_i, 0\right\}
    \label{eqobj}
\end{equation}

Where $X_{ijk}$ is the binary variable equal to 1 if the vehicle $k$ travels from node $i$ to $j$. $t_{i j k}$ is time travelled by vehicle $k$ between node $i$ and $j$, $T_{i k}$ is the arrival time of the vehicle $k$ at node $i$, $\alpha_1$ and $\alpha_2$ are positive monetary conversion factors pf utilizing UAVs and ADRs respectively per hour. Also, the $\alpha_3$ and $\alpha_4$ are the monetary coefficients of waiting time penalty for customers that the delivery company has to compensate in the case of delayed delivery, in case of lateness at the pickup and delivery node, respectively. Additionally, the equations that constrain this objective function are discussed below. 

\begin{equation}
 \sum_{j=0}^{P \cup D} X_{i j k}=1,  k \in K \;  \forall i \in D^{\prime}
    \label{con1}
\end{equation}
\begin{equation}
 \sum_{i=0}^{P \cup D} X_{i j k}=1, \forall  k \in K\;  \forall j \in D^{\prime}
    \label{con2}
\end{equation}
\begin{equation}
 \sum_{k=1}^{N^r \cup N^d} \sum_{i=0}^N X_{i j k}=1, \forall i \in\{P \cup D\}
    \label{con3}
\end{equation}
\begin{equation}
 \sum_{i=0}^{P \cup D} X_{i j k}=\sum_{i=0}^{P \cup D} X_{i\left(j+n\right) k}, \forall j \in P, \; \forall  k \in K
    \label{con4}
\end{equation}
\begin{equation}
T_{i k} \geq e_i, \forall  k \in K, \forall i \in\{P \cup D\}
    \label{con5}
\end{equation}
\begin{equation}
 T_{i k} \leq T_{k, i+n}, \forall i \in\left\{P\right\}, \forall  k \in K 
    \label{con6}
\end{equation}
\begin{equation}
 T_{k i}+ t_{i j k} + T_{r}^kR_{ik} \leq T_{k j} + (1-M)X_{i j k}, \forall  k \in K, \forall(i, j) \in \{P \cup D\}
    \label{con7}
\end{equation}  
\begin{equation}
 0 \leq u^k_i \leq Q_k, \forall  k \in K, \forall i \in\{P \cup D\}
    \label{con8}
\end{equation}
\begin{equation}
u^k_i=u^k_j+q_j + (1-M)X_{i j k}, \forall  k \in K, \forall(i, j) \in \{P \cup D\}
    \label{con9}
\end{equation}
\begin{equation}
e^k_i-e_{i j k}+B^k\left(1-X_{i j k}\right) \geq e_j, \forall  k \in K, \forall(i, j) \in \{P \cup D\}
    \label{con10}
\end{equation}
\begin{equation}
e_{min} \leq e^k_i \leq B^k,  k \in K, \forall i \in \{P \cup D\}
    \label{con11}
\end{equation}
\begin{equation}\textcolor{black}{
 e_{i k} \ge e_{min} - M(1-R_{ik}), \; \forall i \in D^{\prime}, \forall  k \in K}
    \label{con14}
\end{equation}
\begin{equation}
e^k_i=B^k, \forall  k \in K, i \in D^{\prime}
    \label{con12}
\end{equation}
\begin{equation}
 X_{i j k} \in\{0,1\}, \forall(i, j) \in \{P \cup D\}, \forall  k \in K
    \label{con13}
\end{equation}
\begin{equation}\textcolor{black}{
 R_{ik} = 0, \;\forall i \notin D^{\prime}, \forall k \in K}
    \label{con15}
\end{equation}

The constraints in Equations \ref{con1} and \ref{con2} ensure the vehicles depart from and return to any depot, whether for recharging or ending the tour. The Equations \ref{con3} and \ref{con4} ensure the requests are only served once and by the same vehicle. The constraints in Equations \ref{con5}, \ref{con6}, \ref{con7} ensure that vehicles cannot start service at a location earlier than its early time window and cannot be later than the service at its corresponding delivery location, and $M$ is a large positive number. All vehicles must service the pickup location earlier than the delivery node. Besides, if the vehicle goes to the depot for recharging, it should wait at the depot to be recharged by $T_r^k$, which is the recharging time for vehicle $k$. The capacity constraint of the vehicles and the load balance update are given in Equations \ref{con8} and \ref{con9}, with $u^k_i$ denoting the $ k$th vehicle's load after the service at the $i$th location is done. Equations \ref{con10} and  \ref{con11} denote battery level update and upper and lower battery capacity constraints, respectively, where $e_{ijk}$ is the energy consumption of the vehicle $k$ travelling from node $i$ to node $j$. \textcolor{black}{Equations \ref{con14} and \ref{con12} hold for recharging criteria and battery state after visiting depots, with $e^k_i$ denoting the $ k$th vehicle's battery level after going to the $ i$th location. A binary variable of $R_{ik}$ is defined to track if vehicle $k$ recharges at node $i$. } Besides, $e_{min}$ is the minimum battery threshold below which the vehicle battery level cannot be. 
Lastly, Equation \ref{con13} notes that the binary variable, \textcolor{black}{and Equation \ref{con15} notes the recharging variable to be zero when not visiting a depot node. }

\begin{table}[!ht]
\footnotesize
    \centering
    \caption{Parameters and Definitions}
    \label{tab3}
    \begin{tabular}{cl}
        \hline
        \textbf{Parameter} & \textbf{Definition} \\
        \hline
        $D^{\prime}$ & Depot instances set \\
        $P$ & Set of pickup nodes \\
        $D$ & Set of Delivery nodes \\
        $N$ & Set of all nodes\\
        $E^d$ & Set of UAV's edges network \\
        $E^r$ & Set of ADR's edges network \\
        $[e_i,l_i]$ & Early and late time window \\
        $x_i$ & location of node $i$  \\
        $q_i$ & Demand of customer $i$ \\
        $N^d$ & Number of UAVs \\
        $N^r$ & Number of ADRs \\
        $K$ & Number of vehicles \\
        $Q_k$ & Capacity of vehicle $k$\\
        $B_k$ & Battery level of vehicle $k$\\
        $n$ & Number of customers \\
        $T_{i k}$ & Arrival time of the vehicle $k$ at node
        $i$ \\
        $T^k_r$ & Recharging time for vehicle $k$\\ 
        $t_{i j k}$ & Time traveled by vehicle $k$ between node $i$ and $j$ \\
        $e_{ijk}$ & Energy consumption of the vehicle $k$ travelling from node $i$ to node $j$. \\
        $e_{min}$ & The minimum battery capacity of vehicle $j$ \\
        $e^k_i$ &battery level of vehicle $k$ at node $i$ \\
        $u^k_i$ &Load of vehicle $k$ at node $i$ \\
        $X_{i j k}$ & Binary decision variable for whether the vehicle $k$ travels from node $i$ to node $j$ \\
        $R_{ik}$ &Binary variable, 1 if vehicle $k$ recharges at node $i$ (only allowed at depots)\\
        \hline
    \end{tabular}
\end{table}

\subsection{Setup and Stylized Case Study}

\begin{figure}[ht]
\centering
\begin{subfigure}[b]{0.45\textwidth}
    \centering
    \includegraphics[width=0.75\textwidth]{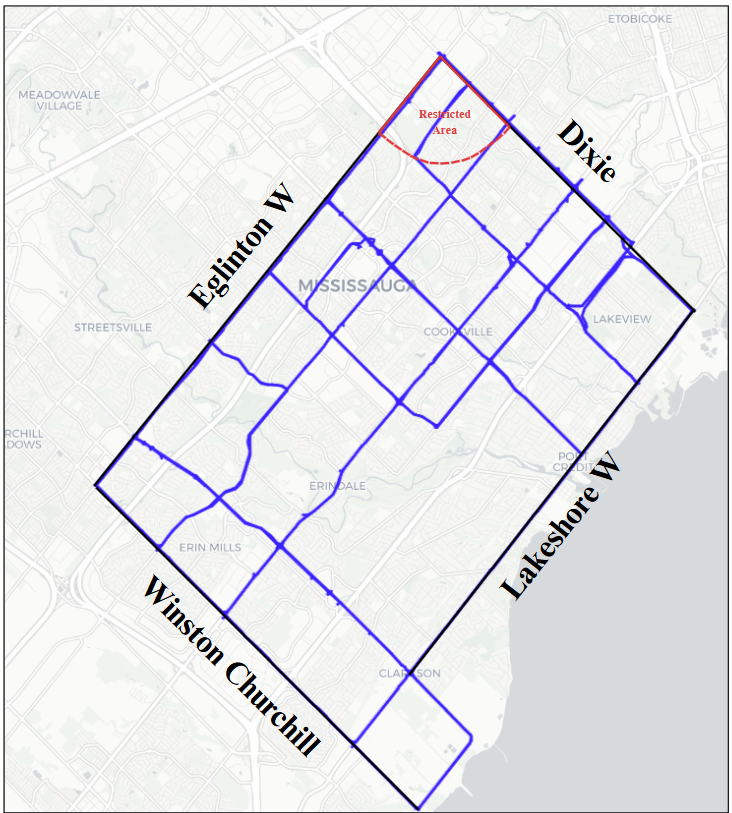}
    \caption{The network zone boundary.}
    \label{fig6a}
\end{subfigure}
\hfill
\begin{subfigure}[b]{0.45\textwidth}
    \centering
    \includegraphics[width=0.75\textwidth]{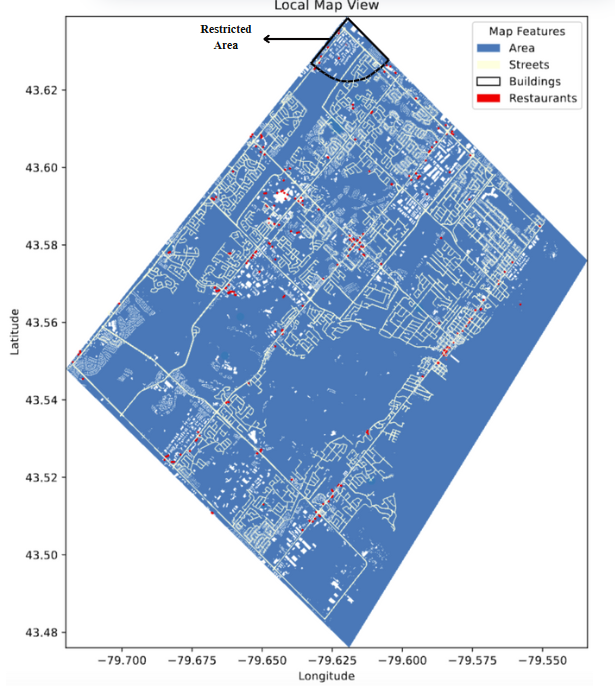}
    \caption{Mississauga city transportation network.}
    \label{fig6bb}
\end{subfigure}
\caption[Case study network]{Case study network.}
\label{fig6b}
\end{figure}

\begin{figure}[!ht]
\centering
     \includegraphics[width=1\textwidth]{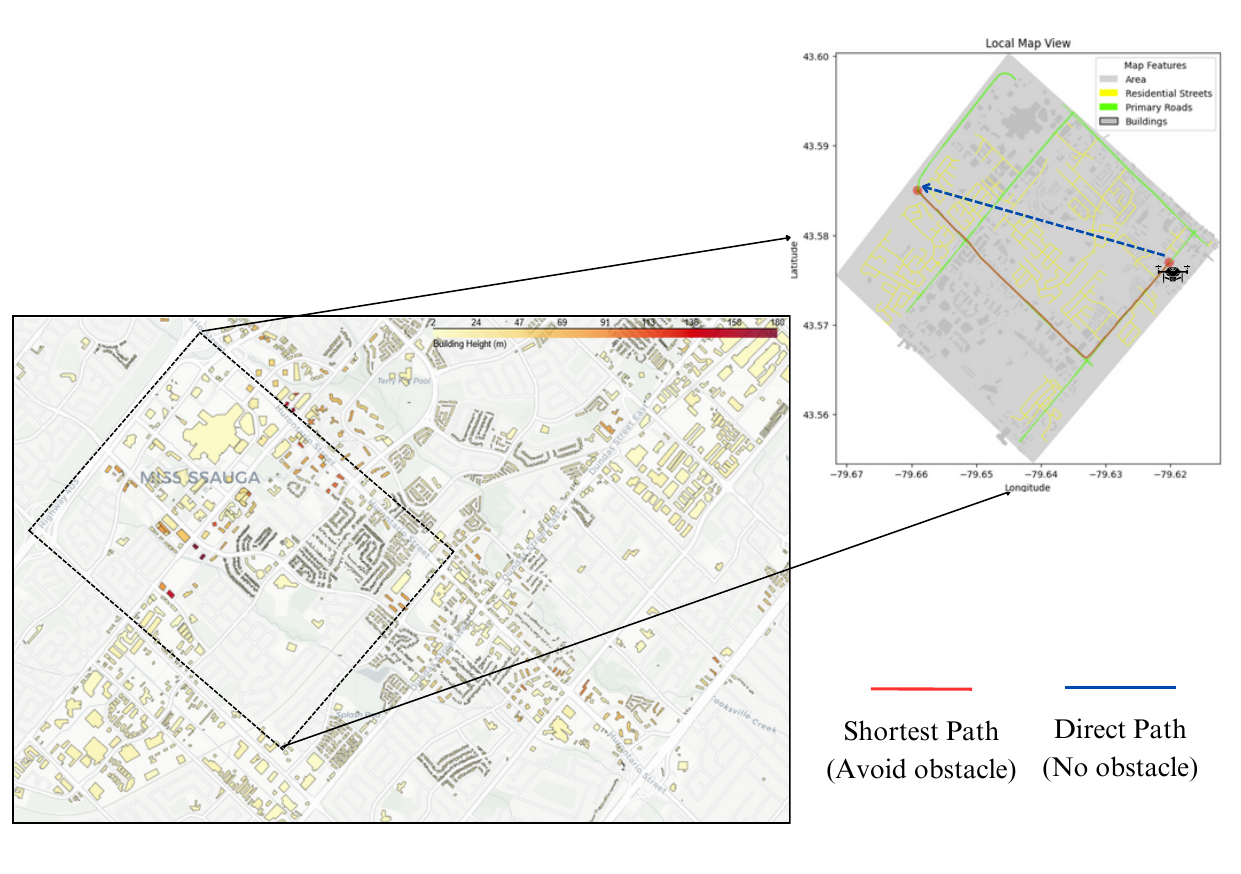}
     \caption{UAV Network Routing. This figure shows a top view of part of Mississauga with the pattern of the urban infrastructures labelled in yellow. The height heatmap is placed on the top right corner, which shows buildings' height from ground level to the highest building in the network. In case of delivery within a cluttered space, the shortest path instead of the direct path is used.}
     \label{fig8a}
\end{figure}

In this study, we adopt the transportation network of Mississauga, Canada (Figure \ref{fig6a}) to collect the essential information for data processing, such as nodes and edges of both ADR and UAV networks. 
We obtain data from the OpenStreetMap city map using OSMnx \citep{boeing2017osmnx} as the region of interest is shown in Figure \ref{fig6a}, where the two-dimensional projection of the UAV network is shown. Furthermore, a more extensive network of the case study is depicted in Figure \ref{fig6b} as the ADR operation network bounded to the residential area and traversing along any link except those leading to highways, like primary and secondary roads. This stems from the municipal and safety regulations and operational limitations due to battery constraints. Therefore, the only region allocated to the ADRs is designated residential areas where the ADR can move along sidewalks. The residential edges are in yellow in Figure \ref{fig8a}, which is part of the case study area, to better illustrate the network. Similar to ADRs, UAVs pose some limitations in certain areas. For example, the red arced region in Figure \ref{fig8b} shows the restricted circular area near Pearson airport, requiring at least a $5.6 \;km$ distance. As a result, they interactively operate to complement their delivery tasks by joint environment coverage.

On the other hand, the notion of the road network is adopted for aerial delivery. Primary and secondary roads are utilized as the main delivery routes for UAVs, which extend at multiple levels in the airspace with a limit of operation of 400 feet according to the \citep{Dronlaw}. This has the benefit of bypassing the high rises and city infrastructure, as can be visible in Figure \ref{fig6a}, which shows a blue route for direct delivery or a red path for the next shortest route delivery due to obstruction. The UAV will freely move horizontally and vertically along the tubes and traverse straight in the absence of obstacles.  
If municipal regulations permit, UAVs can travel through straight lines when the airspace is clear. Otherwise, they move through the edges of the elevated road network as depicted in Figure \ref{fig8b}.

Moreover, the street's intersections, restaurants, and parking facilities denote the delivery, pickup, and depot nodes, respectively, connected by corresponding edges, where each node is obtained from the OSMnx Python package. The graph data, as well as the footprint and height of the buildings, are collected utilizing the web-based data mining tool for Open Street Map, overpass turbine. \textcolor{black}{In addition, the demand order arrival time used in this study is based on the Poission distribution during the evening peak.}

\section{Methodology}
\label{methodology}

This section describes our solution approach for cost optimization and the cooperative strategy of the CE-CPDPTW problem. We first present the reinforcement learning Markov decision process, followed by the encoder and decoder design and the multi-agent policy gradient training algorithm for sequential routing learning. Lastly, the coalition game application to the CE-CPDPTW is defined through a pipeline to check how agents can work more efficiently by forming a coalition by harnessing the core properties and the proposed cost allocation mechanism.
In what follows, the problem model and dynamics are defined by the Markov decision process.

\subsection{Markov Decision Process}

A generic MDP consists of four components: state space $\mathcal{S}$, action space $\mathcal{A}$, reward function $r(s, a)$, and state transition probability $p\left(s^{\prime} \mid s, a\right)$. Each agent takes action in the environment and receives a signal reward and the probability of going to the next state, given its current action and state. 
Specifically, for the CE-CPDPTW environment described earlier in this paper, these components are defined as follows:

\begin{itemize}
    \item \textit{States:} Composed of the graph node state and the vehicle state. The node state is $s_t = (x_t,q_t,e_t,l_t)$, consisting of location, demand, and time window of the node at time $t$. The kth vehicle state $v^k_t$ is composed of its load $u^t_t$, battery level $e^k_t$, and traveled time $\tau^t$, expressed as $v_t=\left[ \tau^k_t, u^k_t, e^k_t\right]$ at a time $t$.

    \item \textit{Actions:} $a_t$ determines the vehicle's node selection at step t. The sequence of actions generated from the initial to the final step should be a combination of nodes starting and ending with the depots.

    \item \textit{Reward:} This research aims to minimize fleet delay and travel time. Therefore, the reward function is given in Equation \ref{eqre}.

\begin{equation}
 \; R=\alpha_1\sum_{k \in N^d} \sum_{(i, j) \in N} t_{i j k} X_{i j k} + \alpha_2\sum_{k \in N^r } \sum_{(i, j) \in N} t_{i j k} X_{i j k} +\sum_{i \in P \cup D}   \alpha_3\left|T_{i k}-e_i\right|+\alpha_4\sum_{i \in P \cup D}  max\left\{T_{i k}-l_i, 0\right\} + r^t_k
    \label{eqre}
\end{equation}

Another term, Equation \ref{bat}, is added to the reward when the agent's energy falls below the battery threshold, penalizing the agent whose battery has run out. Ensuring efficient behaviour and battery usage encourages the agents to avoid poor routing decisions, leading to increased operational costs and low-battery situations.  

\begin{equation}
    r^t_k = \lambda R^t_k
    \label{bat}
\end{equation}

\textcolor{black}{
Where $\lambda$ is a positive coefficient, and $R^t_k$ is the travel cost of agent $k$ at step $t$. Therefore, in case the battery exceeds the lower bound, $e_{min}$, a penalty proportional to the travel cost of visiting the unfinished customer is incurred.}
 
    \item \textit{Transition:} The system state will be updated from $s_t$ to $s_{t+1}$ based on the currently executed action $a_t$. The dynamic features of the problem, such as vehicle load, battery level, and travelled time, are being changed through consecutive nodes based on the vehicle's features (Equations \ref{tran1}, \ref{tran2}, \ref{tran3}, and \ref{tran4}). First, the system time is updated based on these equations. 

\begin{equation}
    \tau^{t+1}=\left\{\begin{array}{l}
\max \left(\tau^t, l_i\right)+ t_{i j k}, \text { if } i \in P \cup D \\
\tau^t+\left(B^k-e^k_t\right)/\eta_k+t_{i j k}, \text { if } i \in C 
\end{array}\right.
\label{tran1}
\end{equation}
where $\eta_k$ is the charging rate to charge the battery from the given level for any vehicle $k$. Next, the battery level of the vehicle is updated:

\begin{equation}
e^{t+1}=\left\{\begin{array}{l}
e^t-e_{i j k}, \text { if } i \in P \cup D \\
B^k, \text { if } i \in C
\end{array}\right.
\label{tran2}
\end{equation}

\textcolor{black}{The initial battery size is given and updated based on the energy consumed in travelling from node $i$ to $j$, $e_{ijk}$. The power consumption model for UAV and ADR batteries is discussed in \ref{B}. Their consumption model takes velocity and payload demand as input and computes the energy used in the battery.}   

Finally, the vehicles load $u^t$, and the remaining demand, $d_i^t$, at each node are updated as follows.

\begin{equation}
    u^{t+1}=\left\{\begin{array}{l}
u^t + d_i^{t}, \text { if } (i \in P ) \cap (\tau^t \in [e_i,l_i]) \\
u^t - d_i^{t}, \text { if } (i \in D ) \cap (\tau^t \in [e_i,l_i])\\
u^t, \text { if } i \in C
\end{array}\right.
\label{tran3}
\end{equation}

\begin{equation}
d_i^{t+1}=\left\{\begin{array}{l}
0, \text { if } (i \in P \cup D) \cap (\tau^t \in [e_i,l_i]) \\
d_i^t, \text { if } i \in C
\end{array}\right.
\label{tran4}
\end{equation}
            
    \item \textit{Policy:} The stochastic policy $p_\theta$ automatically selects a node at each time step. This process is repeated iteratively until all pickup-delivery services are completed with respect to the problem constraints. The final outcome engendered by performing the policy is a permutation of all nodes, which prescribes the order of each node for the vehicle to visit, i.e., $\pi=\left\{\pi_0, \pi_1, \ldots, \pi_T\right\}$. Based on the chain rule, the probability of an output solution is factorized as Equation \ref{eqpol}.
\begin{equation}
    P(\pi \mid s)=\prod_{t=0}^{T-1} p_\theta\left(\pi_t \mid s, \pi_{1: t-1}\right) 
    \label{eqpol}
\end{equation}

\textcolor{black}{where $s$ is the state input to the problem instance}. The decision-making about the node selection will be performed based on the learned $p_\theta$.
\end{itemize}

\textcolor{black}{This policy is trained by a policy gradient algorithm, which is a multi-agent system learning the sequential task of the delivery problem by graph attention and a transformer model.}

\subsection{Reinforcement Learning Model}
\label{S:t4}

\begin{figure}[!ht]
\centering
     \includegraphics[width=\textwidth]{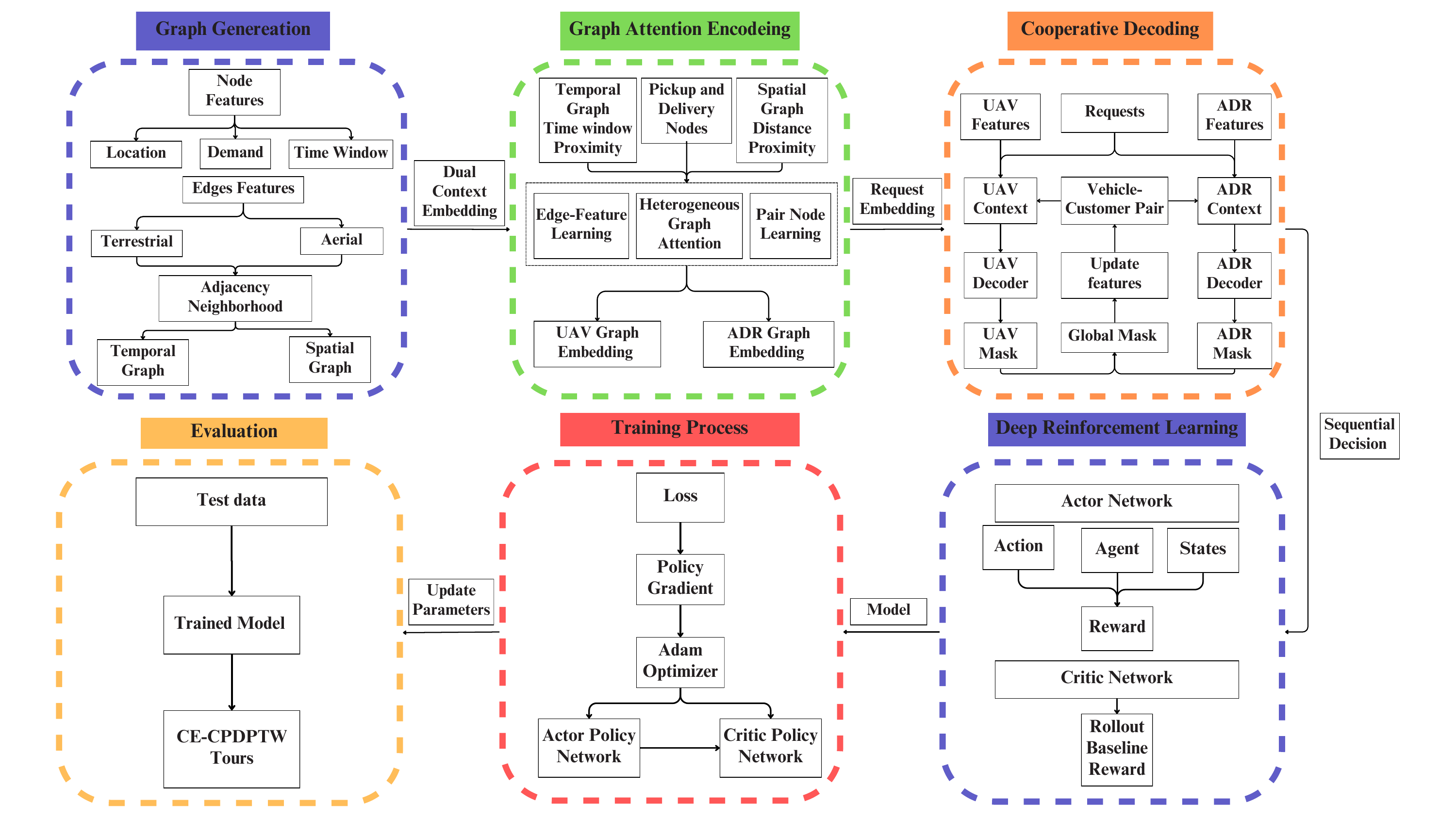}
     \caption{Overview of the methodology}
     \label{fig12}
\end{figure}

\textcolor{black}{
The overall framework of the RL model is to initially prepare the initial graph based on the problem states and each UAV and ADR network (aerial and terrestrial, as shown in Figure \ref{fig11}), as mentioned in the previous section. Subsequently, a MARL centralized controller is designed, comprising an encoder to extract the initial problem feature, and a decoder to learn the selection of nodes by contextual vehicle and routing information of the environment. Finally, an actor-critic training updates the policy parameters, leading to an optimal tour. Figure \ref{fig12} demonstrates the problem methodology flowchart}

\textcolor{black}{
In the graph generation phase, node information is the same for both modes, like terrestrial for ADRs and aerial for UAVs; however, the edge features are different in terms of weight of connectivity, meaning that a node's connections are the same, but with different weights. This Aerial network can be affected by urban building density. The more obstacles scattered along the customers' location, the less direct a path can be built (refer to Figure \ref{fig8a}; thereby, closer to the terrestrial network. These separate sets of edges, together with node features, are embedded by the dual encoder, which is discussed in the next section.}

\subsubsection{Graph Attention Encoder}

\noindent

\begin{figure}[!ht]
\centering
     \includegraphics[width=0.85\textwidth]{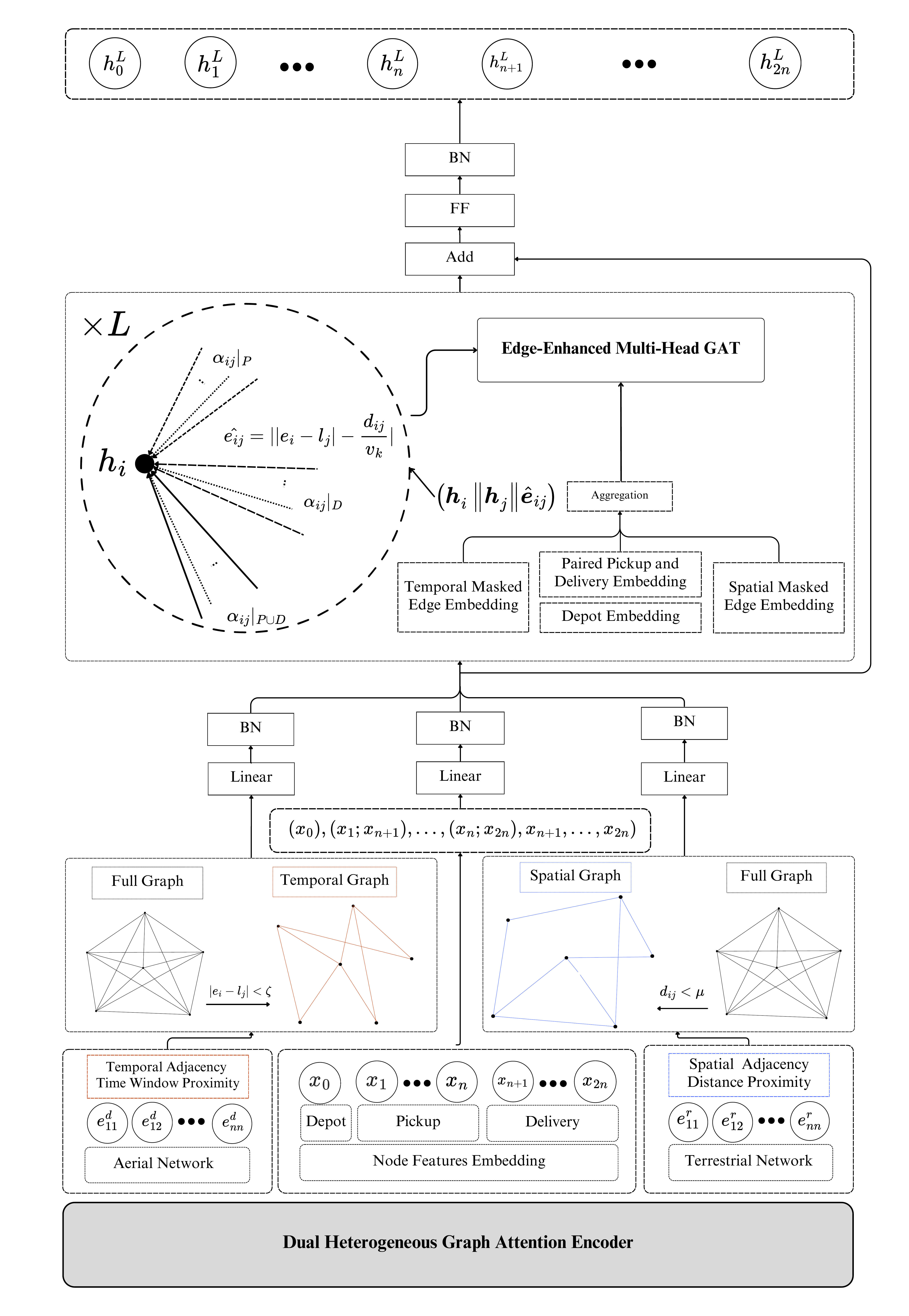}
     \caption{The encoder architecture and graph attention adjacency mechanism}
     \label{fig91}
 \end{figure}

\textcolor{black}{
The graph attention is utilized as an encoder which takes the node features and dual edge features and uses multi-head attention to encode node-edge interactions based on spatial-temporal dependencies across multiple layers. This problem is governed by heterogeneous temporal and spatial feature distribution, where there are non-linear correlations between the locations and the customer order arrival time (pickup time), where distance-based edge embedding into node-to-node attention cannot learn the complex coupling relationship among visiting locations in the problem. In this regard, a novel encoder using an adjacency mask mechanism based on a temporal and spatial graph, and feature-aware edge-enhancement to the node embedding is designed. Figure \ref{fig91} illustrates a detailed design of the encoder.}

\textcolor{black}{
It is noted that the spatial graph is the terrestrial graph with the shortest path distance as the edge weight. On the other hand, the temporal graph is featured by the time window value of each node rather than its location. This graph takes the aerial graph nodes; having said that, the edges are each node's time separation. In other words, the edge weights are the time window difference between nodes.}
\textcolor{black}{
The adjacency masking mechanism is provided to guide the multi-head attention layer to account for proximity nodes rather than irrelevant ones. The proximity nodes are defined based on the edge weight values in the graph. It can be defined for each node as a set of neighbour nodes with which there is an edge connection. Accordingly, the spatial and temporal graph proximity are determined as $NB^S_i=\left\{G^r\mid  \left\|x_{i}-x_{{j}}\right\|_2 \leq \mu, (i,j) \in N\right\}$ and $NB^T_i=\left\{G^d \mid |e_i-l_j| \leq \zeta, (i,j) \in N\right\}$, respectively, where $\|\cdot\|_2$ denotes the L2 norm. This way, local adjacency information of either graph is extracted to give more importance to nodes with more similar features.} 
In this regard, an adjacency matrix is defined as $A_{i, j}=1$ if there is a link between node $i$ and $j$, and zero otherwise. Therefore, the temporal and spatial connection for a node $i$ is defined by $\left\{A_{ij}| j \in{NB}^T_i\right\}$ and $\left\{A_{ij}| j \in{NB}^S_i\right\}$, which limits the neighbourhood in adjacency matrices with a time window by a threshold $\zeta$ and distance threshold by $\mu$. In addition, a parameter called density, $\rho$, is defined to determine the probability of how dense the environment is, particularly, the probability of the existence of an obstacle along two connecting nodes.\textcolor{black}{ This is the parameter that can either make two aerial and terrestrial networks (road networks) the very same for $\rho=1$ or make a fully directional connection network (like free airspace) for $\rho=0$}. A broad spectrum of customers' time and location correlation is extracted by varying thresholds in addition to different network embeddings for sparse and dense urban environments. 
First, the initial embedding for each node and fleet edges is computed in Equation \ref{eq:2} and \ref{eq:3} through a linear layer. \textcolor{black}{Depots, pickups, deliveries are encoded separately into the initial hidden embedding, $\mathbf{h}_i^0$, based on their features $\mathbf{s}$. }

\begin{equation}
\textcolor{black}{\mathbf{h}_i^0= 
\begin{cases}\operatorname{BN}\left(\boldsymbol{W}_0(\mathbf{s}_i ) + \boldsymbol{b}_0\right), & \text{ if } i \in D^{\prime}, \\ 
\operatorname{BN}\left(\boldsymbol{W}_1(\mathbf{s}_i ; \mathbf{s}_{i+n}) + \boldsymbol{b}_1\right), & \text{ if } i \in P, \\ 
\operatorname{BN}\left(\boldsymbol{W}_2(\mathbf{s}_i) + \boldsymbol{b}_2\right), & \text { if } i \in D\end{cases}
}\label{eq:2}
\end{equation}

Where \textcolor{black}{$\boldsymbol{W}_0$}, $\boldsymbol{W}_1$, $\boldsymbol{W}_2$ , $\boldsymbol{W}_3$,  $\boldsymbol{W}_4$, \textcolor{black}{$\boldsymbol{b}_0$}, $\boldsymbol{b}_1$, $\boldsymbol{b}_2$, $\boldsymbol{b}_3$, and $\boldsymbol{b}_4$ represents the learnable parameters and $BN(.)$ represent batch normalization. Similarly, two linear projections for edge features of UAV and ADR networks are used.

\begin{equation}
\mathbf{e^d}_{ij}= 
\operatorname{BN}\left(\boldsymbol{W}_3\hat{E^d}_{ij} + \boldsymbol{b}_3\right), \\
\mathbf{e^r}_{ij}= 
\operatorname{BN}\left(\boldsymbol{W}_4\hat{E^r}_{ij} + \boldsymbol{b}_4\right)
\label{eq:3}
\end{equation}

In this setup, we have defined the attention weight in a time dimension manner, capturing the vehicle's travelling feature as well. In other words, each connecting edge weight in the graph attention network is considered as the relative time that can be passed by a certain vehicle mode within the time window between two nodes. Therefore, $\hat{E^d}_{ij} = \big |e_i - l_j - \frac{d_{ij}}{v_d} \big |$ for UAVs, $d_{ij} \in E^d$ and $\hat{E^r}_{ij} = \big|e_i - l_j - \frac{d_{ij}}{v_r} \big|$ exist for ADRs, $d_{ij} \in E^r$. Where $d_{ij}$ is the shortest distance between node $i$ and $j$, $v_d$ and $v_r$ are the UAV and ADR maximum velocities, respectively. This configuration has been a trade-off between the case of time-window and distance proximity, by incorporating the network and vehicle specifications, leading to a distinguished and agent feature-aware embedding.

 \textcolor{black}{
Afterwards, a multiple GAT Convolution layer is utilized to aggregate the node and edge information to obtain the final node embedding. To capture the precedence constraint for pickup and delivery nodes, a joint encoding of pickup and delivery sets separately is fused into the attention mechanism. That is, two additional attention subsets specifically for the pickup and delivery network are aggregated to the node embedding. According to \citep{lei2022solve}, the attention score between the node and edge embedding for all sets of nodes and edges can be determined by Equation \ref{GAt}.}

\begin{equation}
\alpha_{i j}^{\ell}=\displaystyle \frac{\exp \left(\sigma\left(\boldsymbol{g}^{\ell^T}\left[\boldsymbol{W}^{\ell}\left(\boldsymbol{h}_i^{(\ell-1)}\left\|\boldsymbol{h}_j^{(\ell-1)}\right\| {\boldsymbol{e}}_{i j}\right)\right]\right)\right)}{\sum_{z} \exp \left(\sigma\left(\boldsymbol{g}^{\ell T}\left[\boldsymbol{W}^{\ell}\left(\boldsymbol{h}_i^{(\ell-1)}\left\|\boldsymbol{h}_z^{(\ell-1)}\right\| e_{i z}\right)\right]\right)\right)}
    \label{GAt}
\end{equation}

\textcolor{black}{
where $(\cdot)^T$ represents transposition, $\cdot \| \cdot$ is the concatenation operation, $\boldsymbol{g}^{\ell}$ and $\boldsymbol{W}^{\ell}$ are learnable weight vectors and matrices respectively, and $\sigma(\cdot)$ is the LeakyReLU activation function However, our GAT attention scores between nodes are computed based on the node role through the weighted message passing mechanism by $a^{ij}$ at $l$th layer as in Equations \ref{eq:43}, \ref{eq:5}, and \ref{eq:55}, for all, pickup, and delivery nodes attentions.}
\textcolor{black}{
Additionally, the temporal and spatial graph edges subset is used to mask the message passing of nodes that are not in the proximity graph for UAVs ($NB^T$) and ADRs ($NB^S$), respectively. This can let ADR focus on close-range distances due to the speed limitation and not be impacted by long travel time delivery. On the other hand, UAV can go to a wide range of distances and are more efficient in cases where the pickup and drop-off locations are far, but in a near time window. Therefore, the attention scores for nodes with a wider time window are masked.} 
By doing so, each pickup node would be influenced by the corresponding delivery and learn the heterogeneous relation between the location-time proximity of other pickup nodes.

\begin{equation}
\textcolor{black}{\alpha_{i j}^{\ell} = \begin{cases}
    \alpha_{i j}^{\ell}|_{A}^{UAV}, & \text { if } z = {NB}^T_i, \;\; e_{ij} = \mathbf{e^d}_{ij} \\
    \alpha_{i j}^{\ell}|_{A}^{ADR}, &\text { if } z = {NB}^S_i,  \;\; e_{ij} = \mathbf{e^r}_{ij}
\end{cases} 
\;\;\;\;  \forall (i,j) \in N, \; \boldsymbol{W^l} = \boldsymbol{W_A}, \; \boldsymbol{g^l} = \boldsymbol{g_A}
}\label{eq:43}
\end{equation}

\begin{equation}
\textcolor{black}{\alpha_{i j}^{\ell} = \begin{cases}
    \alpha_{i j}^{\ell}|_{P}^{UAV}, & \text { if } z = {NB}^T_i, \;\; e_{ij} = \mathbf{e^d}_{ij} \\
    \alpha_{i j}^{\ell}|_{P}^{ADR}, &\text { if } z = {NB}^S_i,  \;\; e_{ij} = \mathbf{e^r}_{ij}
\end{cases} 
\;\;\;\;  \forall (i,j) \in P, \; \boldsymbol{W^l} = \boldsymbol{W_P}, \; \boldsymbol{g^l} = \boldsymbol{g_P}
}\label{eq:5}
\end{equation}

\begin{equation}
\textcolor{black}{\alpha_{i j}^{\ell} = \begin{cases}
    \alpha_{i j}^{\ell}|_{D}^{UAV}, & \text { if } z = {NB}^T_i, \;\; e_{ij} = \mathbf{e^d}_{ij} \\
    \alpha_{i j}^{\ell}|_{D}^{ADR}, &\text { if } z = {NB}^S_i,  \;\; e_{ij} = \mathbf{e^r}_{ij}
\end{cases} 
\;\;\;\;  \forall (i,j) \in D, \; \boldsymbol{W^l} = \boldsymbol{W_D}, \; \boldsymbol{g^l} = \boldsymbol{g_D}
}\label{eq:55}
\end{equation}

where $\boldsymbol{g_A},\boldsymbol{g_P}, \boldsymbol{g_D}$ and $\boldsymbol{W_A},\boldsymbol{W_P}, \boldsymbol{W_D}$ are learnable weight vectors and matrices, for every node in the graph, pickup, and delivery set, respectively. Subsequently, each subset of attention is used to compute the weight value fused embedding of each node using a non-shareable weight parameter of $\boldsymbol{W}_V^l$, in Equation \ref{eqmean} and \ref{eqmean1}, in a multi-head heterogeneous attention network in each layer for each mode, respectively.

\begin{equation}\textcolor{black}{
\boldsymbol{h}_i^{l}|^{UAV}=\sum_{j \in N \cap NB^T} a^l_{i j}|^{UAV}_{A} \boldsymbol{W}_V^l \boldsymbol{h}_j^{(\mathbf{l-1})} + \sum_{j \in  P\cap NB^T} a^l_{i j}|^{UAV}_{P} \boldsymbol{W}_{Vp}^l \boldsymbol{h}_j^{(\mathbf{l-1})} + \sum_{j \in  D \cap NB^T} a^l_{i j}|^{UAV}_{D} \boldsymbol{W}_{Vd}^l \boldsymbol{h}_j^{(\mathbf{l-1})}}
\label{eqmean}
\end{equation}

\begin{equation}\textcolor{black}{
\boldsymbol{h}_i^{l}|^{ADR}=\sum_{j \in N \cap NB^S} a^l_{i j}|^{ADR}_{A} \boldsymbol{W}_V^l \boldsymbol{h}_j^{(\mathbf{l-1})} + \sum_{j \in  P\cap NB^S} a^l_{i j}|^{ADR}_{P} \boldsymbol{W}_{Vp}^l \boldsymbol{h}_j^{(\mathbf{l-1})} + \sum_{j \in  D\cap NB^S} a^l_{i j}|^{ADR}_{D} \boldsymbol{W}_{Vd}^l \boldsymbol{h}_j^{(\mathbf{l-1})}}
\label{eqmean1}
\end{equation}

Then, we calculate the attention weights $K^{\prime}$ times in a multi-head mechanism. The final weight value vector is the summation over all heads, and each weight for each layer and $k$th head is $ W_3^kh^ {\prime}_{i,k}$. The final multi-head attention is shown, followed by a BN layer in Equation \ref{eqk}. Afterwards, a feed-forward layer with a residual connection and BN layer is employed and demonstrated in Equations \ref{qek} and \ref{mmm}, respectively.

\begin{equation}
\widehat{\boldsymbol{h}}_i^{(l)}=\mathrm{BN}\left(\boldsymbol{h}_i^{(l-1)}+\sum_{k=1}^{K^{\prime}} \boldsymbol{W}_3^k \boldsymbol{h}_{i,k}^{\prime}\right)\
\label{eqk}
\end{equation}

\begin{equation}
\boldsymbol{\tilde{h}}_i^{(l)}=\operatorname{BN}\left(\widehat{\boldsymbol{h}}_i^{(l)}+\varphi\left(\widehat{\boldsymbol{h}}_i^{(1)}\right)\right)
\label{qek}
\end{equation}
\begin{equation}
     \varphi(\boldsymbol{x})=\operatorname{ReLu}\left(\boldsymbol{W}_5 \boldsymbol{x}+\boldsymbol{b}_2\right)
\label{mmm}
\end{equation}

\textcolor{black}{As a result, the node embedding in one GAT layer can be figured by the output of Equation \ref{qek}, and the next layer embedding can be found after the multi-head attention calculation in \ref{eqk}. Note that this process is done twice for each Equation of \ref{eqmean}
and \ref{eqmean1} for UAV and ADR node-edge encoding, respectively.}
Finally, after $L$ convolution layers, the output of the encoder is two distinct sets of embeddings, $\tilde{h_i}^L|^{UAV}$ and $\tilde{h_i}^L|^{ADR}$ for $i \in N$, which is used to get the final and average embedding shown in Equation \ref{eqmean2}. Note that ADRs and UAVs do not share parameters through the encoding mechanism and are embedded separately with unique network configurations. 

\begin{equation}
\boldsymbol{\bar{h}}_j|^{UAV,ADR}= \frac{1}{N}\sum_{i \in N} (\boldsymbol{\tilde{h}}_i^{L}|^{UAV,ADR})_j
\label{eqmean2}
\end{equation}

\noindent
\subsubsection{Decoder}

\textcolor{black}{
After the final node embedding for each mode, the decoder performs node assignment for each mode iteratively until all nodes are served or no feasible nodes remain.
In the decoder, first, the vehicle features are concatenated and projected through linear layers, followed by a batch normalization layer. These states include vehicle features, load, accumulated travel time and remaining battery, concatenated with the current node embedding. Equation \ref{eq:4} shows the UAV and ADR initial feature embedding into a higher hidden dimensional representation, $\mathbf{v^d}$ and $\mathbf{v^r}$, respectively.  
}

\begin{equation}\textcolor{black}{
\mathbf{v^d}= 
\operatorname{BN}\left(\boldsymbol{W}_d[v^d;\boldsymbol{\tilde{h}}|^{UAV}] + \boldsymbol{b}_d\right), \\
\mathbf{v^r}= 
\operatorname{BN}\left(\boldsymbol{W}_r[v^r;\boldsymbol{\tilde{h}}|^{ADR}] + \boldsymbol{b}_r\right)
}\label{eq:4}
\end{equation}

\textcolor{black}{
Where $\boldsymbol{W}_d$, $\boldsymbol{W}_r$, $\boldsymbol{b}_d$, and $\boldsymbol{b}_r$ represents the learnable parameters. The context embedding of each vehicle will be aggregated by dynamic states of current vehicle features with aggregated global graph embeddings, $\boldsymbol{h}_{(N)}$, to get the agent-vehicle-context embedding via linear transformations. Equation \ref{eq5} and \ref{eq51} combine per-vehicle features with global context embeddings for each mode, respectively, at each step $t$ of decoding.}

\begin{equation}\textcolor{black}{
\boldsymbol{x}_k^{(a)}|^{UAV}=\mathbf{v^k_t}+\boldsymbol{W}_{5} \cdot\left[\overline{\boldsymbol{h}}_{(N)}|^{UAV} ; \boldsymbol{v}^1_{ t} ; \boldsymbol{v}^2_{ t} ; \ldots ; \boldsymbol{v}^{N^d}_{t}\right], \quad \forall k \in N^d
 }   \label{eq5}
\end{equation}

\begin{equation}\textcolor{black}{
\boldsymbol{x}_k^{(a)}|^{ADR}=\mathbf{v^k_t}+\boldsymbol{W}_{6} \cdot\left[\overline{\boldsymbol{h}}_{(N)}|^{ADR} ; \boldsymbol{v}^1_{ t} ; \boldsymbol{v}^2_{ t} ; \ldots ; \boldsymbol{v}^{N^r}_{t}\right], \quad \forall k \in N^r
   } \label{eq51}
\end{equation}

\begin{figure}[!htbp]
\centering
     \includegraphics[width=1\textwidth]{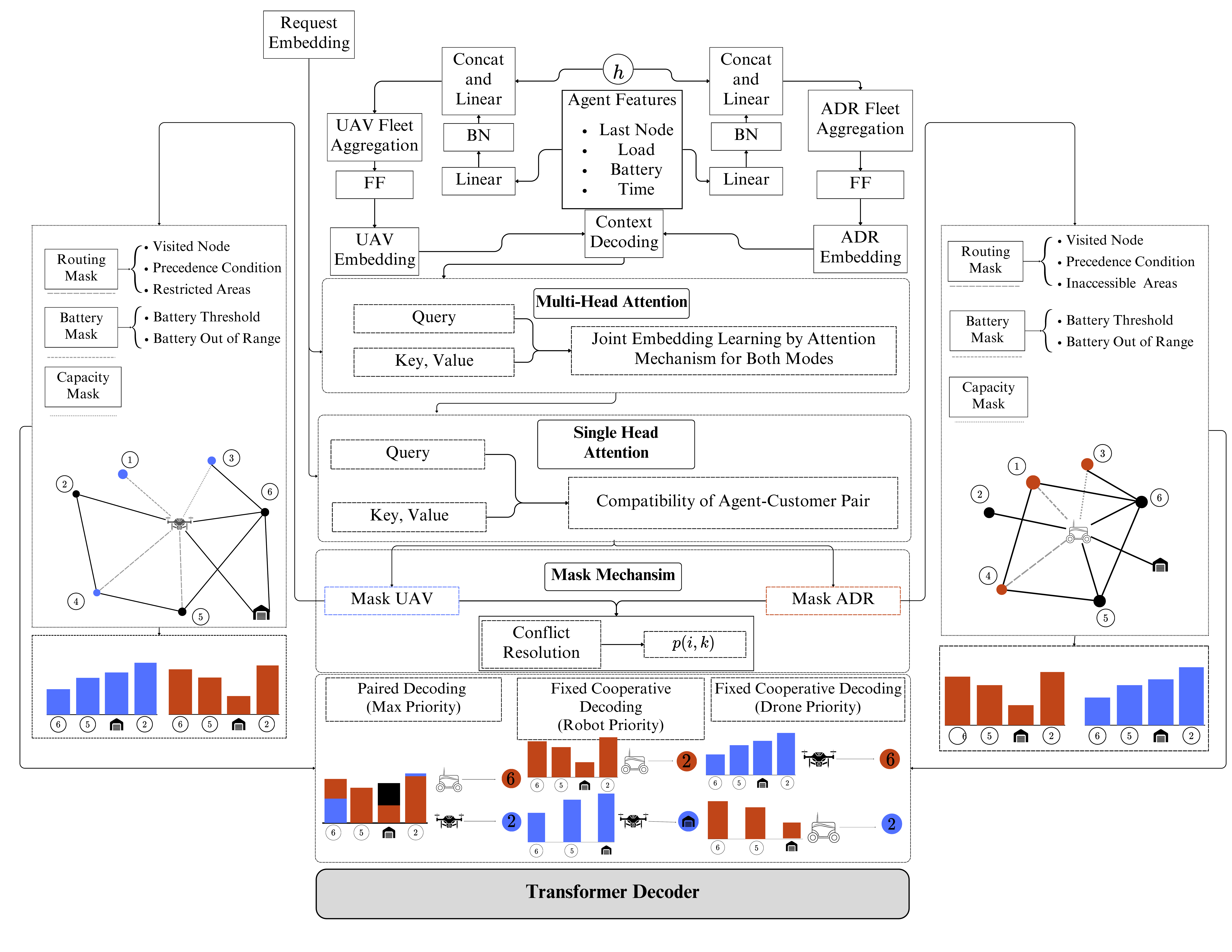}
     \caption{The decoder architecture strategies and masking scheme representation}
     \label{fig9}
 \end{figure}

We define the query vector as the concatenated agent embedding for both mode into $\boldsymbol{x}_k^{(a)}$, key vectors as the customer embedding ${\tilde{h}}_i$, and value vectors to utilize the multi-head attention mechanism to compute the compatibility scores between vehicles and nodes of $u_{k, i}$ for each customer $i$ to agent $k$ as in Equations \ref{eq1}.

\begin{equation}
    \begin{aligned}
{q}_k & =\boldsymbol{W}_{7} \cdot \boldsymbol{x}_k^{(a)}, \quad \forall k \in N^d, N^r \\
\kappa_i & =\boldsymbol{W}_{8} \cdot {\tilde{h}}_i, \quad \forall i \in N \\
{V}_i & =\boldsymbol{W}_{9} \cdot {\tilde{h}}_i, \quad \forall i \in N \\
u_{k, i} & ={q}_k^T \cdot \kappa_i, \quad \forall i \in N, \quad \forall k \in N^d, N^r
\label{eq1}
\end{aligned}
\end{equation}

 Next, we calculate the agent-customer joint information embedding as the weighted sum of value vectors in Equation \ref{eq:7}.
\begin{equation}
    h_{v,k}=\sum_{j \in N} \frac{e^{u_{k, i}}}{\sum_{j \in N} e^{u_{k, j}}} \cdot \mathcal{V}_i, \quad \forall k \in N^d, N^r.
\label{eq:7}
\end{equation}

Furthermore, the joint information embedding to a query of nodes in a single-attention mechanism is used, followed by \citep {zhang2022transformer}. It compares with each customer's key to acquire the attention coefficient, representing the compatibility between vehicle $k$ and customer $i$ at time $t$, according to Equation \ref{eq:8}. 
\begin{equation}
    \tilde{h}_{k, i}  =(\boldsymbol{W}_{10} \cdot h_{v,k})^T \cdot (\boldsymbol{W}_{11} \cdot \tilde{h}_i), \quad \forall i \in N, \quad \forall k \in N^d, N^r.
   \label{eq:8}
\end{equation}
To guarantee that each vehicle would not select the same node, a global mask is used to handle such situations and other operational and delivery constraints; the masking procedure is used for both fleets in the probability of selecting node $i$, which is noted in Equation \ref{eq7}:

\begin{equation}
 P(i,k) = softmax(C \cdot \tanh \left(\tilde{h}_{k, i}\right))
    \label{eq7}
\end{equation}

with Clip parameter of C, and $\boldsymbol{W}_{5}, \boldsymbol{W}_{6}, \boldsymbol{W}_{7}, \boldsymbol{W}_{8}, \boldsymbol{W}_{9}, \boldsymbol{W}_{10}, \boldsymbol{W}_{11} $ are learnable weight parameters. As a result, a tour can be generated by selecting vehicle-node pairs at every step. The detailed design of the decoder is depicted in Figure \ref{fig9}. In the assignment step, two strategies were adopted to test if a priority-based model in the decoding stage would perform superiorly in contrast to joint learning. The latter is when a node is assigned to the highest probability in two modes, whereas the former is for either of the modes; there is a priority in the assignment, the UAV and ADR priority case.

\subsubsection{Masking Scheme}

To ensure every node assignment to an available vehicle is feasible, a set of masking rules will be applied before the final step of the decoder. This way, only unmasked nodes and vehicles will remain in the probability of Equation \ref{eq7}. The mask rules consist of the following.

\begin{itemize}
    \item Each node is visited only once, except for charging stations. 

    \item Each vehicle must be reachable to a customer within its remaining load and battery capacity.

    \item The vehicle capacity must accommodate the request for the whole trip; it would return to depots if it cannot service any two sets of pickup and delivery nodes. 

    \item The out-of-range nodes will be masked based on the temporal and spatial adjacency neighbourhood threshold.

    \item For any pickup node, all delivery nodes, except the corresponding one, will be masked.
\end{itemize}

\begin{figure}[!ht]
\centering
     \includegraphics[width=\textwidth]{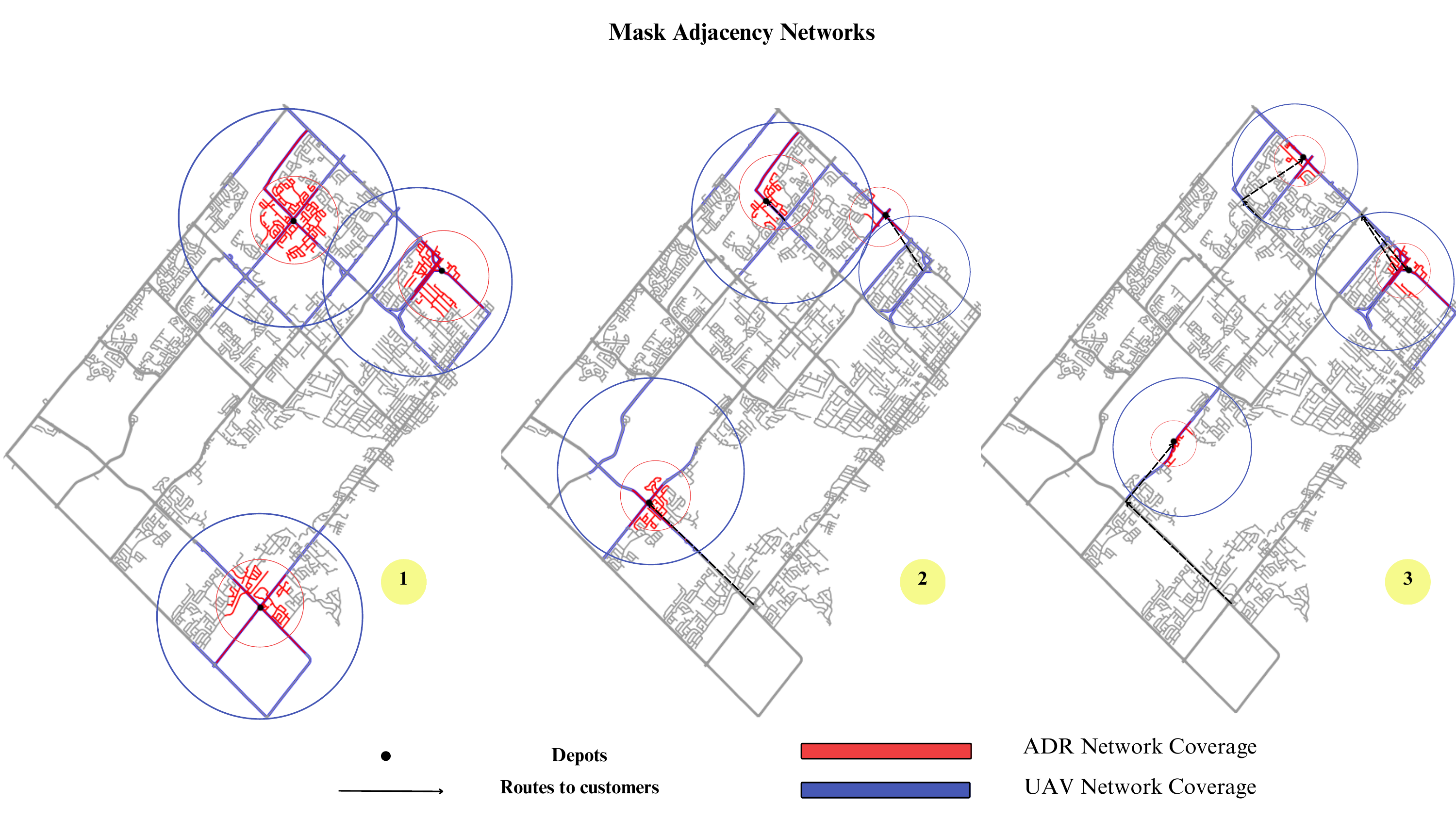}
     \caption{UAV and ADR network coverage during the delivery service}
     \label{fig1222}
\end{figure}

Figure \ref{fig1222} illustrates a visualization case of masking steps based on the dynamic features. Red and blue area networks account for the ADRs and UAVs, respectively, which, in the first step, can let each mode access more customers. In contrast, in the third step, the coverage radius is lowered due to battery constraints.  Both modes of network coverage are limited due to adjacency network restrictions on how the masking algorithm applies to accessible nodes. 
There are, in this case, three depots throughout the map where either of the modes can begin operating, and by taking each step of the routing, the network coverage where a delivery service is available varies.

Next, the problem model is trained using the policy gradient reinforcement learning method. More detail is provided in \ref{training}.

\textcolor{black}{
\subsection{Coalition Game for CE-CPDPTW}
}

In this section, we develop a coalition game for a CE-CPDPTW involving two modes of delivery, i.e., UAVs and ADRs. These agents differ in terms of speed and payload capacity, as well as battery capacity. 
\textcolor{black}{This research is a centralized model using a global conflict resolution layer to better coordinate the multi-agent learning. This implies that shared information across the entities is required. We also aim to prove and quantify the collaboration gain, if it exists, so last-mile delivery companies can be encouraged to share their resources or rent such a vehicle combination, which leads to flexible urban coverage and reduces operational costs in a cooperative manner. To do so, first, based on the coalition game theory, the existence of such cooperation must be verified. Next, the advantages of cooperative delivery can be assessed by a cost allocation mechanism, which can fairly share the cost of forming coalitions.}

\textcolor{black}{ We consider a set of UAV agents in a coalition $ S_D = \{D_1, D_2, \ldots, D_m\}$ and similarly for ADR $S_R = \{R_1, R_2, \ldots, R_n\} $ consisting of $m$ UAVs and $n$ ADRs. These vehicles can jointly cooperate in smaller groups of $S_1$ and $S_2$, form a sub-coalition in such a way that $(S_1 \subset S_D) \cup (S_2 \subset S_R)$. 
Consequently, according to section \ref{S:t22}, in this case, if the characteristic function of the game satisfies super-additivity \ref{sup}, followed by the existence of a core in the grand coalition, it is always profitable for the two modes to cooperate \citep{osicka2020cooperative}.  
This case can be held if one of the efficiency, Equation \ref{eff} or coalitional rationality, Equation \ref{rat}, is found with a non-feasible allocation. 
}

\textcolor{black}{
Generally, proving such statements for our problem is not a trivial task, given the highly non-linear and data-driven bi-optimization model. As an alternative, we use the RL solution as a characteristic function, which leverages a reinforcement learning trained model of CE-CPDPTW across different numbers of agents, accounting for different coalitions.}
In this regard, after training the RL model for multi-agent, we use the solution to the test instances on the generalized model to obtain the characteristic function among different coalitions of ADRs and UAVs, to check super-additivity and if the core exists. Note that the core can sometimes be empty, meaning cooperation is not preferred. 
Next, we show why it is possible to verify these properties using the RL solution.

\begin{algorithm}[ht]
\footnotesize
\caption{Core Coalition for CE-CPDPTW with UAVs and ADRs}
\label{core_check}

\begin{algorithmic}[1]
\State \textbf{Initialization:} Trained RL actor model for ten agents, $C(K) \simeq C^{opt}(D,R) $, \; $\forall$  $D \subseteq {N^d}$ and $R\subseteq {N^r}$ for the coalition

\State \textbf{Input:}  Initialize $D$ and $R$ for generalized trained model $C(S) = C^{opt}(D,R)$ for any coalition $S \subseteq {N^d \cup N^r}$
    
    \State Define efficiency condition $C(K)$ for grand coalition $K$ (all agents) such that: $\sum_{i \in N} C(S_i) = C(K)$

    \State Define coalition rationality condition: For any subset $S \subseteq K$: $ \sum_{i \in S} C(S_i) \leq C(S)$
       
    \Statex
    \Statex \textbf{Step 1: Solve for Cost of Individual Coalitions}
    \State Compute cost $C(D)$ for $D$ UAVs of CE-CPDPTW
    \State Compute cost $C(R)$ for $R$ ADRs of CE-CPDPTW
    \State Compute cost $C(D+R)$ for $D$ UAVs and $R$ ADRs working together of CE-CPDPTW

    \Statex
    \Statex \textbf{Step 2: Check Sub-additivity Condition}
    
    \State \textbf{if:} {$C(D+R) \leq C(D) + C(R)$}
    \State  \;\;\;\; The game is sub-additive (cooperation reduces cost)
    \State \textbf{else:}
    \State \;\;\;\; The game is not sub-additive
    \State \textbf{end}

    \Statex
    \Statex \textbf{Step 3: Efficiency and Coalition Rationality Conditions}
    \State Set $C(S_i)$ as the cost share allocated to each agent in $D$ UAVs and $R$ ADRs
    \State \textbf{for} {each coalition $S \subseteq K$:}
        \State \;\;\;\;\textbf{if:} {$\sum_{i \in S} C(S_i) > C(S)$ for any coalition $S$:}
           \State \;\;\;\; The core is empty 
    \State \;\;\;\;\textbf{elif:} {$\sum_{i \in N} C(S_i) = C(K)$ is satisfied for all $C(S_i)$}
            \State \;\;\;\;\ The core is non-empty 
        \State \;\;\;\; \textbf{end}
    \State \textbf{end}

\end{algorithmic}
\end{algorithm}

\textcolor{black}{We prove this by contradiction, as if we assume the super-additive game is satisfied, and we then find the condition that meets Equation \ref{sup}. Since the RL provides the upper bound solution to the problem, it is obvious that the solution to the CE-CPDPTW for a coalition $S$ is equal to or more than its true optimal value, $C^{opt}(S)$; as a result, Equation \ref{co} must be held.}
\begin{equation}
C^{opt}(S) \leq C^R(S)
\label{co}    
\end{equation}

\textcolor{black}{Where $C^R(S)$ is the value for cost as the solution from the reinforcement learning problem. To solve the super-additive game for two modes of coalitions $S_1 \subseteq K$ and $S_2 \subseteq K$, Equation \ref{sup} must hold for both $S_1 \cup S_2$, as shown in Equation \ref{co1} }

\begin{equation}
C^{o p t}(S_1 \cup S_2) \leq C^{o p t}(S1)+C^{o p t}(S_2)
    \label{co1}
\end{equation}

\textcolor{black}{If by the assumption, this statement is true, afterwards, using Equation \ref{co} for each coalition, $C^{opt}(S_1) \leq C^R(S_1)$ and $C^{opt}(S_2) \leq C^R(S_2)$, and substituting in Equation \ref{co1}, the sub-additive property of the cooperative coalition will hold as Equation \ref{co2}}
\begin{equation}
C^{o p t}(S_1 \cup S_2) \leq C^R(S_1)+C^R(S_2)     , S_1 \subseteq K, S_2 \subseteq K, S_1 \cap S_2=\emptyset
\label{co2}
\end{equation}

\textcolor{black}{
Finally, if for any coalition Equation \ref{co2} exists, we can conclude that the game is super-additive, thus our assumption is proved, and the core can exist, conditioning on Equations \ref{eff} and \ref{rat}; thereby, cooperation is stable and fair, followed by Shapley value allocation to equal distribution of the marginal profit. The workflow of this coalition game solution is given in Algorithm \ref{core_check}.
}

\section{Results}
\label{S:t5}
This section conducts extensive experiments on
synthetic and real-world datasets, with different request sizes, vehicles, and depots. All simulations are carried out with PyTorch on an NVidia A100 GPU.

\subsection{Experimental Setup}
Node locations are drawn uniformly from the square area of $[0,5]\;km$, and the pickup time window of nodes is driven by a Poisson distribution for the evening peak hour, in addition to a random delivery time window from $[30,60]\;min$ based on the node. The maximum speed of the vehicles is set at $20 \;m/s$ and $8.3\; m/s$ for UAVs and ADRs, respectively, though they will adjust their speed based on the horizon time window to arrive at the destination on time. The demand volume of a pickup node $d_i$ is uniformly sampled from (1,10), the capacity limit of each vehicle is 5 and 10, and the maximum battery energy for each mode is set as $6.5 \;KJ$ and $4.5\;KJ$ for UAV and ADR, respectively. The vertical movement for takeoff and landing is considered 2 minutes for UAVs, and the recharging time at each 
charging station for UAV and ADR is up to 10 and 20 minutes, respectively. Also, the speed of either of the vehicles visiting the depot for recharging is half of the maximum speed. The lower limit threshold for the battery is $30\%$ and $20\%$ for UAVs and ADRs, respectively, since they can manage energy while going back to depots. 
We assume a higher technology readiness level in some parameters, such as battery recharging efficiency. The penalty monetary coefficients for utilizing the UAV and ADR are set separately as $\alpha_1 = 0.6 \; \frac{\$}{min}$ and $\alpha_2 =0.1 \;\frac{\$}{min}$ monetary units per minute. According to \cite{sudbury2016cost} and \cite{MarketResearchReport}, these factors are considered the cost of automated vehicles' utilization for delivery companies. We assume that these vehicles are rented for the peak hours of operation. The pickup time window is set as a soft constraint, and if the vehicle arrives \textcolor{black}{later than the pickup time window, it will be penalized for customer extra waiting time since the order has been prepared to be picked up}; the cost penalty rate will be ($\alpha_3 = 0.05 \;\frac{\$}{min})$. On the other hand, the delivery time window is designed flexibly to encourage vehicles to arrive as soon as possible. If it arrives earlier than the time window, no penalty is incurred. Nonetheless, the penalty factor for time delay is $\alpha_4 = 0.05\; \frac{\$}{min}$, the monetary unit per minute delay. More specifically, the delay penalty in case of arriving after the time window (pickup and delivery node) is $0.05$, as a compensation for the customers' time, even though the delay penalty for arriving before the pickup time window is $0.01$ in case of \textcolor{black}{arriving early since the order is not ready yet to be picked up and also vehicles are idle and not operational, therefore, the the incurring cost is not directly contribute to customer waiting time and is considered a lower value in reward signal to cost function.} Furthermore, we set a negative penalty value in the cost function if the battery constraint is violated and falls below the threshold. This factor is considered if there would be another delivery for that unfinished delivery customer; thereby, the battery penalty factor is $\lambda = 1$.

We also present a detailed comparison study between the proposed methodology with state-of-the-art methods, including Google OR Tools and available state-of-the-art methods, using transform attention models, described as follows.

\begin{enumerate}
    \item \textbf{Google OR-Tools}, a commonly used software for solving vehicle routing problems. For a Python version, we implement the algorithm to solve the CE-CPDPTW, which models our problem constraints with a limit of 3600 seconds.
    
    \item \textbf{Gurobi Optimizer}, a mathematical optimization software for solving mixed-integer linear and quadratic
    optimization problems. We adopt our mixed integer programming in the mathematical formulation for CE-CPDPTW, and solve the problem with the upper limit of solution time as 3,600 seconds.

    \item \textbf{Attention model (AM) } \cite{kool2018attention}, a multi-head attention transformer structure, utilizes both the encoder and decoder for vehicle routing. \textcolor{black}{Instead of edge-enhanced graph attention, we manually use conventional multi-head attention and customize it for the encoder in our setting, combined with our masking scheme in the decoder, to compare the encoding power in the urban delivery application.}

    \item \textbf{Heterogeneous AM(HetAM) } \cite{li2021heterogeneous}, a heterogeneous attention model for the pickup and delivery problem, in which overall seven types of attention layers were designed to consider different roles played by nodes while considering the precedence constraint. Their architecture has been incorporated in the encoder, augmented with our masking scheme, likewise the previous method.

\end{enumerate}

\subsection{Policy Network Training }

We have conducted several experiments to evaluate our model's performance and justification fairly and effectively. Initially, to verify the training performance with respect to both datasets and neural network parameter trade-offs, we tested experiments with three scales, 2N = 20, 50, 120, and with the number of agents  2, 4, 6, and 10 vehicles. For convenience, we will use the problem name in the following manner. For example, ``4CE-CPDPTW20-d2'' indicates the delivery problem of 4 vehicles, and 20 requests with 2 depot stations. The networks are trained via Adam optimizer with four layers of convolution and eight heads, the node and edge hidden embedding dimension as $128$ and $16$ respectively, and learning rate $l_r = 0.00005$.
We apply greedy decoding for the baseline to reduce the reward variance in the back-propagation phase and evaluate the actor parameters in the training process with sampling decoding.
The training is done for the cost of 100 epochs, and in each epoch, 2500 batches with 512. The training times for the following cases, 20-1d one vehicle each, 50-3d two vehicles each, and 120-3d three vehicles each, are 21 min, 55 min, and 112 min per epoch, respectively.

\textcolor{black}{Secondly, to show the applicability and power of the proposed model in urban delivery scenarios subject to high-density areas and heterogeneous demand distribution in terms of location and time window, the model is trained, influenced by density and adjacency threshold parameters. Each parameter is randomly chosen from a specific bound, which allows a variety of instances, and the distribution of temporal-spatial with a varying weighted connection can be absorbed. The bounds are considered uniformly as $\zeta \in [60, 80]\;min$, $\mu \in [1, 3]\; km$, $\rho \in [0.4, 0.7]$, in a limited difference since the model struggles to smoothly converge due to high variance in rewards}. Moreover, the dataset carries the stochastic wind model by \cite{johnson1985wind}. It substantially impacts the cost function since it can work either in favour of the UAVs or adversely impact when the direction is on the opposite side of the UAV's speed. Besides, UAV capacity is lower than that of ADRs and carrying heavier payloads will use more power for the same delivery; thereby, training on a variety of datasets and their influence on the cost function, we expect that heavier payload in short distances will be mainly assigned to ADRs and UAVs deliver the far requests, as well the requests which fall in the unfavorable-wind locations for UAVs, will be done by ADRs. However, to avoid over-fitting and trapping in the sub-optimal space, we consider two main wind directions, eastward and westward, to perturb the UAV power consumption model. Otherwise, the reward distribution would be noisy and convergence under such criteria is not guaranteed. Also, we allow the agents to reach a negative battery level and instead add a term to the cost function for a penalty. Otherwise, the action space would be limited, and exploring the optimal solution through the learning process would be difficult.

\begin{figure}[!ht]
\centering
     \includegraphics[width=0.8\textwidth]{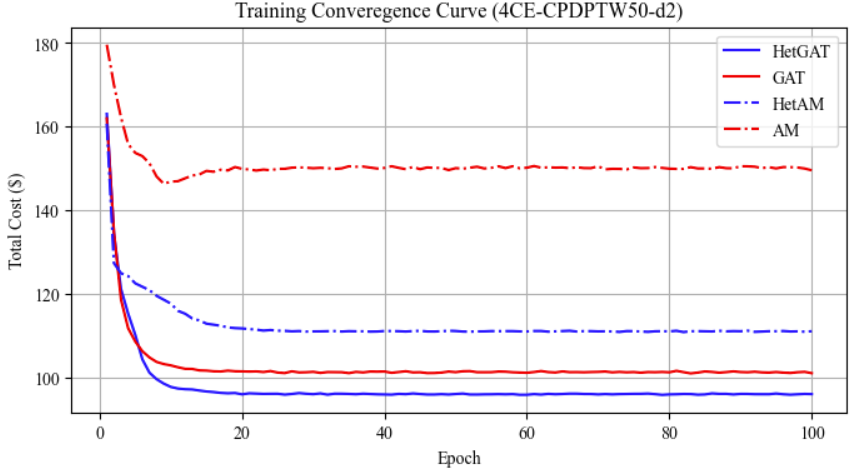}
     \caption{Training convergence curve in 100 epochs for comparing encoder design}
     \label{train2}
\end{figure}

\textcolor{black}{
To test the efficacy of the model design, specifically spatial and temporal edge-enhancement and learning curve performance, the baseline test cases are considered for training comparison. Figure \ref{train2} demonstrates the training performance, the average cumulative total cost \textcolor{black}{with monetary unit of dollar currency}, versus the number of epochs. The training is done for our model, called for short ``HetGat", the adjacency mask for the graph attention encoder with problem feature-aware edge enhancement, focuses its attention on only the meaningful neighbours. Whereas, the ``GAT" model uses the full edge connection to graph attention with distance-based edge weight. As this Figure shows, the multi-head attention models fall short in finding a better solution, and despite the smooth curve, their convergence rate is lower than edge-based models. The performance of our model and the regular graph attention model is competitive; however, the model aggregates the distance-priority nodes and does not incorporate the time-sensitive delivery information in the encoder; therefore, it fails to capture the time priority over the relationship between the nodes. Note that in all models, the problem features are kept identical in the encoding process as well as pickup and delivery node embedding, and only the central attention mechanism has been replaced. The decoder, on the other hand, is the same since all use transformers.}

\textcolor{black}{
In addition, the training parameters and running time for each model are represented in Table \ref{compl}. The complexity of the proposed model is moderately higher than other baselines, except HetAm, since it uses seven types of attention, implying more parameters and higher computational cost. However, this model can learn the interaction of a multi-modal fleet of a UAV and ADR specification for heterogeneous spatial-temporal data distribution. 
}

\begin{table}[ht]
\scriptsize
\centering\textcolor{black}{
\caption{Training parameters of encoder and runtime per epoch for 2CE-CPDPTW20-1d}
\begin{tabular}{lcccc}
\hline & AM & HetAM & GAT & HetGat \\
\hline Trainable parameters &$594,816$& $1,109,504$ & $413,312$ &$716,640$  \\
Running time & $18 \mathrm{~min}$ & $26 \mathrm{~min}$ & $16 \mathrm{~min}$ & $21 \mathrm{~min}$ \\
\hline \hline
\end{tabular}
\label{compl}}
\end{table}

\begin{figure}[ht]
\centering
     \includegraphics[width=0.8\textwidth]{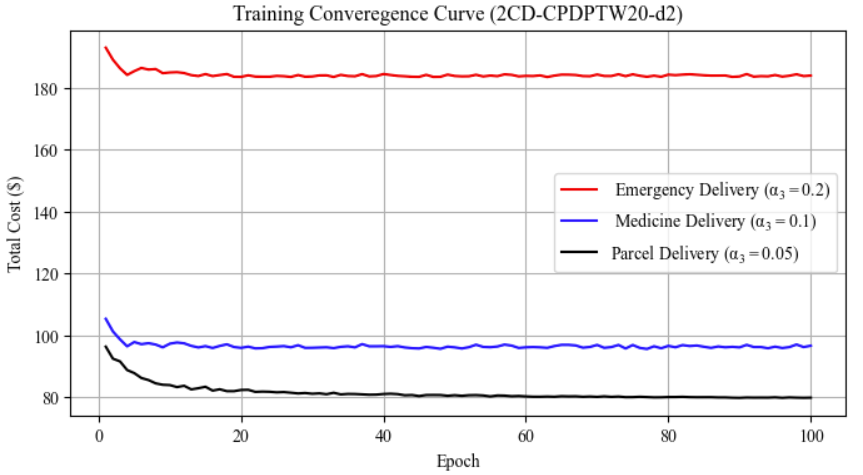}
     \caption{The learning curve for the various delivery cases from low to high critical}
     \label{fig231}
\end{figure}

Moreover, rigorous training criteria have been set to consolidate the desired level of service in time-critical delivery applications when the time delay is with more linear penalties accounted for different delivery cases. It has been assumed that the delay penalty is highest in the case of an emergency when customers are patients in the medical centers awaiting an organ for surgery or transplant. Another case is assumed to be of priority but not as critical as the first one, medicine delivery for hospitals or customers. The penalty for obtaining this learning curve is set as $\alpha_3 = 0.1$, and, for the former, is set as a higher rate, $\alpha_3 = 0.2$, which reinforces the time-sensitive delivery. In addition, the rate of delay penalty when vehicles arrive earlier than the pickup time window is set as \textcolor{black}{ $0.05$ to make the reward signal higher for not arriving early, yet not large since it will fall into the after pickup time window, which is more costly}. Finally, the baseline training curve for a guaranteed service level is considered a typical parcel or meal delivery, though with a stricter penalty on battery violation and increasing the battery threshold by 20\%, ensuring a vehicle would not fail during the delivery. The output of these scenarios is depicted in Figure \ref{fig231}. According to the Figure, emergency and medicine delivery both result in a higher overall cost and drop drastically from the initial point where the weights of the training parameters are randomly generated, leading to the initial cost being high, and after a few epochs, they can manage to converge; however, it cannot improve the policy much further due to higher prioritization to delay penalty and remain in the local minima. This behaviour can be explained by agents increasing travel costs to ensure no battery and time delay violation happens. On the other hand, the parcel delivery case does not give much importance to the time delay, leading to the model learning to find a balance of battery usage with delay penalties to avoid delivery failures and the training curve can converge and perform more stably than other cases.

\subsection{Results Analysis}

The comparison results of the overall performance are shown in Table \ref{res}. The total cost in monetary units of US dollars, the gap of each method, and the CPU time spent during inference with three different customer scales are reported, and the optimal solution is considered the minimum cost function. Accordingly, gap denotes the gap between a result solution and the optimal baseline solution, written in Equation \ref{rqb}. Besides sampling decoding, another test is denoted as E-CPDPTW(1280), where the decoder samples 1280 times at each step to obtain multiple solutions and selects the solution with the highest reward as the routing solution.

\begin{equation}\textcolor{black}{
\text { Gap }=\frac{O b j_{\text {best }}-O b j}{O b j_{\text {best }}} \times 100 \%
  }  \label{rqb}
\end{equation}

\begin{table}[!ht]
\footnotesize
\centering
\caption{Costs and running computing time for CE-CPDPTW}
\resizebox{\textwidth}{!}{
\begin{tabular}{@{}llccccccccccccc@{}}
\toprule
Vehicles & \textbf{Method} & \multicolumn{3}{c}{\textbf{E-CPDPTW10-d1}} & \multicolumn{3}{c}{\textbf{CE-CPDPTW20-d1}} & \multicolumn{3}{c}{\textbf{CE-CPDPTW50-d2}} & \multicolumn{3}{c}{\textbf{CE-CPDPTW120-d3}} \\
 &  & \textbf{Cost (\$)} & \textbf{Gap, \%} & \textbf{CPU Time (s)} & \textbf{Cost (\$)} & \textbf{Gap, \%} & \textbf{CPU Time (s)} & \textbf{Cost (\$)} & \textbf{Gap, \%} & \textbf{CPU Time (s)} & \textbf{Cost (\$)} & \textbf{Gap, \%} & \textbf{CPU Time (s)} \\ \midrule
\multirow{6}{*}{2} 
 & Gurobi & 47.25 & 0.00 & 6.87  & 81.15  & 9.92 & 3600 & -  & - & - & -  & - & - \\
 & OR-Tools & 47.25 & 0.00 & 7.87 & 89.55  & 21.28 & 3600 & -  & - & - & -  & - & - \\
 & Heterogeneous AM & 56.45 & 19.48 & 0.44 & 76.87  & 4.12 & 0.33 & 177.11  & 3.88 & 0.73 & 492.42  & 7.52 & 1.38 \\
 & AM (greedy) & 55.53 & 14.91 & 0.27 & 82.02  & 11.09 & 0.21 & 180.55  & 5.88 & 0.80 & 506.64  & 10.68 & 1.52 \\
 & CE-CPDPTW (greedy) & 50.50 & 6.88 & 0.57 & 75.01  & 1.60 & 0.58 & 175.58  & 2.94 & 1.32 & 460.21  & 0.53 & 2.45 \\ 
  & CE-CPDPTW (1280) & 48.79 & 3.26 & 0.65 & 73.83  & 0.00 & 0.66 & 170.54  & 0.00 & 1.35 & 457.77  & 0.00 & 3.17 \\ \midrule
\multirow{4}{*}{4} & Gurobi & 39.23 & 0.00 & 3.45  & 57.04  & 19.05 & 3600 & -  & - & - & -  & - & - \\
& OR-Tools & 39.23 & 0.00 & 1.68 & 57.04  & 19.05 & 3600 & -  & - & - & -  & - & - \\
 & Heterogeneous AM & 45.23 & 15.30 & 1.21 & 49.19  & 2.67 & 0.42 & 102.20  & 8.06 & 0.83 & 306.83  & 14.63 & 1.47 \\
 & AM (greedy) & 48.88 & 24.60 & 0.74 & 51.23  & 6.93 & 0.59 & 108.13  & 14.34 & 0.78 & 313.29  & 17.05 & 1.13 \\
 & CE-CPDPTW (greedy) & 41.33 & 5.35 & 0.49 & 50.68  & 5.78 & 0.64 & 98.81 & 4.54 & 0.98 & 283.35 & 5.86 & 2.75 \\ 
   & CE-CPDPTW (1280) & 40.13 & 2.29 & 0.77 & 47.91  & 0.00 & 0.75 & 94.54  & 0.00 & 1.66 & 267.67  & 0.00 & 3.16 \\ \midrule
\multirow{4}{*}{6} & OR-Tools & - & - & - & -  & - & - & -  & - & - & -  & - & - \\
 & Heterogeneous AM & - & - & - & 40.85  & 4.92 & 0.62 & 78.14  & 20.18 & 0.81 & 165.12  & 9.79 & 1.96 \\
 & AM (greedy) & - & - & - & 41.02  & 5.92 & 0.53 & 82.11  & 20.72 & 0.96 & 184.66  & 16.09 & 1.72 \\
 & CE-CPDPTW (greedy) & - & - & - & 40.60  & 4.91 & 0.87 & 68.92  & 5.98 & 1.84 & 152.66  & 0.80 & 2.80 \\
   & CE-CPDPTW (1280) & - & - & - & 38.59  & 0.00 & 1.15 & 65.02  & 0.00 & 2.42 & 151.56  & 0.00 & 3.43 \\ \bottomrule
\end{tabular}
\label{res}}
\end{table}

Regarding solution quality, our model outperforms classical methods and, in most cases, transformer-based methods. The computation time increases exponentially as the problem scale increases for the exact 
solvers of Gurobi and OR Tools, which fail to solve instances with $>50$ customers within an acceptable time, as well as failure to address constraints to find the optimal solution, which cannot complete the delivery request set. \textcolor{black}{Additionally, the classical solvers struggled to find the optimal solution for the network size of 20 in the time limit due to the model complexity. It could only achieve the optimal solution for a small size of 10. The CPU running time shows a inference time which as the agents and problme size scales, it becomes higher showing the almost linear incremental rate.} Our model, on the other hand, incorporates operational constraints, which add layers of complexity to the routing problem, and the learning process is accounted for by two different network setups, which again impose more parameters to train, especially for discontinuous reward functions, and lead to unstable learning signals. Considering the spatial-temporal information, our transformer performs significantly better due to the novel exclusive encoder-decoder design to distinguish between each mode assignment and cooperation tasks over other benchmarks. Moreover, it can be observed that the quality of the solution outperforms with a larger gap as the network size becomes larger compared to baselines. Moreover, this model is designed to capture the hard and soft constraints and the variety of heterogeneity of different datasets. Therefore, high representation and embedding dimensions and training parameters must be defined to handle such restrictions and capture unpredictable and irregular patterns of on-demand delivery. Nevertheless, considering addressing such operational constraints to this extent, it can still compete with most modern algorithms at the same scale in comparable times.

\subsection{Robustness Test}

The framework for a multi-modal delivery system has been designed to address issues with the delivery system in urban environments. However, it is appealing to incorporate issues from customer perspectives to establish a robust performance. Therefore, it is crucial that our model produces reasonable results for cases that come with uncertainty or undergo more realistic applications, leading to a sustainable level of service. 

\subsubsection{Decoding Scenarios}

The strategy of decoding has been examined to determine if the service level would differ based on the paired and fixed cooperative priority-based assignment of the modes to the customers. 

\begin{figure}[!ht]
\centering
     \includegraphics[width=0.8\textwidth]{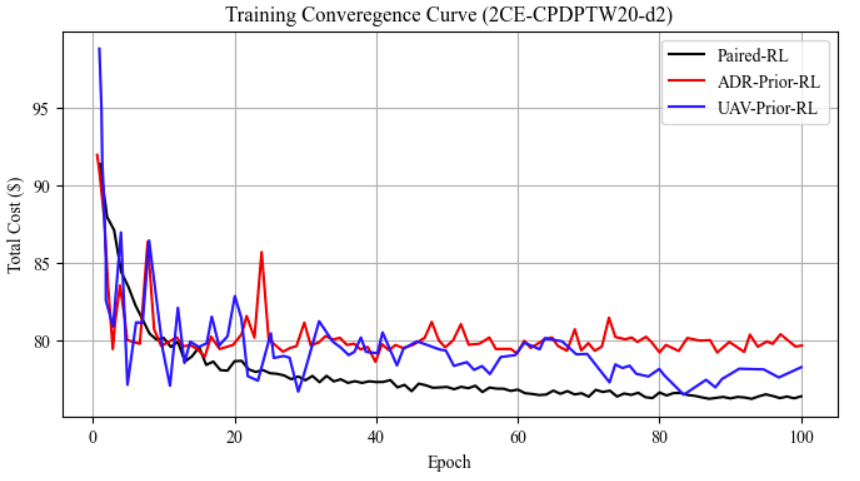}
     \caption{Training convergence curve in 100 epochs for prior-based decoding}
     \label{train}
\end{figure}

The training curve for our proposed learning architecture decoding with 20 requests is shown in Figure \ref{train}. The average cumulative total cost \textcolor{black}{with monetary unit of dollar currency}, versus the number of epochs, is depicted. Note that three different assignment methods have been shown: decoding, where both modes are accounted for as a pair to pick the highest probability of node for all the agents, and two other decoding assignments, where one of the modes comes into priority. By priority, we explicitly mean picking the highest node assignment probability for any mode that comes first. The former is called Paired-RL, and the latter is called UAV-Prior-RL and ADR-Prior-RL, which separately give the fixed assignment order to one mode. The paired-learning decoding would encompass a broader solution space to be explored, yet required for the multi-agent policy network training curve to be smoother and less noisy. Otherwise, limiting the decisions for either of the modes might cause agents to receive a high variance in reward. In this regard, we add an entropy regularization term with decaying weight to the loss function to reduce over-fitting.  

It can be observed by the training convergence curves that the Paired decoding learning shows a stable average cost over time and quickly converges to a 12\% cost gap at epoch 40. This is due to the model's ability to allocate tasks based on the optimal agent for each request, leading to optimal choice over the training process. UAV-Prior, however, shows efficient behaviour by decreasing the cost significantly at earlier epochs, being able to adapt to the environment. Additionally, it can converge to a Paired case, though the curve is under fluctuations, possibly due to the UAV's sensitivity to specific conditions like different edge embedding and weather in every batch, which can impact its performance consistency as well as exploring the sub-optimal space. On the other hand, ADR-Prior shows less variance domain throughout the training with slower convergence and higher average reward. This can be explained by the fact that, whereas in the UAV-Prior, this decoding struggles to match the efficiency of UAVs in certain scenarios, such as longer distances or time-sensitive demands. Therefore, it can limit the exploration space. In the end, given the performance quality and stability, Paired learning acquires a better solution.

\subsubsection{Uncertain Scenarios}

First, to ensure the adaptability of the proposed model to real-world scenarios, a few layers of uncertainty will be added to both the spatial and temporal aspects of delivery requests, such as non-uniform customer locations, tighter time windows, and directed wind conditions. Second, by enhancing the functionality of this study to a broader application, the practical implications are expanded to critical cases of deliveries, such as medicine deliveries, where there is an urge for the parcels to be delivered on time. 

The application of our trained models in real-world scenarios is
described, where depot and delivery locations are simulated within real-world road networks retrieved from OpenStreetMap for the city of Mississauga.
In this regard, these scenarios will verify how the multi-agent reinforcement learning model's robustness performs in different wind situations and spatial-temporal patterns of order arrival. This will facilitate comprehending various circumstances that affect the model's capacity to produce effective routing options. We make different scenarios where the test dataset is not uniformly distributed in the case study; the pickup and delivery locations are distributed based on a non-uniform distribution for scattered places, as well as a case of a tightened time window.  Furthermore, the wind has a constant direction over the case study, such as Eastward and Westward, to evaluate how it would affect each mode's assignment pattern, given that UAVs can either benefit from the wind direction or be confined due to more battery usage. The wind speed magnitude can be up to $12 m/s$.

\begin{table}[ht]
\footnotesize
\centering
\resizebox{\textwidth}{!}{
\caption{Costs and computing time for CE-CPDPTW with non-uniform dataset}
\begin{tabular}{@{}llcccccccccc@{}}
\toprule
Case & \textbf{Method} & \multicolumn{3}{c}{\textbf{E-CPDPTW20-d1}} & \multicolumn{3}{c}{\textbf{CE-CPDPTW50-d1}} & \multicolumn{3}{c}{\textbf{CE-CPDPTW120-d2}} \\
 &  & \textbf{Cost (\$)} & \textbf{Gap, \%} & \textbf{CPU Time (s)} & \textbf{Cost (\$)} & \textbf{Gap, \%} & \textbf{CPU Time (s)} & \textbf{Cost (\$)} & \textbf{Gap, \%} & \textbf{CPU Time (s)} \\ \midrule
\multirow{4}{*}{Eastward wind} 
 & Heterogeneous AM  & 38.51  & 15.10 & 0.74 & 76.51  & 12.67 & 0.84 & 170.68  & 11.55 & 1.76 \\
 & AM (greedy) & 39.92  & 19.34 & 0.71 & 78.24  & 15.20 & 0.73 & 192.23  & 25.58 & 1.82 \\
 & CE-CPDPTW (greedy) & 35.80  & 6.99 & 0.93 & 69.57  & 2.43 & 1.76 & 158.41  & 3.52 & 1.63 \\
 & CE-CPDPTW (1280) & 33.46  & 0.00 & 1.02 & 67.92  & 0.00 & 1.92 & 153.03  & 0.00 & 3.42 \\ \midrule
\multirow{4}{*}{Westward wind} 
 & Heterogeneous AM  & 40.31  & 22.89 & 0.77 & 77.97  & 17.16 & 0.81 & 168.31  & 11.96 & 1.52 \\
 & AM (greedy) & 41.74  & 27.33 & 0.70 & 75.04  & 12.75 & 0.83 & 185.60  & 23.43 & 1.74 \\
 & CE-CPDPTW (greedy) & 34.11  & 4.02 & 0.88 & 69.33  & 4.16 & 1.48 & 155.09  & 3.16 & 1.66 \\
 & CE-CPDPTW (1280) & 32.79  & 0.00 & 1.13 & 66.56  & 0.00 & 1.73 & 150.35  & 0.00 & 2.89 \\ \midrule
\multirow{4}{*}{Tight time window} 
 & Heterogeneous AM  & 43.16  & 11.20 & 0.80 & 82.12  & 8.26 & 0.95 & 207.06  & 24.20 & 1.68 \\
 & AM (greedy) & 44.43  & 14.43 & 0.75 & 81.90  & 8.00 & 0.70 & 216.25  & 29.78 & 2.15 \\
 & CE-CPDPTW (greedy) & 42.55  & 9.61 & 0.94 & 77.81  & 2.59 & 1.88 & 168.22  & 1.17 & 1.85 \\
 & CE-CPDPTW (1280) & 38.82  & 0.00 & 1.21 & 75.85  & 0.00 & 1.91 & 166.63  & 0.00 & 3.71 \\ \bottomrule
\end{tabular}
\label{res1}}
\end{table}

Like the previous analysis, Table \ref{res1} demonstrates a performance for different scenarios, their cost, gap, and computation time for a fleet of three vehicles for each mode. It is noted that the analysis is conducted only for transformer-based methods since the traditional approaches were not superior in the last section. The results show that heterogeneous attention mechanisms can handle these variations better due to their inherent capabilities in managing data inputs and variable constraints, rather than the original transformer model. Moreover, our model executes in a comparable running time, though it is considerably more robust than the regular situation. This led to the advantage of our model over the other models in terms of model specification and the power of incorporation of the graph attention network in capturing inter-nodal information, as well as how our novel approach can produce high-quality solutions. The cost value, however, does not always increase in the case of uncertainty. For example, in the case of eastward wind and $n=20$, the cost is reduced, and for $n=50$, it is increased to some point. The reason for such behaviour is that the wind may influence the ADR adversely, leading to higher battery consumption. Also, it can take place as it assists the UAV in moving downstream airflow, which is in the same direction. Both windy cases suggested that the approach finds optimal solutions as it conducts graph data processing much more efficiently, yet with more computation effort. Additionally, the tight time-window case is when the difference between pickups and deliveries is up to 25 minutes (about $20\%$ tightened), showing the variation of the solution under such modification. It can be seen that presumably, for shorter time windows, the model will not be effective, and also, the result alteration is more significant when it comes to smaller networks, most probably due to the lack of sufficiently large customer nodes, so it can counteract wind influence and time window distribution.  

\begin{figure}[!ht]
\centering
     \includegraphics[width=\textwidth]{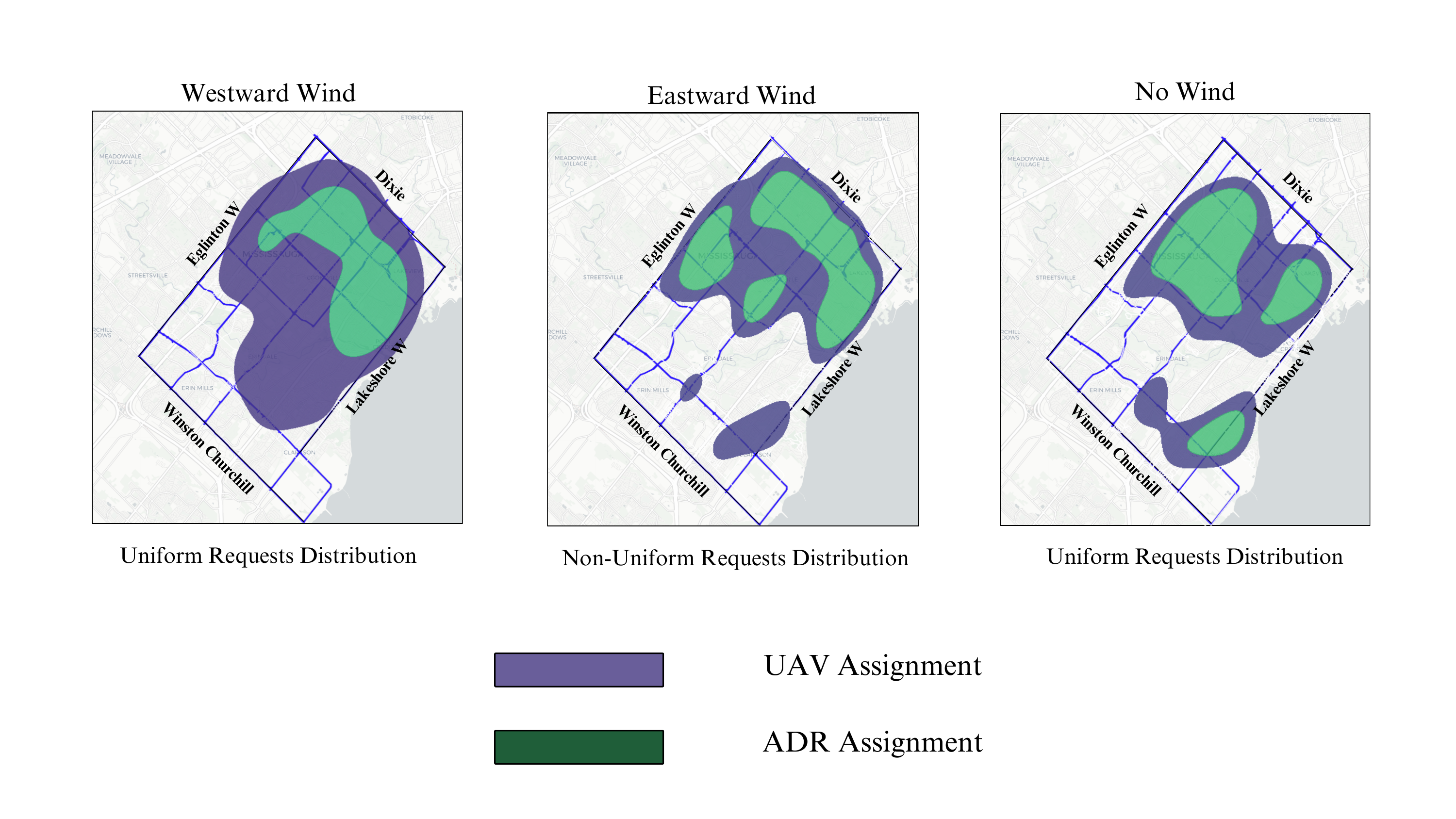}
     \caption{Multi-modal assignment distribution in the existence of uncertain conditions}
     \label{aa}
\end{figure}

Furthermore, to draw the patterns of spatial-temporal requests and how the assignment of multi-modal delivery in the presence of wind can be done, for each case, the assignment map for both modes is represented in Figure \ref{aa}. The deviation from the no-wind case is noticeable, where UAVs tend to take the windward delivery action. 
For the tight time window case, the battery limit constraint has been treated less strictly due to some cases that might have appeared with long distances from the depot, making it less efficient for the electric delivery. Both modes can mainly do one delivery at a time in such situations. In contrast, with a uniform test set, they can conduct multiple deliveries and even visit the depot during the travel. However, the system's overall performance is comparably acceptable due to changing conditions in real-time and assigning UAVs to the opposite side of delivery to move in favour of the wind.

\textcolor{black}{
\subsection{Coalitional Analysis}
}

This section analyzes the cooperation potential of the multi-modal system in the context of urban on-demand delivery, as well as analyzes the algorithm's cost allocation along different scales of the problem. First, the model's generalization power is investigated, followed by coalitional analysis using the high-quality solution extracted by the trained model of generalizing varying among agents.  

\subsubsection{Coalition Generalizability}

To verify the generalization of our method, we investigate three different configurations determined by the density parameter $\rho$ threshold, the probability of obstacles along each node. First, (a) shows a low connectivity and scattered environment, (b) accounts for medium connectivity and density of the obstacles, and (c) demonstrates a high-density area with most of the nodes connected. The computation time will increase from the former to the latter scenario owing to the existence of more intermediary nodes, which will do the shortest path computation for the UAVs. 

\begin{figure}[!ht]
\centering
     \includegraphics[width=0.6\textwidth]{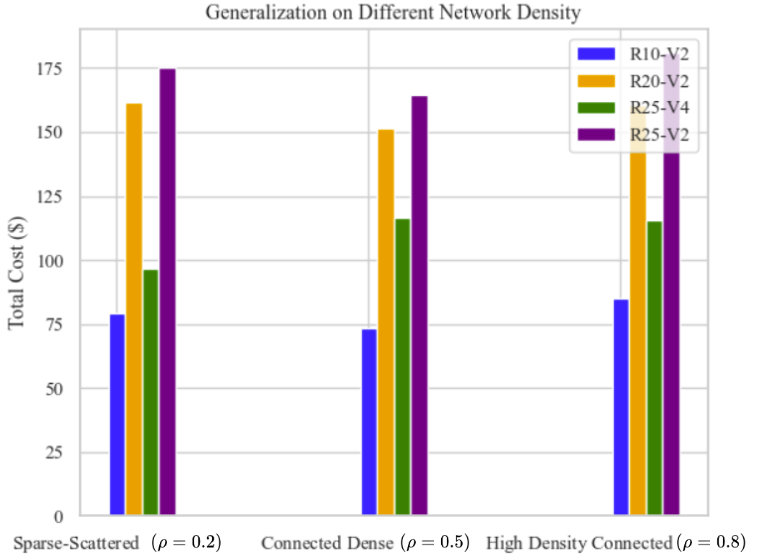}
     \caption{Generation from sparse to fully connected and dense graph network.}
     \label{gen}
\end{figure}

We analyze the trained model's generalization performance, focusing on its ability to handle problems with different configurations: obstacle density and time-proximity network. Figure \ref{gen} demonstrates the generalization result of the problem from 10, 20 and 25 requests on three different network configurations, where the urban environment becomes denser from case (a) to (c) and also becomes fully connected from a sparse time-based weighted-graph-network. We have utilized the trained model for each request with the case (b) criteria and applied that to the 1000 test instances to obtain how these models generate solutions for the two other cases.

The problem configuration shows reasonable performance when generalized to fully connected and high-density cases, as the edge weights are supposed to increase due to a non-existent point-to-point travel path, and UAVs have to use intermediary nodes, thereby higher travel costs. The generalization of the larger case to the smaller case is not expected to work well because of the excessive information carried by the larger case. It takes into account every node connection regardless of their time proximity, which will pose more time for computation on encoding unnecessary datasets sparsified on the lower scale problem. However, even models trained on small-scale problems can achieve comparable results when applied to larger-scale problems, and the cost function will be increased since more graph distance calculations are computed. The core of the performance improvement is due to the nature of the customized graph attention, which gives importance to the time-based graph connectivity encoding, reflected in the decoding scheme to output the closest time window node candidates. This has not been addressed in previous studies. One of the advantages of our study is that for larger-scale problems, it is not inferior, but rather outperforms the computation cost of generalization and scalability, such that it can be used for any type of urban network delivery since it bypasses the municipality and operational constraints, together with incorporating uncertainties. Overall, it can be observed that the cost of sparse-scattering is higher due to higher edge weights from fewer direct connections, which restricts route decision-making flexibility. On the other hand, the fully connected network's primary advantage lies in flexibility since the increased connectivity does not necessarily mean lower costs, as the model already has efficient routes available in the Connected-Dense setup. As a result, we can conclude that more edge connections do not always lead to better cost estimation, although the weight magnitude from the denser case to the scattered case might not work as efficiently as the scattered to the denser edge-weighted graph.

\begin{figure}[!ht]
\centering
     \includegraphics[width=0.8\textwidth]{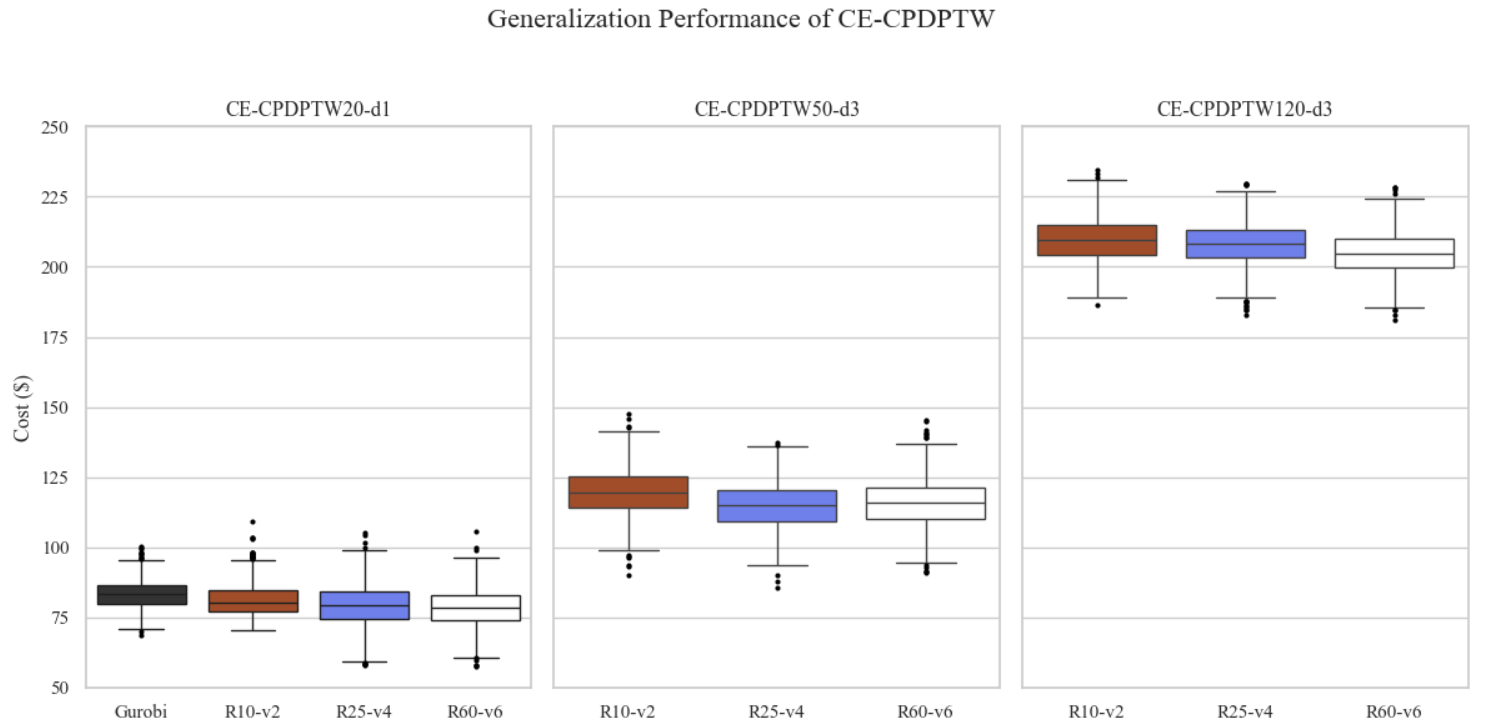}
     \caption{Generalizing on different problem scales.}
     \label{gen1}
\end{figure}

Besides the internal component of different layouts of urban configuration, the model generalization on different problem scales is worth exploring and, in fact, crucial for the time-saving of using the trained model of the smaller network for larger cases, especially generalizing on a different number of vehicles to verify the effectiveness of the proposed model. The policy learnt from the training is designed independently of the number of agents, incorporating mean-pooling of the agent's context embedding and the network embedding. To this end, we use the pre-trained models for three scales of 20, 50, and 120 to generate solutions for smaller to larger and larger to smaller networks. Figure \ref{gen1} demonstrates validating the generalized solution in a box plot for all three networks.

\normalsize

The overall result shows an acceptable solution generated from each of the scales to others; most notably, the generalization on smaller networks from the trained model of larger networks produces better results. This is probably due to the inclusion of smaller and more information on the network embedding. This variation can also be explained by the different number of depots in each case, which causes the training parameter not to capture the depots' embedding well. In addition, as the scale of the problem gets larger, generalization results undergo variability, particularly due to managing the fewer vehicles within the larger network and generalizing from lower vehicle-to-request ratio distributions, which causes ineffectiveness in high request densities assignment with limited resources. Nevertheless, in the constant size of the graph, the model’s ability to generalize is still effective, and in the next part, we analyzed the core game theory for the constant size of the network, with various numbers of agents.

\subsubsection{Coalitional Ability}

We use the pre-trained model for E-CPDPTW120-d3 for ten agents (five of each mode) to conduct a test to find the core coalition among the subset of agents. To evaluate the cost function for each coalition, we use generalization on the pre-trained model to obtain cost values to avoid the high computational expense of training for each coalition. The coalitions are of the two separate modes of ADRs ($r \in\left\{1, \ldots, N^{r}\right\} $) and UAVs ($d \in\left\{1, \ldots, N^{d}\right\}$) set, assumed to have no intersection.

\begin{figure}[ht]
	
\centering

\begin{subfigure}[b]{0.45\textwidth}
	\centering
	\includegraphics[width=0.75\textwidth]{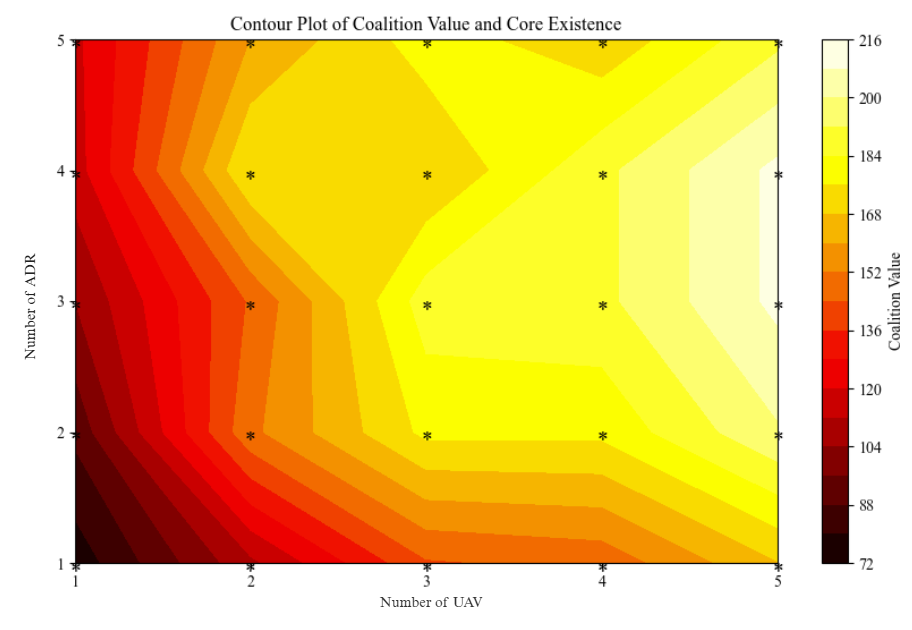}
	\caption{Coalitional analysis for the network size of 50.}
	\label{coal1}
\end{subfigure}

\begin{subfigure}[b]{0.45\textwidth}
	\centering
	\includegraphics[width=0.75\textwidth]{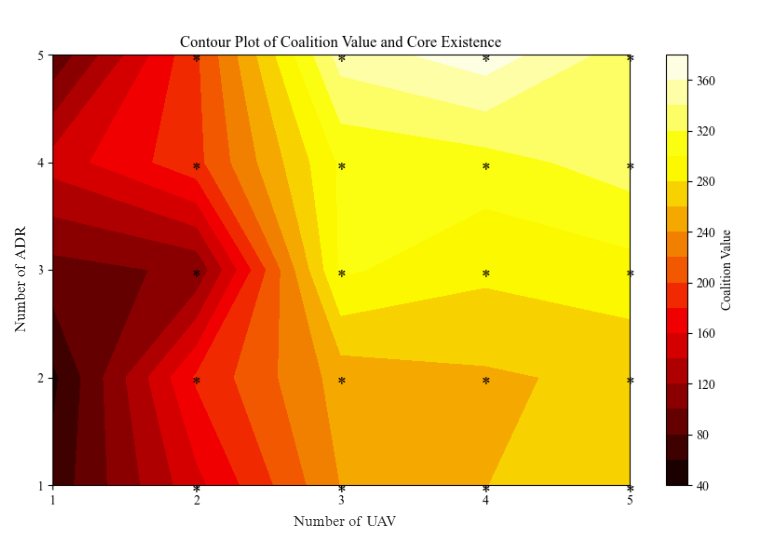}
	\caption{Coalitional analysis for the network size of 120.}
	\label{coal2}
\end{subfigure}
\caption[Core coalition counter-plot for five agents in each mode. The sidebar shows the cost difference and denotes the incentives in forming coalitions and working cooperatively.] {Core coalition counter-plot for five agents in each mode. The sidebar shows the cost difference and denotes the incentives for forming coalitions and working cooperatively.}
\label{coal}
\end{figure}

The Algorithm \ref{core_check} is used here to assess how coalitions will perform and if the core exists. For a group consisting of either UAVs or ADRs, the individual cost of both modes will be calculated, as well as their coalition. Subsequently, the sub-additivity and efficient conditions are checked so that not only must the cooperative cost be less than both individual modes, but also, the newly allocated cost of the coalition must be less than the average mode cost alone. Figure \ref{coal} displays contour plots for coalition gains; the heat map bar shows the difference between the sum of individual and cooperative modes. Note that the spectrum level is set to be sufficiently high to obtain a smooth colour transition graph. However, it is clear that the core analysis plot is discrete. The star sign in the figure indicates if the core exists, given its criteria. It was observed in both scales of the problem that the core exists except in the network size of 120, where only one UAV is operating. This can be mainly because in the larger network, having one UAV independently from the ADR numbers cannot accommodate the demand due to high delays in the deliveries of ADRs, yet it is achievable in the $n=50$. Nonetheless, lower coalition values are observed in both graphs where fewer UAVs are involved, and the lowest value gain is obtained, which fails to perform well in such a large network. It is, however, toward yellowish regions where the most value is gained by acting cooperatively, which is increasing where both the number of UAVs and ADRs increase. This implies that a balanced configuration, often considered a homogeneous coalition, leads to stable and cost-effective matching policies. Furthermore, the non-homogeneous formations can also be a feasible and comparatively fair candidate solution for both sub-figures. In fact, most of the coalitions will fit into the core game of CE-CPDPTW and deploy at least one UAV in the fleet; the requests are delivered in such a network size, yet not as efficiently as operating more ADRs to deal with the high-capacity demands. In other words, in delivery systems, especially in large networks, it is more convenient if one UAV goes to multiple pickup and delivery locations than if one of them is assigned to an ADR if not mandated by a capacity constraint violation. Moreover, the battery consumption rate of UAVs is significantly higher than that of ADRs compared to the current state-of-the-art technology, which leads them to visit recharging stations more frequently and puts more load on urban areas' serviceability. Therefore, using coalitions of ground mode, which can keep away from this shortcoming, is of paramount importance, as can be implied from the result of the core analysis. 

\begin{figure}[ht]
\centering

\begin{subfigure}[b]{0.45\textwidth}
	\centering
	\includegraphics[width=0.75\textwidth]{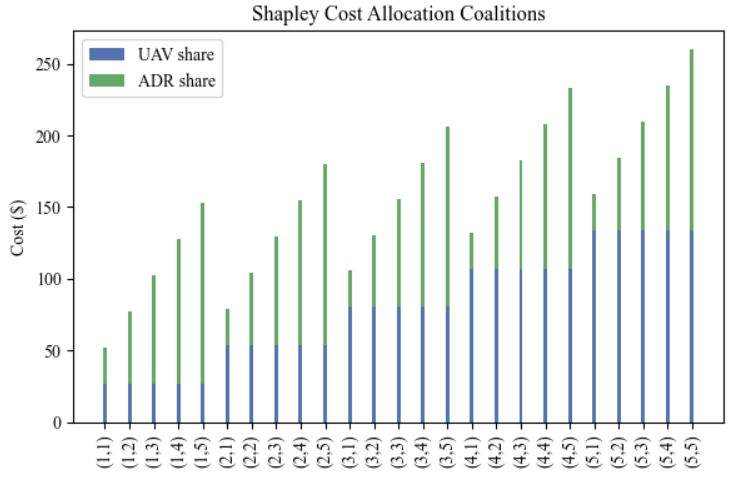}
	\caption{Shapely value for network size of 50.}
	\label{shap2}
\end{subfigure}

\begin{subfigure}[b]{0.45\textwidth}
	\centering
	\includegraphics[width=0.75\textwidth]{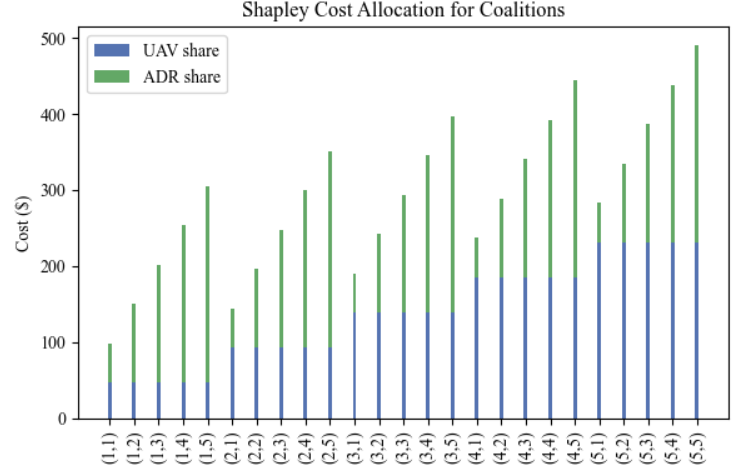}
	\caption{Shapely value for the network size of 120.}
	\label{shap1}
\end{subfigure}

\caption[Cost allocation for all coalitions. The numbers on the $x$ label shows UAV and ADR, respectively.] {Cost allocation for all coalitions. The first number on the left show UAV and second ADR.}
\label{shap}
\end{figure}

\begin{figure}[ht]
	\centering
	
	\begin{subfigure}[b]{0.45\textwidth}
		\centering
		\includegraphics[width=0.75\textwidth]{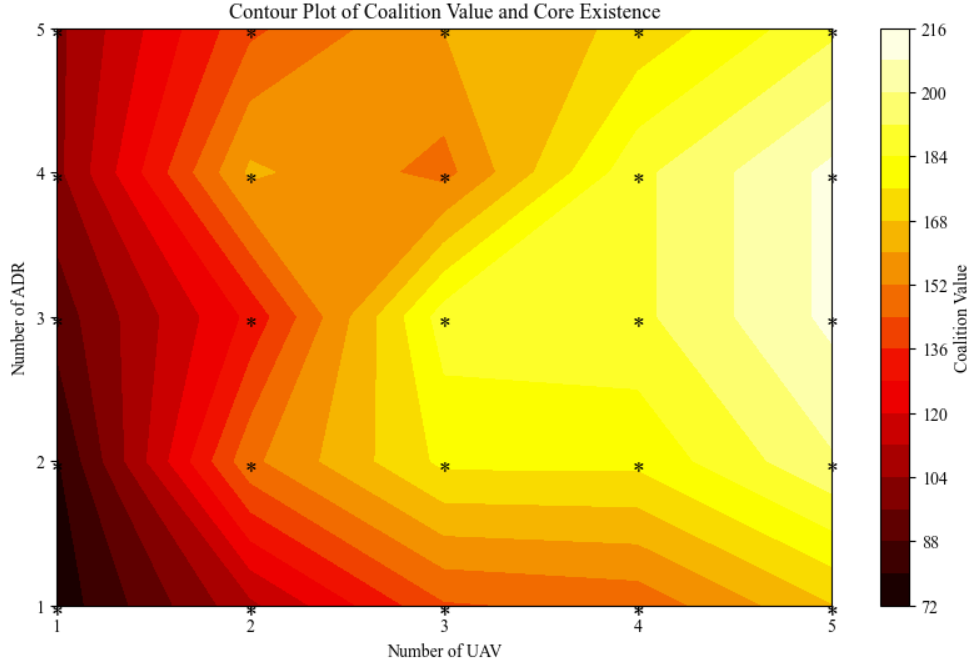}
		\caption{Coalitional analysis for the network size of 50.}
		\label{coala2}
	\end{subfigure}
	
	\begin{subfigure}[b]{0.4\textwidth}
		\centering
		\includegraphics[width=0.75\textwidth]{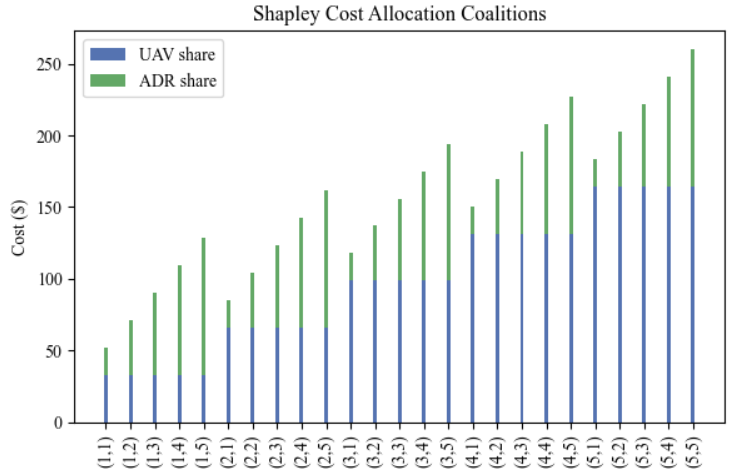}
		\caption{Shapely value for the network size of 50.}
		\label{coala1}
	\end{subfigure}
	
\caption[Cost allocation for non-homogeneous demand distribution.] {Cost allocation for non-homogeneous demand distribution.}
	\label{coala}
\end{figure}

\textcolor{black}{
Additionally, the cost allocation by Shapley value is illustrated in Figure \ref{shap}, distribution of share for each mode across each coalition is shown. }
\textcolor{black}{
The cost share over all coalition is fixed since the UAV and ADR fleet are homogeneous and they represent a higher gain for UAVs on dominance conditions since they provide higher marginal value whereas ADRs are less sensitive to scale and are more efficient in terms of cost sharing in smaller networks. Next, a case of non-homogeneous demand distribution is considered, where payload sizes are from $(3,8)$ bound as well as a gaussian location distribution to experiment the case of deliveries are clustered in dense area. The additivity check and cost allocation are shown in Figure \ref{coala}.}
\textcolor{black}{
In this Figure, the potential marginal contribution for UAVs is subjected to more reducing share in cost due to more load on ADRs. Specifically, they are more effective in coalitions which in case of uniform data distribution was more UAV-oriented. Additionally, cost shares in ADR shows faster as number of agent which accounts for marginal contribution of ADR in such scneraios. }

Finally, according to Figure \ref{coal}, the optimal coalition for both cases is not when all resources are available; instead, the middle regions of the plot, where starting from 3-5 UAVs and 3-5 ADRs, tend to exhibit higher gain. Although it has been inferred that having more vehicles will substantiate the core existence and applicability, it can be achieved in the case of $n=120$ with only $80\%$ of resources. This is also observable in the case of $n=50$ when ADRs are of $80\%$ of resources. Certain scenarios, such as $r=4, d=\{2,3\}$ on the scale of 50 or $r=3, d=\{1,2,3\}$ on the scale of 120, do not support the overall trend of the graph, which tends to gain more values for utilizing more vehicles. The rationale behind this is mainly due to a misfit in the generalization of larger to smaller fleets, which causes the graph not to be smooth and continuously changing. However, the explanation for the former statement of maximum utilization gain is believed to be underlying the fact that systems have found that specific coalition to be more optimal, given the prioritizing of a sufficient number of UAVs to avoid the high cost. Note that not necessarily adding more vehicles will lead to redundancy, where each additional agent contributes less benefit to the coalition and may put extra effort into other vehicles and reduce the overall efficiency. As a result, achieving the optimal coalition value depends on finding the right balance between the number of UAVs and ADRs. \textcolor{black}{This balance can be affected by different test cases, and each mode share can vary significantly, as the recent analysis shows it will be more ADR or UAV needed for larger scale, and in general, the more agents involved, the more efficient the cost allocation results, in addition to the cases which both can complement their delivery task which the other cannot achieve}.


\section{Conclusions and Future Directions}
\label{S:t6}


We present a comprehensive framework for multi-modal autonomous delivery utilizing ADRs and UAVs in urban environments. By addressing critical challenges related to on-demand delivery in high-density areas and efficient vehicle allocation, the proposed framework aims to enhance the accessibility of delivery services. Developing an interactive cooperative allocation system ensures a centralized decision-making process to determine the most effective delivery mode regarding efficiency and time delay.
The stochastic nature of the delivery problem is acknowledged, considering factors such as a heterogeneous fleet interaction in urban areas cluttered with buildings, airspace regulations, and unpredictable weather conditions. 
Our solution tailored a deep reinforcement learning approach, specifically the transformer architecture with \textcolor{black}{edge-enhanced agent-aware attention models capturing spatial and temporal coupling effects, to achieve optimal assignment and routing in urban environments.} By simulating the delivery task for the fleet, our methodology accounts for delivery priority, vehicle states, and environmental changes. Moreover, the coalitional game theory model has been adopted to investigate potential cooperative cost-profit by which various coalitions are formed. In addition, core properties where they proved to exist have been established to identify multi-modal delivery advantages over uni-modal and cost distribution among groups of vehicles.

Our proposed model exhibits superior performance in urban on-demand delivery systems applied in various delivery scenarios, giving importance from customers' to commercial perspectives, utilizing potential cooperative operation of UAVs, where reducing delay costs and handling time-sensitive deliveries is accomplished, and ADRs carrying larger payloads and decreasing the burden on the fleet is done. The problem of CE-CPDPTW has been investigated in a variety of aspects, including robustness tests of operating uncertain and \textcolor{black}{heterogeneous} patterns of time-location distribution as well as wind conditions, which could be potentially utilized in such situations owing to its sustainable performance. \textcolor{black}{In addition, the coalition following the generalization ability of such a mode has been analyzed in various configurations, scales, and multiple agents, demonstrating the cooperative benefit in large-scale and non-homogeneous demand distribution. }  

There are several future directions where the research can be further developed. \textcolor{black}{Firstly, the centralized conflict resolution three-dimensional network plan to manage the UAV traffic throughout the vertical and horizontal aerial links to avoid potential collisions. Exploration of additional factors, such as non-homogeneous sub-fleet vehicles ranging in different capacities and battery sizes, will contribute to a more realistic and prioritized order in the delivery system.} Additionally, the system's scalability in the context of multi-agent systems by extending the framework to handle a larger fleet of autonomous vehicles, diverse urban environments of different layouts, and decentralized execution will be crucial for practical implementation and widespread adoption. \textcolor{black}{ Furthermore, evaluating the proposed on-demand delivery system under varying external factors, such as vehicle failure and split delivery, by incorporating the adaptive generalization techniques, will be a focus of future work.}
\appendix


\section{UAV Energy Consumption}
\label{B}

UAVs and ADRs are powered by batteries, and due to the limited capacity of the battery charge, they need to recharge their battery once they return to depots. Additionally, there are some factors in the urban area where the power consumption of these electric vehicles is not constantly changing, such as the magnitude and direction of the wind, ground friction, and variable speed of vehicles, which all affect the power consumption. For each vehicle, we use a standard model in previous studies. 
For UAVs, the model proposed by \citep{stolaroff2018energy} for power consumption is utilized, and we incorporate a stochastic wind with random turbulent flow from \citep{kimon2015handbook}, where the power consumption is summarized as Equation \ref{pw}.

\begin{equation}
   P=\frac{T\left(v_a \sin \alpha+v_i\right)}{\eta}
   \label{pw}
\end{equation}

where $\alpha$ is the angle of attack, $T$ is thrust, $v_a$ is the UAV speed, $\eta$ is power transfer efficiency, and $v_i$ is the induced speed, which can be found by solving Equation \ref{vi}.

\begin{equation}
   v_i=\frac{g \sum_{k=1}^3 m_k}{2 n \rho \varsigma \sqrt{\left(v_a \cos \alpha\right)^2+\left(v_a \sin \alpha+v_i\right)^2}}
   \label{vi}
\end{equation}
where n is the number of rotors, and $\varsigma$ is the area of the spinning blade disc of one rotor.
Besides, the angle of attack $\alpha$ and thrust are given by Equation \ref{alph} and \ref{thru}, respectively.

\begin{equation}
\alpha=\tan ^{-1}\left(\frac{\frac{1}{2} \rho\left(\sum_{k=1}^3 C_{D_k} A_k\right) v_a{ }^2}{g \sum_{k=1}^3 m_k}\right)
   \label{alph}
\end{equation}

\begin{equation}
T=W+D=g \sum_{k=1}^3 m_k+\frac{1}{2} \rho \sum_{k=1}^3 C_{D_k} A_k v_a^2
   \label{thru}
\end{equation}

The first term reflects payload and UAV weight, and the second term is the parasite drag force, with a coefficient of $CD_k$ and the projected area perpendicular to travel $A_k$ for each UAV component. It is noted that the stochastic wind model in Equation \ref{wind} in which a constant wind is combined with a random turbulence flow is incorporated from \citep{kimon2015handbook}. 

\begin{equation}
   \begin{aligned}
& \dot{x}=v_a \cos \psi+v_w \cos \psi_w \\
& \dot{z}=v_a \sin \psi+v_w \sin \psi_w
\end{aligned}
\label{wind}
\end{equation}

here $v_a$ and $v_w$ indicate the airspeed and wind speed respectively; $\chi=\tan ^{-1}(z / x)$ is the course angle; $\psi$ is the heading angle of the UAV; and $\psi_w$ is the wind course angle. we consider the UAV's velocity to be constant in the course of doing the delivery task; thus, Equation \ref{windd} shows the final corrected velocity.

\begin{equation}
   v_a=\sqrt{2 v_g^2+2 v_w^2-2 v_g v_w \cos \left(\psi_w-\chi\right)}
   \label{windd}
\end{equation}

Furthermore, for the ADR, we use the simple kinematic model and incorporate friction as the only force applied to ADR operation, Equation \ref{pww}, where the parameter is followed by \citep{xiao2014energy}.

\begin{equation}
P = C_r(M + m_{pl})gv/\nu
    \label{pww}
\end{equation}

Where $C_r$ is the friction coefficient, $M$ is the ADR weight, $m_{pl}$ is the payload weight, $v$ and $\nu$ are the velocity and power efficiency, respectively.

\section{Training Algorithm}
\label{training}

The policy gradient network is adopted for the training with Adam optimizer \citep{sutton1999policy}, which algorithm is shown in Algorithm \ref{tabpol}. Since the action space in a routing problem is discrete and expands exponentially as the scale of the problem increases. A policy-based reinforcement method is commonly employed to address this, comprising an actor and a critic network \citep{li2021heterogeneous}. In this method, the actor-network generates a probability vector over all actions based on the current state at each step and selects an action accordingly. This process iterates until the terminal condition is met. The reward for the actor-network is computed by summing up the cumulative rewards at each step throughout the entire process. Serving as a baseline for the actor-network, the critic network calculates the baseline reward solely based on the initial state to reduce variance. Following the receipt of the rewards from the actor-network and the baseline reward from the critic network, the policy gradient method is applied to update the parameters of both networks. In this update, the actor-network is trained to discover solutions of higher quality. 
Critic and rollout baselines are the network for comparison. At each episode, a batch of samples is fed to the RL agent, and by applying a transformer to the instances, the output probability of the decoder generates a candidate sequential node for routing, followed by collecting a reward. Afterwards, the policy gradient estimator will update the parameters $\theta$ with a baseline $\pi^{B}$, using Equation \ref{eqtrain}.

\begin{equation}
\nabla_\theta L(\theta) = \frac{1}{B} \sum_{i=1}^B \left[ \frac{1}{N^{d,r}} \sum_{k=1}^{N^{d,r}} \left(\mathcal{R}(\pi_i) - \mathcal{R}(\pi^B)_k \right) \nabla_{\theta_k} \log p_{\theta_k}(\pi_i) \right]
    \label{eqtrain}
\end{equation}

where $B$ and $R\left(\pi^{B L}\right)$ are the batch size and the baseline reward, respectively. When using a rollout baseline, $\theta$ is replaced by the baseline reward at the end of each episode if the test results are significant with confidence of $95 \%$.

\begin{algorithm}[ht]
\caption{Policy gradient algorithm}
\label{tabpol}
\footnotesize
\begin{algorithmic}[1]

\State { \textbf{Input}: the number of iterations } $N$ \text {; iteration size } $I$ \text {; batch size } $b$ \text {; number of batches } $B=I / b$ 
\text { maximum decoding length } $T$ \text {; t-test threshold } $\alpha$

\State { \textbf{Initialization}: initial parameters } $\theta$ \text { for policy network } $\pi_\theta$ \text { initial parameters } $\varphi$ \text { for policy network } $\pi_{\varphi}$

\State \text { Generate } A \text { E-PDPTW instances randomly }

\State { \textbf{for} epoch }=1,2 \ldots N \text { \textbf{do} }
\State \;\;\;\; \text { Calculate the baseline network solution and reward } $R\left(\pi_{\varphi}\right)$
\State \;\;\;\;\text { \textbf{for} } $i=1,2 \ldots B$ \text { \textbf{do} }
\State \;\;\;\;\;\;\;\;\text { \textbf{for} } $t=1,2 \ldots T$ \text { \textbf{do} }
\State \;\;\;\;\;\;\;\;\;\;\;\;\text { Calculate the output action of policy network step } $t, a_t \sim \pi_\theta\left(a_t \mid s_t\right)$
\State \;\;\;\;\;\;\;\;\;\;\;\;\text { Observe reward } $r_t$ \text { and next state } $S_{t+1}$ \text {; }
\State \;\;\;\;\;\;\;\;\text { \textbf{end} }
\State \;\;\;\;\;\;\;\;\text { Calculate the reward } $R\left(\pi_\theta\right)$ ; 
\State \;\;\;\;\;\;\;\;\text { Calculate gradient } $\nabla_\theta J(\theta)$
\State \;\;\;\;\;\;\;\;\text { Update the parameters; }
\State \;\;\;\; \textbf{end} 
\State \;\;\;\;\text { Calculate the value of the } $\mathrm{t}$ \text {-test } $\varepsilon$ \text {; }
\State \;\;\;\;\text { \textbf{if} } $\varepsilon<\alpha$
\State \;\;\;\;\;\;\;\;\text { Update baseline network parameters } $\varphi \leftarrow \theta$ ; 
\State \;\;\;\; \textbf{end} 
\State   \textbf{end} 
\State   \textbf{end} 

\end{algorithmic}
\end{algorithm}
\normalsize

\bibliographystyle{elsarticle-harv}
\bibliography{traj.bib}

\begin{thebibliography}{67}
\expandafter\ifx\csname natexlab\endcsname\relax\def\natexlab#1{#1}\fi
\providecommand{\url}[1]{\texttt{#1}}
\providecommand{\href}[2]{#2}
\providecommand{\path}[1]{#1}
\providecommand{\DOIprefix}{doi:}
\providecommand{\ArXivprefix}{arXiv:}
\providecommand{\URLprefix}{URL: }
\providecommand{\Pubmedprefix}{pmid:}
\providecommand{\doi}[1]{\href{http://dx.doi.org/#1}{\path{#1}}}
\providecommand{\Pubmed}[1]{\href{pmid:#1}{\path{#1}}}
\providecommand{\bibinfo}[2]{#2}
\ifx\xfnm\relax \def\xfnm[#1]{\unskip,\space#1}\fi
\bibitem[{Ahmad et~al.(2023)Ahmad, Shah and Al-Fagih}]{ahmad2023applications}
\bibinfo{author}{Ahmad, F.}, \bibinfo{author}{Shah, Z.}, \bibinfo{author}{Al-Fagih, L.}, \bibinfo{year}{2023}.
\newblock \bibinfo{title}{Applications of evolutionary game theory in urban road transport network: A state of the art review}.
\newblock \bibinfo{journal}{Sustainable Cities and Society} \bibinfo{volume}{98}, \bibinfo{pages}{104791}.
\bibitem[{Zhang~et al.(2022)}]{zhang2022transformer}
\bibinfo{author}{Zhang~et al., K.}, \bibinfo{year}{2022}.
\newblock \bibinfo{title}{Transformer-based reinforcement learning for pickup and delivery problems with late penalties}.
\newblock \bibinfo{journal}{IEEE Trans. on ITS} \bibinfo{volume}{23}, \bibinfo{pages}{24649--24661}.
\bibitem[{Alfandari et~al.(2022)Alfandari, Ljubi{\'c} and da~Silva}]{alfandari2022tailored}
\bibinfo{author}{Alfandari, L.}, \bibinfo{author}{Ljubi{\'c}, I.}, \bibinfo{author}{da~Silva, M.D.M.}, \bibinfo{year}{2022}.
\newblock \bibinfo{title}{A tailored benders decomposition approach for last-mile delivery with autonomous robots}.
\newblock \bibinfo{journal}{European Journal of Operational Research} \bibinfo{volume}{299}, \bibinfo{pages}{510--525}.
\bibitem[{Archetti and Bertazzi(2021)}]{archetti2021recent}
\bibinfo{author}{Archetti, C.}, \bibinfo{author}{Bertazzi, L.}, \bibinfo{year}{2021}.
\newblock \bibinfo{title}{Recent challenges in routing and inventory routing: E-commerce and last-mile delivery}.
\newblock \bibinfo{journal}{Networks} \bibinfo{volume}{77}, \bibinfo{pages}{255--268}.
\bibitem[{Beliaev et~al.(2023)Beliaev, Mehr and Pedarsani}]{beliaev2023congestion}
\bibinfo{author}{Beliaev, M.}, \bibinfo{author}{Mehr, N.}, \bibinfo{author}{Pedarsani, R.}, \bibinfo{year}{2023}.
\newblock \bibinfo{title}{Congestion-aware bi-modal delivery systems utilizing drones}.
\newblock \bibinfo{journal}{Future Transportation} \bibinfo{volume}{3}, \bibinfo{pages}{329--348}.
\bibitem[{Boeing(2017)}]{boeing2017osmnx}
\bibinfo{author}{Boeing, G.}, \bibinfo{year}{2017}.
\newblock \bibinfo{title}{Osmnx: New methods for acquiring, constructing, analyzing, and visualizing complex street networks}.
\newblock \bibinfo{journal}{Computers, Environment and Urban Systems} \bibinfo{volume}{65}, \bibinfo{pages}{126--139}.
\bibitem[{Bogyrbayeva et~al.(2022)Bogyrbayeva, Meraliyev, Mustakhov and Dauletbayev}]{bogyrbayeva2022learning}
\bibinfo{author}{Bogyrbayeva, A.}, \bibinfo{author}{Meraliyev, M.}, \bibinfo{author}{Mustakhov, T.}, \bibinfo{author}{Dauletbayev, B.}, \bibinfo{year}{2022}.
\newblock \bibinfo{title}{Learning to solve vehicle routing problems: A survey}.
\newblock \bibinfo{journal}{arXiv preprint arXiv:2205.02453} .
\bibitem[{Chalkiadakis et~al.(2022)Chalkiadakis, Elkind and Wooldridge}]{chalkiadakis2022computational}
\bibinfo{author}{Chalkiadakis, G.}, \bibinfo{author}{Elkind, E.}, \bibinfo{author}{Wooldridge, M.}, \bibinfo{year}{2022}.
\newblock \bibinfo{title}{Computational aspects of cooperative game theory}.
\newblock \bibinfo{publisher}{Springer Nature}.
\bibitem[{Chen et~al.(2023)Chen, Wang, Pan, Wu, Zheng and Ding}]{chen2023matching}
\bibinfo{author}{Chen, J.}, \bibinfo{author}{Wang, L.}, \bibinfo{author}{Pan, Z.}, \bibinfo{author}{Wu, Y.}, \bibinfo{author}{Zheng, J.}, \bibinfo{author}{Ding, X.}, \bibinfo{year}{2023}.
\newblock \bibinfo{title}{A matching algorithm with reinforcement learning and decoupling strategy for order dispatching in on-demand food delivery}.
\newblock \bibinfo{journal}{Tsinghua Science and Technology} \bibinfo{volume}{29}, \bibinfo{pages}{386--399}.
\bibitem[{Chu et~al.(2021)Chu, Zhang, Bai and Chen}]{chu2021data}
\bibinfo{author}{Chu, H.}, \bibinfo{author}{Zhang, W.}, \bibinfo{author}{Bai, P.}, \bibinfo{author}{Chen, Y.}, \bibinfo{year}{2021}.
\newblock \bibinfo{title}{Data-driven optimization for last-mile delivery}.
\newblock \bibinfo{journal}{Complex \& Intelligent Systems} , \bibinfo{pages}{1--14}.
\bibitem[{Das et~al.(2020)Das, Sewani, Wang and Tiwari}]{das2020synchronized}
\bibinfo{author}{Das, D.N.}, \bibinfo{author}{Sewani, R.}, \bibinfo{author}{Wang, J.}, \bibinfo{author}{Tiwari, M.K.}, \bibinfo{year}{2020}.
\newblock \bibinfo{title}{Synchronized truck and drone routing in package delivery logistics}.
\newblock \bibinfo{journal}{IEEE Transactions on Intelligent Transportation Systems} \bibinfo{volume}{22}, \bibinfo{pages}{5772--5782}.
\bibitem[{Doole et~al.(2020)Doole, Ellerbroek and Hoekstra}]{doole2020estimation}
\bibinfo{author}{Doole, M.}, \bibinfo{author}{Ellerbroek, J.}, \bibinfo{author}{Hoekstra, J.}, \bibinfo{year}{2020}.
\newblock \bibinfo{title}{Estimation of traffic density from drone-based delivery in very low level urban airspace}.
\newblock \bibinfo{journal}{Journal of Air Transport Management} \bibinfo{volume}{88}, \bibinfo{pages}{101862}.
\bibitem[{Elsayed and Mohamed(2020)}]{elsayed2020impact}
\bibinfo{author}{Elsayed, M.}, \bibinfo{author}{Mohamed, M.}, \bibinfo{year}{2020}.
\newblock \bibinfo{title}{The impact of airspace regulations on unmanned aerial vehicles in last-mile operation}.
\newblock \bibinfo{journal}{Transportation Research Part D: Transport and Environment} \bibinfo{volume}{87}, \bibinfo{pages}{102480}.
\bibitem[{Fellek et~al.(2023)Fellek, Farid, Gebreyesus, Fujimura and Yoshie}]{fellek2023graph}
\bibinfo{author}{Fellek, G.}, \bibinfo{author}{Farid, A.}, \bibinfo{author}{Gebreyesus, G.}, \bibinfo{author}{Fujimura, S.}, \bibinfo{author}{Yoshie, O.}, \bibinfo{year}{2023}.
\newblock \bibinfo{title}{Graph transformer with reinforcement learning for vehicle routing problem}.
\newblock \bibinfo{journal}{IEEJ Transactions on Electrical and Electronic Engineering} \bibinfo{volume}{18}, \bibinfo{pages}{701--713}.
\bibitem[{Fernando et~al.(2023)Fernando, Senanayake, Choi and Swany}]{fernando2023graph}
\bibinfo{author}{Fernando, M.}, \bibinfo{author}{Senanayake, R.}, \bibinfo{author}{Choi, H.}, \bibinfo{author}{Swany, M.}, \bibinfo{year}{2023}.
\newblock \bibinfo{title}{Graph attention multi-agent fleet autonomy for advanced air mobility}.
\newblock \bibinfo{journal}{arXiv preprint arXiv:2302.07337} .
\bibitem[{Fortune(2023)}]{MarketResearchReport}
\bibinfo{author}{Fortune, B.}, \bibinfo{year}{2023}.
\newblock \bibinfo{title}{Delivery robots market size, share, and growth}.
\newblock \bibinfo{howpublished}{\url{https://www.marketsandmarkets.com/Market-Reports/delivery-robot-market-263997316.html/}}.
\newblock \bibinfo{note}{Accessed: 2024-10-20}.
\bibitem[{Fuertes et~al.(2023)Fuertes, del Blanco, Jaureguizar, Navarro and Garc{\'\i}a}]{fuertes2023solving}
\bibinfo{author}{Fuertes, D.}, \bibinfo{author}{del Blanco, C.R.}, \bibinfo{author}{Jaureguizar, F.}, \bibinfo{author}{Navarro, J.J.}, \bibinfo{author}{Garc{\'\i}a, N.}, \bibinfo{year}{2023}.
\newblock \bibinfo{title}{Solving routing problems for multiple cooperative unmanned aerial vehicles using transformer networks}.
\newblock \bibinfo{journal}{Engineering Applications of Artificial Intelligence} \bibinfo{volume}{122}, \bibinfo{pages}{106085}.
\bibitem[{Gansterer and Hartl(2020)}]{gansterer2020shared}
\bibinfo{author}{Gansterer, M.}, \bibinfo{author}{Hartl, R.F.}, \bibinfo{year}{2020}.
\newblock \bibinfo{title}{Shared resources in collaborative vehicle routing}.
\newblock \bibinfo{journal}{Top} \bibinfo{volume}{28}, \bibinfo{pages}{1--20}.
\bibitem[{Gu et~al.(2023)Gu, Liu and Poon}]{gu2023dynamic}
\bibinfo{author}{Gu, R.}, \bibinfo{author}{Liu, Y.}, \bibinfo{author}{Poon, M.}, \bibinfo{year}{2023}.
\newblock \bibinfo{title}{Dynamic truck--drone routing problem for scheduled deliveries and on-demand pickups with time-related constraints}.
\newblock \bibinfo{journal}{Transportation Research Part C: Emerging Technologies} \bibinfo{volume}{151}, \bibinfo{pages}{104139}.
\bibitem[{He et~al.(2022)He, He, Li, Zhang and Xiao}]{he2022route}
\bibinfo{author}{He, X.}, \bibinfo{author}{He, F.}, \bibinfo{author}{Li, L.}, \bibinfo{author}{Zhang, L.}, \bibinfo{author}{Xiao, G.}, \bibinfo{year}{2022}.
\newblock \bibinfo{title}{A route network planning method for urban air delivery}.
\newblock \bibinfo{journal}{Transportation Research Part E: Logistics and Transportation Review} \bibinfo{volume}{166}, \bibinfo{pages}{102872}.
\bibitem[{Jahanshahi et~al.(2022)Jahanshahi, Bozanta, Cevik, Kavuk, Tosun, Sonuc, Kosucu and Ba{\c{s}}ar}]{jahanshahi2022deep}
\bibinfo{author}{Jahanshahi, H.}, \bibinfo{author}{Bozanta, A.}, \bibinfo{author}{Cevik, M.}, \bibinfo{author}{Kavuk, E.M.}, \bibinfo{author}{Tosun, A.}, \bibinfo{author}{Sonuc, S.B.}, \bibinfo{author}{Kosucu, B.}, \bibinfo{author}{Ba{\c{s}}ar, A.}, \bibinfo{year}{2022}.
\newblock \bibinfo{title}{A deep reinforcement learning approach for the meal delivery problem}.
\newblock \bibinfo{journal}{Knowledge-Based Systems} \bibinfo{volume}{243}, \bibinfo{pages}{108489}.
\bibitem[{James et~al.(2019)James, Yu and Gu}]{james2019online}
\bibinfo{author}{James, J.}, \bibinfo{author}{Yu, W.}, \bibinfo{author}{Gu, J.}, \bibinfo{year}{2019}.
\newblock \bibinfo{title}{Online vehicle routing with neural combinatorial optimization and deep reinforcement learning}.
\newblock \bibinfo{journal}{IEEE Transactions on Intelligent Transportation Systems} \bibinfo{volume}{20}, \bibinfo{pages}{3806--3817}.
\bibitem[{Jeong et~al.(2019)Jeong, Song and Lee}]{jeong2019truck}
\bibinfo{author}{Jeong, H.Y.}, \bibinfo{author}{Song, B.D.}, \bibinfo{author}{Lee, S.}, \bibinfo{year}{2019}.
\newblock \bibinfo{title}{Truck-drone hybrid delivery routing: Payload-energy dependency and no-fly zones}.
\newblock \bibinfo{journal}{International Journal of Production Economics} \bibinfo{volume}{214}, \bibinfo{pages}{220--233}.
\bibitem[{Johnson(1985)}]{johnson1985wind}
\bibinfo{author}{Johnson, G.L.}, \bibinfo{year}{1985}.
\newblock \bibinfo{title}{Wind energy systems}.
\newblock \bibinfo{publisher}{Citeseer}.
\bibitem[{Kim et~al.(2024)Kim, Jeong and Lee}]{kim2024drone}
\bibinfo{author}{Kim, Y.}, \bibinfo{author}{Jeong, H.Y.}, \bibinfo{author}{Lee, S.}, \bibinfo{year}{2024}.
\newblock \bibinfo{title}{Drone delivery problem with multi-flight level: Machine learning based solution approach}.
\newblock \bibinfo{journal}{Computers \& Industrial Engineering} \bibinfo{volume}{197}, \bibinfo{pages}{110565}.
\bibitem[{Kimon and George(2015)}]{kimon2015handbook}
\bibinfo{author}{Kimon, P.}, \bibinfo{author}{George, J.}, \bibinfo{year}{2015}.
\newblock \bibinfo{title}{Handbook of unmanned aerial vehicles}.
\bibitem[{Kool et~al.(2018)Kool, Van~Hoof and Welling}]{kool2018attention}
\bibinfo{author}{Kool, W.}, \bibinfo{author}{Van~Hoof, H.}, \bibinfo{author}{Welling, M.}, \bibinfo{year}{2018}.
\newblock \bibinfo{title}{Attention, learn to solve routing problems!}
\newblock \bibinfo{journal}{arXiv preprint arXiv:1803.08475} .
\bibitem[{Lei et~al.(2022)Lei, Guo, Wang, Wu and Zhao}]{lei2022solve}
\bibinfo{author}{Lei, K.}, \bibinfo{author}{Guo, P.}, \bibinfo{author}{Wang, Y.}, \bibinfo{author}{Wu, X.}, \bibinfo{author}{Zhao, W.}, \bibinfo{year}{2022}.
\newblock \bibinfo{title}{Solve routing problems with a residual edge-graph attention neural network}.
\newblock \bibinfo{journal}{Neurocomputing} \bibinfo{volume}{508}, \bibinfo{pages}{79--98}.
\bibitem[{Li et~al.(2022)Li, Wu, He, Fan and Pedrycz}]{li2022overview}
\bibinfo{author}{Li, B.}, \bibinfo{author}{Wu, G.}, \bibinfo{author}{He, Y.}, \bibinfo{author}{Fan, M.}, \bibinfo{author}{Pedrycz, W.}, \bibinfo{year}{2022}.
\newblock \bibinfo{title}{An overview and experimental study of learning-based optimization algorithms for the vehicle routing problem}.
\newblock \bibinfo{journal}{IEEE/CAA Journal of Automatica Sinica} \bibinfo{volume}{9}, \bibinfo{pages}{1115--1138}.
\bibitem[{Li et~al.(2021)Li, Xin, Cao, Lim, Song and Zhang}]{li2021heterogeneous}
\bibinfo{author}{Li, J.}, \bibinfo{author}{Xin, L.}, \bibinfo{author}{Cao, Z.}, \bibinfo{author}{Lim, A.}, \bibinfo{author}{Song, W.}, \bibinfo{author}{Zhang, J.}, \bibinfo{year}{2021}.
\newblock \bibinfo{title}{Heterogeneous attentions for solving pickup and delivery problem via deep reinforcement learning}.
\newblock \bibinfo{journal}{IEEE Transactions on Intelligent Transportation Systems} \bibinfo{volume}{23}, \bibinfo{pages}{2306--2315}.
\bibitem[{Liu et~al.(2023)Liu, Shin and Tsourdos}]{liu2023edge}
\bibinfo{author}{Liu, R.}, \bibinfo{author}{Shin, H.S.}, \bibinfo{author}{Tsourdos, A.}, \bibinfo{year}{2023}.
\newblock \bibinfo{title}{Edge-enhanced attentions for drone delivery in presence of winds and recharging stations}.
\newblock \bibinfo{journal}{Journal of Aerospace Information Systems} \bibinfo{volume}{20}, \bibinfo{pages}{216--228}.
\bibitem[{Liu(2019)}]{liu2019optimization}
\bibinfo{author}{Liu, Y.}, \bibinfo{year}{2019}.
\newblock \bibinfo{title}{An optimization-driven dynamic vehicle routing algorithm for on-demand meal delivery using drones}.
\newblock \bibinfo{journal}{Computers \& Operations Research} \bibinfo{volume}{111}, \bibinfo{pages}{1--20}.
\bibitem[{L{\"o}wens et~al.(2022)L{\"o}wens, Ashraf, Gembus, Cuizon, Falkner and Schmidt-Thieme}]{lowens2022solving}
\bibinfo{author}{L{\"o}wens, C.}, \bibinfo{author}{Ashraf, I.}, \bibinfo{author}{Gembus, A.}, \bibinfo{author}{Cuizon, G.}, \bibinfo{author}{Falkner, J.K.}, \bibinfo{author}{Schmidt-Thieme, L.}, \bibinfo{year}{2022}.
\newblock \bibinfo{title}{Solving the traveling salesperson problem with precedence constraints by deep reinforcement learning}, in: \bibinfo{booktitle}{German Conference on Artificial Intelligence (K{\"u}nstliche Intelligenz)}, \bibinfo{organization}{Springer}. pp. \bibinfo{pages}{160--172}.
\bibitem[{Lui(2010)}]{lui2010introduction}
\bibinfo{author}{Lui, J.C.}, \bibinfo{year}{2010}.
\newblock \bibinfo{title}{Introduction to game theory: Cooperative games}.
\newblock \bibinfo{journal}{Department of Computer Science \& Engineering} .
\bibitem[{LVMTech(2025)}]{lvmtechNavigatingSuccess}
\bibinfo{author}{LVMTech}, \bibinfo{year}{2025}.
\newblock \bibinfo{title}{{N}avigating {S}uccess: {R}evolutionizing {L}ast-{M}ile {D}elivery with {F}leet {M}anagement {S}olutions - {L}{V}{M} {T}ech --- lvmtech.com}.
\newblock \bibinfo{howpublished}{\url{https://www.lvmtech.com/navigating-success-revolutionizing-last-mile-delivery-with-fleet-management-solutions/}}.
\newblock \bibinfo{note}{[Accessed 10-07-2025]}.
\bibitem[{Mehra et~al.(2023)Mehra, Saha, Raychoudhury and Mathur}]{mehra2023deliverai}
\bibinfo{author}{Mehra, A.}, \bibinfo{author}{Saha, S.}, \bibinfo{author}{Raychoudhury, V.}, \bibinfo{author}{Mathur, A.}, \bibinfo{year}{2023}.
\newblock \bibinfo{title}{Deliverai: Reinforcement learning based distributed path-sharing network for food deliveries}.
\newblock \bibinfo{journal}{arXiv preprint arXiv:2311.02017} .
\bibitem[{Mohamed~Salleh et~al.(2018)Mohamed~Salleh, Wanchao, Wang, Huang, Tan, Huang and Low}]{mohamed2018preliminary}
\bibinfo{author}{Mohamed~Salleh, M.F.B.}, \bibinfo{author}{Wanchao, C.}, \bibinfo{author}{Wang, Z.}, \bibinfo{author}{Huang, S.}, \bibinfo{author}{Tan, D.Y.}, \bibinfo{author}{Huang, T.}, \bibinfo{author}{Low, K.H.}, \bibinfo{year}{2018}.
\newblock \bibinfo{title}{Preliminary concept of adaptive urban airspace management for unmanned aircraft operations}, in: \bibinfo{booktitle}{2018 AIAA Information Systems-AIAA Infotech@ Aerospace}, p. \bibinfo{pages}{2260}.
\bibitem[{Mohammad et~al.(2023)Mohammad, Nazih~Diab, Elomri and Triki}]{mohammad2023innovative}
\bibinfo{author}{Mohammad, W.A.}, \bibinfo{author}{Nazih~Diab, Y.}, \bibinfo{author}{Elomri, A.}, \bibinfo{author}{Triki, C.}, \bibinfo{year}{2023}.
\newblock \bibinfo{title}{Innovative solutions in last mile delivery: concepts, practices, challenges, and future directions}, in: \bibinfo{booktitle}{Supply Chain Forum: An International Journal}, \bibinfo{organization}{Taylor \& Francis}. pp. \bibinfo{pages}{151--169}.
\bibitem[{Nazari et~al.(2018)Nazari, Oroojlooy, Snyder and Tak{\'a}c}]{nazari2018reinforcement}
\bibinfo{author}{Nazari, M.}, \bibinfo{author}{Oroojlooy, A.}, \bibinfo{author}{Snyder, L.}, \bibinfo{author}{Tak{\'a}c, M.}, \bibinfo{year}{2018}.
\newblock \bibinfo{title}{Reinforcement learning for solving the vehicle routing problem}.
\newblock \bibinfo{journal}{Advances in neural information processing systems} \bibinfo{volume}{31}.
\bibitem[{Osicka et~al.(2020)Osicka, Guajardo and van Oost}]{osicka2020cooperative}
\bibinfo{author}{Osicka, O.}, \bibinfo{author}{Guajardo, M.}, \bibinfo{author}{van Oost, T.}, \bibinfo{year}{2020}.
\newblock \bibinfo{title}{Cooperative game-theoretic features of cost sharing in location-routing}.
\newblock \bibinfo{journal}{International Transactions in Operational Research} \bibinfo{volume}{27}, \bibinfo{pages}{2157--2183}.
\bibitem[{Osler(2021)}]{Dronlaw}
\bibinfo{author}{Osler, Hoskin, H.}, \bibinfo{year}{2021}.
\newblock \bibinfo{title}{Drone law in canada}.
\newblock \URLprefix \url{https://www.osler.com/osler/media/Osler/infographics/CG5049_Drone-Law-Canada.pdf}. \bibinfo{note}{accessed: 2023-11-30}.
\bibitem[{Ostermeier et~al.(2023)Ostermeier, Heimfarth and H{\"u}bner}]{ostermeier2023multi}
\bibinfo{author}{Ostermeier, M.}, \bibinfo{author}{Heimfarth, A.}, \bibinfo{author}{H{\"u}bner, A.}, \bibinfo{year}{2023}.
\newblock \bibinfo{title}{The multi-vehicle truck-and-robot routing problem for last-mile delivery}.
\newblock \bibinfo{journal}{European Journal of Operational Research} \bibinfo{volume}{310}, \bibinfo{pages}{680--697}.
\bibitem[{Park et~al.(2021)Park, Bakhtiyar and Park}]{park2021schedulenet}
\bibinfo{author}{Park, J.}, \bibinfo{author}{Bakhtiyar, S.}, \bibinfo{author}{Park, J.}, \bibinfo{year}{2021}.
\newblock \bibinfo{title}{Schedulenet: Learn to solve multi-agent scheduling problems with reinforcement learning}.
\newblock \bibinfo{journal}{arXiv preprint arXiv:2106.03051} .
\bibitem[{Pingale et~al.(2024)Pingale, Kaur and Agarwal}]{pingale2024collaborative}
\bibinfo{author}{Pingale, S.}, \bibinfo{author}{Kaur, A.}, \bibinfo{author}{Agarwal, R.}, \bibinfo{year}{2024}.
\newblock \bibinfo{title}{Collaborative last mile delivery: A two-echelon vehicle routing model with collaboration points}.
\newblock \bibinfo{journal}{Expert Systems with Applications} \bibinfo{volume}{252}, \bibinfo{pages}{124164}.
\bibitem[{Roger et~al.(1991)}]{roger1991game}
\bibinfo{author}{Roger, B.M.}, et~al., \bibinfo{year}{1991}.
\newblock \bibinfo{title}{Game theory: analysis of conflict}.
\newblock \bibinfo{journal}{The President and Fellows of Harvard College, USA} \bibinfo{volume}{66}.
\bibitem[{Roth(1988)}]{roth1988shapley}
\bibinfo{author}{Roth, A.E.}, \bibinfo{year}{1988}.
\newblock \bibinfo{title}{The Shapley value: essays in honor of Lloyd S. Shapley}.
\newblock \bibinfo{publisher}{Cambridge University Press}.
\bibitem[{Samouh et~al.(2020)Samouh, Gluza, Djavadian, Meshkani and Farooq}]{samouh2020multimodal}
\bibinfo{author}{Samouh, F.}, \bibinfo{author}{Gluza, V.}, \bibinfo{author}{Djavadian, S.}, \bibinfo{author}{Meshkani, S.}, \bibinfo{author}{Farooq, B.}, \bibinfo{year}{2020}.
\newblock \bibinfo{title}{Multimodal autonomous last-mile delivery system design and application}, in: \bibinfo{booktitle}{2020 IEEE International Smart Cities Conference (ISC2)}, \bibinfo{organization}{IEEE}. pp. \bibinfo{pages}{1--7}.
\bibitem[{Santiyuda et~al.(2024)Santiyuda, Wardoyo, Pulungan and Vincent}]{santiyuda2024multi}
\bibinfo{author}{Santiyuda, G.}, \bibinfo{author}{Wardoyo, R.}, \bibinfo{author}{Pulungan, R.}, \bibinfo{author}{Vincent, F.Y.}, \bibinfo{year}{2024}.
\newblock \bibinfo{title}{Multi-objective reinforcement learning for bi-objective time-dependent pickup and delivery problem with late penalties}.
\newblock \bibinfo{journal}{Engineering Applications of Artificial Intelligence} \bibinfo{volume}{128}, \bibinfo{pages}{107381}.
\bibitem[{Shi et~al.(2022)Shi, Lin, Li and Li}]{shi2022bi}
\bibinfo{author}{Shi, Y.}, \bibinfo{author}{Lin, Y.}, \bibinfo{author}{Li, B.}, \bibinfo{author}{Li, R.Y.M.}, \bibinfo{year}{2022}.
\newblock \bibinfo{title}{A bi-objective optimization model for the medical supplies' simultaneous pickup and delivery with drones}.
\newblock \bibinfo{journal}{Computers \& Industrial Engineering} \bibinfo{volume}{171}, \bibinfo{pages}{108389}.
\bibitem[{Son et~al.(2023)Son, Kim, Choi and Park}]{son2023solving}
\bibinfo{author}{Son, J.}, \bibinfo{author}{Kim, M.}, \bibinfo{author}{Choi, S.}, \bibinfo{author}{Park, J.}, \bibinfo{year}{2023}.
\newblock \bibinfo{title}{Solving np-hard min-max routing problems as sequential generation with equity context}.
\newblock \bibinfo{journal}{arXiv preprint arXiv:2306.02689} .
\bibitem[{Soroka et~al.(2023)Soroka, Meshcheryakov and Gerasimov}]{soroka2023deep}
\bibinfo{author}{Soroka, A.}, \bibinfo{author}{Meshcheryakov, A.}, \bibinfo{author}{Gerasimov, S.}, \bibinfo{year}{2023}.
\newblock \bibinfo{title}{Deep reinforcement learning for the capacitated pickup and delivery problem with time windows}.
\newblock \bibinfo{journal}{Pattern Recognition and Image Analysis} \bibinfo{volume}{33}, \bibinfo{pages}{169--178}.
\bibitem[{Stolaroff et~al.(2018)Stolaroff, Samaras, O’Neill, Lubers, Mitchell and Ceperley}]{stolaroff2018energy}
\bibinfo{author}{Stolaroff, J.K.}, \bibinfo{author}{Samaras, C.}, \bibinfo{author}{O’Neill, E.R.}, \bibinfo{author}{Lubers, A.}, \bibinfo{author}{Mitchell, A.S.}, \bibinfo{author}{Ceperley, D.}, \bibinfo{year}{2018}.
\newblock \bibinfo{title}{Energy use and life cycle greenhouse gas emissions of drones for commercial package delivery}.
\newblock \bibinfo{journal}{Nature communications} \bibinfo{volume}{9}, \bibinfo{pages}{409}.
\bibitem[{Sudbury and Hutchinson(2016)}]{sudbury2016cost}
\bibinfo{author}{Sudbury, A.W.}, \bibinfo{author}{Hutchinson, E.B.}, \bibinfo{year}{2016}.
\newblock \bibinfo{title}{A cost analysis of amazon prime air (drone delivery)}.
\newblock \bibinfo{journal}{Journal for Economic Educators} \bibinfo{volume}{16}, \bibinfo{pages}{1--12}.
\bibitem[{Sutton et~al.(1999)Sutton, McAllester, Singh and Mansour}]{sutton1999policy}
\bibinfo{author}{Sutton, R.S.}, \bibinfo{author}{McAllester, D.}, \bibinfo{author}{Singh, S.}, \bibinfo{author}{Mansour, Y.}, \bibinfo{year}{1999}.
\newblock \bibinfo{title}{Policy gradient methods for reinforcement learning with function approximation}.
\newblock \bibinfo{journal}{Advances in neural information processing systems} \bibinfo{volume}{12}.
\bibitem[{Vaswani et~al.(2017)Vaswani, Shazeer, Parmar, Uszkoreit, Jones, Gomez, Kaiser and Polosukhin}]{vaswani2017attention}
\bibinfo{author}{Vaswani, A.}, \bibinfo{author}{Shazeer, N.}, \bibinfo{author}{Parmar, N.}, \bibinfo{author}{Uszkoreit, J.}, \bibinfo{author}{Jones, L.}, \bibinfo{author}{Gomez, A.N.}, \bibinfo{author}{Kaiser, {\L}.}, \bibinfo{author}{Polosukhin, I.}, \bibinfo{year}{2017}.
\newblock \bibinfo{title}{Attention is all you need}.
\newblock \bibinfo{journal}{Advances in neural information processing systems} \bibinfo{volume}{30}.
\bibitem[{Velickovic et~al.(2017)Velickovic, Cucurull, Casanova, Romero, Lio, Bengio et~al.}]{velickovic2017graph}
\bibinfo{author}{Velickovic, P.}, \bibinfo{author}{Cucurull, G.}, \bibinfo{author}{Casanova, A.}, \bibinfo{author}{Romero, A.}, \bibinfo{author}{Lio, P.}, \bibinfo{author}{Bengio, Y.}, et~al., \bibinfo{year}{2017}.
\newblock \bibinfo{title}{Graph attention networks}.
\newblock \bibinfo{journal}{stat} \bibinfo{volume}{1050}, \bibinfo{pages}{10--48550}.
\bibitem[{Vinyals et~al.(2015)Vinyals, Fortunato and Jaitly}]{vinyals2015pointer}
\bibinfo{author}{Vinyals, O.}, \bibinfo{author}{Fortunato, M.}, \bibinfo{author}{Jaitly, N.}, \bibinfo{year}{2015}.
\newblock \bibinfo{title}{Pointer networks}.
\newblock \bibinfo{journal}{Advances in neural information processing systems} \bibinfo{volume}{28}.
\bibitem[{Wang et~al.(2023a)Wang, Wang, Dong, Ren and Xing}]{wang2023reinforcement}
\bibinfo{author}{Wang, X.}, \bibinfo{author}{Wang, L.}, \bibinfo{author}{Dong, C.}, \bibinfo{author}{Ren, H.}, \bibinfo{author}{Xing, K.}, \bibinfo{year}{2023}a.
\newblock \bibinfo{title}{Reinforcement learning-based dynamic order recommendation for on-demand food delivery}.
\newblock \bibinfo{journal}{Tsinghua Science and Technology} \bibinfo{volume}{29}, \bibinfo{pages}{356--367}.
\bibitem[{Wang et~al.(2023b)Wang, Zhou, Sun, Fan, Wang and Wang}]{wang2023collaborative}
\bibinfo{author}{Wang, Y.}, \bibinfo{author}{Zhou, J.}, \bibinfo{author}{Sun, Y.}, \bibinfo{author}{Fan, J.}, \bibinfo{author}{Wang, Z.}, \bibinfo{author}{Wang, H.}, \bibinfo{year}{2023}b.
\newblock \bibinfo{title}{Collaborative multidepot electric vehicle routing problem with time windows and shared charging stations}.
\newblock \bibinfo{journal}{Expert Systems with Applications} \bibinfo{volume}{219}, \bibinfo{pages}{119654}.
\bibitem[{Williams(1992)}]{williams1992simple}
\bibinfo{author}{Williams, R.J.}, \bibinfo{year}{1992}.
\newblock \bibinfo{title}{Simple statistical gradient-following algorithms for connectionist reinforcement learning}.
\newblock \bibinfo{journal}{Machine learning} \bibinfo{volume}{8}, \bibinfo{pages}{229--256}.
\bibitem[{Xiao and Whittaker(2014)}]{xiao2014energy}
\bibinfo{author}{Xiao, X.}, \bibinfo{author}{Whittaker, W.}, \bibinfo{year}{2014}.
\newblock \bibinfo{title}{Energy considerations for wheeled mobile robots operating on a single battery discharge}.
\newblock \bibinfo{journal}{Robot. Inst., Carnegie Mellon Univ., Pittsburgh, PA, USA, Tech. Rep., CMU-RI-TR-14-16} .
\bibitem[{Zhang et~al.(2021)Zhang, Campbell, Sweeney~II and Hupman}]{zhang2021energy}
\bibinfo{author}{Zhang, J.}, \bibinfo{author}{Campbell, J.F.}, \bibinfo{author}{Sweeney~II, D.C.}, \bibinfo{author}{Hupman, A.C.}, \bibinfo{year}{2021}.
\newblock \bibinfo{title}{Energy consumption models for delivery drones: A comparison and assessment}.
\newblock \bibinfo{journal}{Transportation Research Part D: Transport and Environment} \bibinfo{volume}{90}, \bibinfo{pages}{102668}.
\bibitem[{Zhang et~al.(2023a)Zhang, Li, Wang, Li and Lin}]{zhang2023two}
\bibinfo{author}{Zhang, K.}, \bibinfo{author}{Li, M.}, \bibinfo{author}{Wang, J.}, \bibinfo{author}{Li, Y.}, \bibinfo{author}{Lin, X.}, \bibinfo{year}{2023}a.
\newblock \bibinfo{title}{A two-stage learning-based method for large-scale on-demand pickup and delivery services with soft time windows}.
\newblock \bibinfo{journal}{Transportation Research Part C: Emerging Technologies} \bibinfo{volume}{151}, \bibinfo{pages}{104122}.
\bibitem[{Zhang et~al.(2023b)Zhang, Lin and Li}]{zhang2023graph}
\bibinfo{author}{Zhang, K.}, \bibinfo{author}{Lin, X.}, \bibinfo{author}{Li, M.}, \bibinfo{year}{2023}b.
\newblock \bibinfo{title}{Graph attention reinforcement learning with flexible matching policies for multi-depot vehicle routing problems}.
\newblock \bibinfo{journal}{Physica A: Statistical Mechanics and its Applications} \bibinfo{volume}{611}, \bibinfo{pages}{128451}.
\bibitem[{Zhang et~al.(2022)Zhang, Wang, Huang, Yu and Fang}]{zhang2022heterogeneous}
\bibinfo{author}{Zhang, Q.}, \bibinfo{author}{Wang, Z.}, \bibinfo{author}{Huang, M.}, \bibinfo{author}{Yu, Y.}, \bibinfo{author}{Fang, S.C.}, \bibinfo{year}{2022}.
\newblock \bibinfo{title}{Heterogeneous multi-depot collaborative vehicle routing problem}.
\newblock \bibinfo{journal}{Transportation Research Part B: Methodological} \bibinfo{volume}{160}, \bibinfo{pages}{1--20}.
\bibitem[{Zibaei et~al.(2016)Zibaei, Hafezalkotob and Ghashami}]{zibaei2016cooperative}
\bibinfo{author}{Zibaei, S.}, \bibinfo{author}{Hafezalkotob, A.}, \bibinfo{author}{Ghashami, S.S.}, \bibinfo{year}{2016}.
\newblock \bibinfo{title}{Cooperative vehicle routing problem: an opportunity for cost saving}.
\newblock \bibinfo{journal}{Journal of Industrial Engineering International} \bibinfo{volume}{12}, \bibinfo{pages}{271--286}.
\bibitem[{Zong et~al.(2022)Zong, Zheng, Li and Jin}]{zong2022mapdp}
\bibinfo{author}{Zong, Z.}, \bibinfo{author}{Zheng, M.}, \bibinfo{author}{Li, Y.}, \bibinfo{author}{Jin, D.}, \bibinfo{year}{2022}.
\newblock \bibinfo{title}{Mapdp: Cooperative multi-agent reinforcement learning to solve pickup and delivery problems}, in: \bibinfo{booktitle}{Proceedings of the AAAI Conference on Artificial Intelligence}, pp. \bibinfo{pages}{9980--9988}.

\end{thebibliography}
\end{document}